\documentclass{article}

\usepackage{styles/jfrExamplee}
\usepackage{graphicx}
\usepackage{subfig}
\usepackage{setspace}
\usepackage{epstopdf}
\usepackage{enumitem} 		% easier-to-edit lists
\usepackage[tight]{units} 	% includes nicefrac package
\usepackage{amsmath}		% Needed for align environment
\usepackage{amssymb}		% Needed for \lesssim symbol
\usepackage{algorithm}		% algorithm-environment
\usepackage{algpseudocode}  % For \algorithmic
\usepackage{gensymb}		% \degree command
\usepackage{booktabs}  		% Allows layouting nicer tables
\usepackage{multirow}		% Multirows in tables
\usepackage{caption}		% Make captions width-adjustable
\usepackage{tikz}			% Tikz-figures
\usetikzlibrary{shapes,arrows,quotes}
\usetikzlibrary{calc}
\DeclareMathAlphabet{\pazocal}{OMS}{zplm}{m}{n}
\DeclareMathAlphabet{\mathcal}{OMS}{cmsy}{m}{n} % to make sure L1 stays to the standard mathcal symbol
\usepackage[toc,page]{appendix}		% Appendices
\usepackage[bottom]{footmisc}		% Load before hyperref to avoid problems
\usepackage[colorlinks=true,citecolor=blue]{hyperref} %Hyperlinks for refs&cites in the PDF
\usepackage{mathtools}		% \coloneq (define as equal to) operator and more
\usepackage{bm}				% For \bm
\usepackage[capitalize,noabbrev]{cleveref} % CleverRefs ; Must come last
\usepackage[nolist,nohyperlinks]{acronym}
\usepackage{array}
\usepackage{pseudocode}  	% For \pseudocode
\usepackage[sort&compress,sectionbib,numbers]{natbib}

\crefname{equation}{Eq.}{Eqs.}
\Crefname{equation}{Equation}{Equations}
\crefname{appsec}{Appendix}{Appendices}
\crefname{pseudocode}{algorithm}{algorithms}
\Crefname{pseudocode}{Algorithm}{Algorithms}

\usepackage{color} % Colors and highlighting in the sould package
\usepackage{soul} % Proper text highlighting with the \hl command
\sethlcolor{green}
\definecolor{OliveGreen}{rgb}{0,0.6,0}

\DeclareMathAlphabet{\pazocal}{OMS}{zplm}{m}{n}
\let\oldnl\nl% Store \nl in \oldnl
\newcommand{\nonl}{\renewcommand{\nl}{\let\nl\oldnl}}% Remove line number for one line

\newcommand{\matr}[1]{\bm{#1}}
\def\code#1{\texttt{#1}}
\newcommand{\note}[1]{} %for not displaying notes

\makeatletter
\newcommand{\nosemic}{\renewcommand{\@endalgocfline}{\relax}}% Drop semi-colon ;
\newcommand{\dosemic}{\renewcommand{\@endalgocfline}{\algocf@endline}}% Reinstate semi-colon ;
\renewcommand\paragraph{\@startsection{paragraph}{4}{\z@}{-1.5ex plus -0.5ex minus -.2ex}{0.5ex plus .2ex}{\normalsize\itshape\raggedright}}
\makeatother

\setlist[itemize]{itemsep=0pt,parsep=3pt,partopsep=0pt, topsep=3pt, leftmargin=25pt,rightmargin=25pt}
\setlist[enumerate]{itemsep=0pt,parsep=3pt,partopsep=0pt,topsep=0pt,leftmargin=25pt,rightmargin=25pt}

\newcolumntype{L}[1]{>{\raggedright\arraybackslash}p{#1}} %Defines new table column type which is ragged right

\title{Towards Fully Environment-Aware UAVs: Real-Time Path Planning with Online 3D Wind Field Prediction in Complex Terrain}
\author{
\begin{minipage}{35em}
\begin{center}
Philipp Oettershagen, Florian Achermann, Benjamin M{\"u}ller, Daniel Schneider and Roland Siegwart
\end{center}
\end{minipage} \\
\\
Autonomous Systems Lab\\Swiss Federal Institute of Technology Zurich (ETH Zurich)\\Leonhardstrasse 21\\ 8092 Zurich\\+41 44 632 7395\\
\texttt{philipp.oettershagen@mavt.ethz.ch} \\
}

\begin{document}

\maketitle

\begin{abstract}
Today, low-altitude fixed-wing Unmanned Aerial Vehicles (UAVs) are largely limited to primitively follow user-defined waypoints. To allow fully-autonomous remote missions in complex environments, real-time environment-aware navigation is required both with respect to terrain and wind, which can easily exceed the aircraft maximum airspeed or vertical speed in complex terrain. This paper presents two relevant initial contributions: First, the literature's first-ever local 3D wind field prediction method which can run in real time onboard a UAV is presented. The simple downscaling approach retrieves low-resolution global weather data, and uses potential flow theory to adjust the wind field such that terrain boundaries, mass conservation, and the atmospheric stratification are observed. Synthetic test cases show good qualitative results. A comparison with 1D LIDAR data shows an overall wind error reduction of 23\% with respect to the zero-wind assumption that is mostly used for UAV path planning today. However, given that the vertical winds are not resolved accurately enough further research is required and identified. Second, a sampling-based path planner that considers the aircraft dynamics in non-uniform wind iteratively via Dubins airplane paths is presented. Performance optimizations, e.g. obstacle-aware sampling and fast 2.5D-map collision checks, render the planner 50\% faster than the Open Motion Planning Library (OMPL) implementation. Synthetic and simulated test cases in Alpine terrain show that the wind-aware planning performs up to 50x less iterations than shortest-path planning and is thus slower in low winds, but that it tends to deliver lower-cost paths in stronger winds. More importantly, in contrast to the shortest-path planner, it always delivers collision-free paths. Overall, our initial research demonstrates the feasibility of 3D wind field prediction from a UAV and the advantages of wind-aware planning. This paves the way for follow-up research on fully-autonomous environment-aware navigation of UAVs in real-life missions and complex terrain.
\end{abstract}

\begin{acronym}
\acro{EKF}{Extended Kalman Filter}
\acro{UKF}{Unscented Kalman Filter}
\acro{SLAM}{Simultaneous Localization And Mapping}
\acro{IMU}{Inertial Measurement Unit}
\acro{2D}{two-dimensional}
\acro{2.5D}{2.5-dimensional}
\acro{3D}{three-dimensional}
\acro{6D}{six-dimensional}
\acro{LiDAR}{Light Detection And Ranging}
\acro{UAV}{Unmanned Aerial Vehicle}
\acro{STC}{Standard Test Conditions}
\acro{ESC}{Electronic Speed Control}
\acro{MPPT}{Maximum Power Point Tracker}
\acro{AoI}{Angle of Incidence}
\acro{FM}{Full Model}
\acro{CAM}{Conceptual Analysis Model}
\acro{CDM}{Conceptual Design Model}
\acro{CFD}{Computation Fluid Dynamics}
\acro{DEM}{Digital Elevation Model}
\acro{DP}{Dynamic Programming}
\acro{DoF}{Degrees of Freedom}
\acro{ABL}{Atmospheric Boundary Layer}
\acro{FEM}{Finite Element Method}
\acro{PDE}{Partial Differential Equation}
\acro{NWP}{Numerical Weather Prediction}
\acro{ROS}{Robot Operating System}
\acro{RMSE}{Root Mean Square Error}
\acro{COSMO}{Consortium for Small-Scale Modeling}
\acro{ECMWF}{European Centre for Medium-Range Weather Forecasts}
\acro{ZHAW}{Consortium for Small-Scale Modeling}
\acro{TSP}{Traveling Salesman Problem}
\acro{TDATSP}{Time Dependent Asymmetric Traveling Salesman Problem}
\acro{SPP}{Shortest Path Problem}
\acro{ZHAW}{Consortium for Small-Scale Modeling}
\acro{SoC}{State of Charge}
\acro{GPS}{Global Positioning System}
\acro{OMPL}{Open Motion Planning Library}
\acro{HIL}{Hardware-in-the-loop}
\acro{SAR}{Search and Rescue}
\acro{PID}{Proportional-Integral-Derivative}
\acro{FCL}{Flexible Collision Library}
\end{acronym}

%%%%%%%%%%%% Just copy stuff from the thesis here

%TODO
% Florian general comments
% Names of test cases check. E.g. in Figure 4.24 it is different than in text.

%\acresetall

%%%%%%%%%%%%%%%%%%%%%%%%%%%%%%%%%%%%%%%%%%%%%%%%%%%%%%%%%%%%%%%%%
%%%%%%%%%%%%%%%%%%%%%%%%%%%%%%%%%%%%%%%%%%%%%%%%%%%%%%%%%%%%%%%%%
\section{Introduction}
\label{sec:PL_Introduction}
%%%%%%%%%%%%%%%%%%%%%%%%%%%%%%%%%%%%%%%%%%%%%%%%%%%%%%%%%%%%%%%%%
%%%%%%%%%%%%%%%%%%%%%%%%%%%%%%%%%%%%%%%%%%%%%%%%%%%%%%%%%%%%%%%%%

%%%%%%%%%%%%%%%%%%%%%%%%%%%%%%%%%%%%%%%%%%%%%%%%%%%%%%%%%%%%%%%%%
\paragraph{Motivation}
\label{sec:PL_Intro_Motivation}
%%%%%%%%%%%%%%%%%%%%%%%%%%%%%%%%%%%%%%%%%%%%%%%%%%%%%%%%%%%%%%%%%

\begin{figure}[htb]
\centering
\includegraphics[width=0.85\columnwidth]{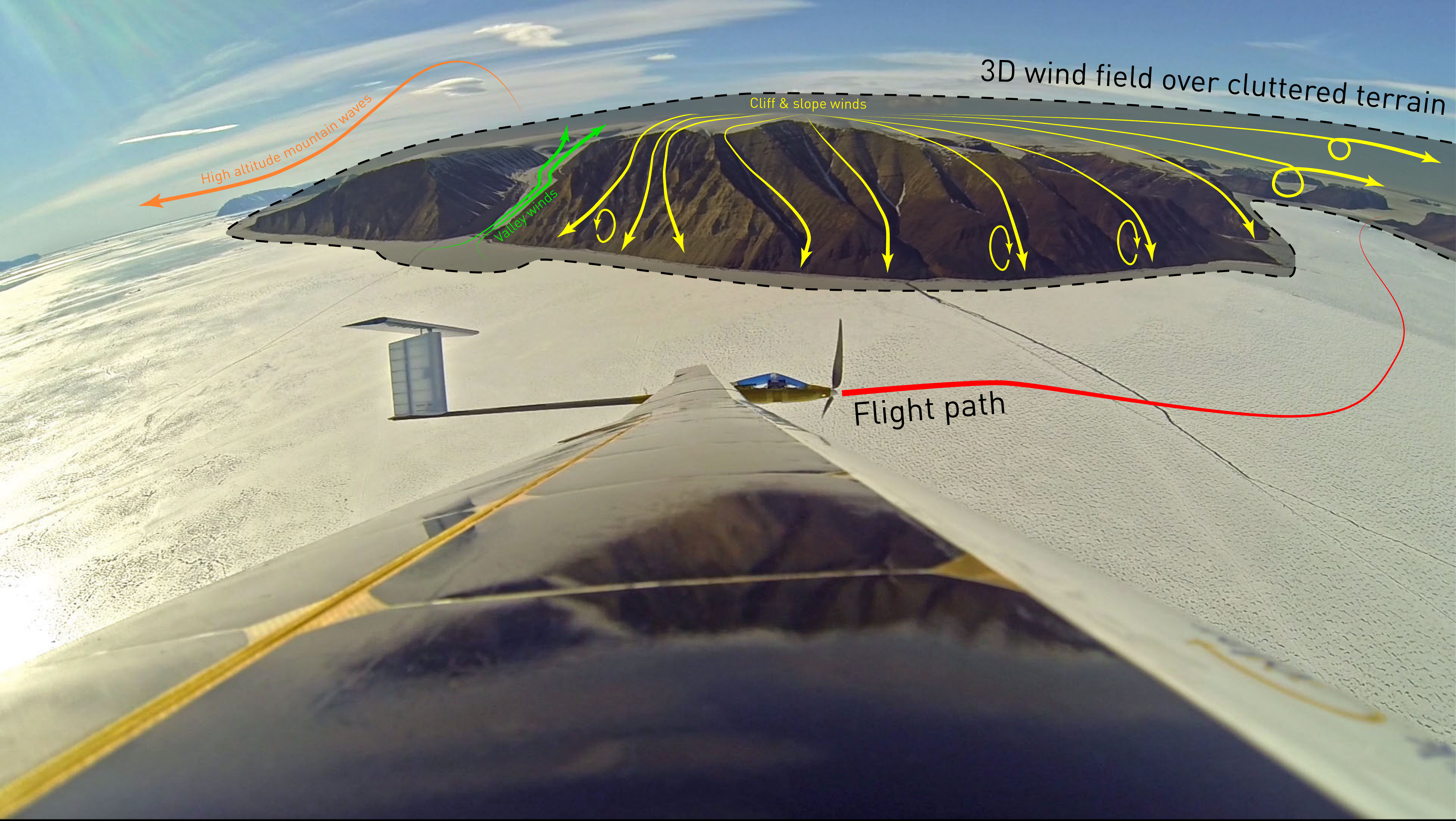}
\caption{
Cluttered terrain and strong winds pose a significant risk to a \ac{UAV}. This paper therefore presents a downscaling method to predict the local 3D wind field in real time and a path planner to calculate optimal trajectories with respect to the perceived terrain and the calculated 3D wind field. \emph{AtlantikSolar}'s flights next to the mountains and strong winds above the Arctic in summer 2017 are a prime application for such a framework.
}
\label{fig:PL_Intro_Collage}
\end{figure}

The aerial scanning capabilities provided by \acp{UAV} are of significant benefit in search-and-rescue support, agricultural sensing, industrial inspection or border patrol~\cite{NASA_Pathfinder}. However, today's \acp{UAV} are largely limited to a primitive set of waypoint-following tasks and have little awareness of the environment in which they fly. As a result, today, low-altitude \ac{UAV} missions are mostly performed on a small scale. To penetrate into applications that can be of pivotal societal and commercial use, UAVs need to be capable of fully autonomous large-scale operations including beyond-visual-line-of-sight (BVLOS) conditions. In such missions, cluttered terrain poses a significant risk to a \ac{UAV} not only because of a potential terrain collision but also because of local meteorological effects: While rain, thunderstorms and global winds are large-scale effects that are predictable using \ac{NWP} systems, local turbulence including updrafts and downdrafts is usually caused by terrain features that are not resolved by the low-resolution \ac{NWP} system. However, unexpected turbulence or downdrafts behind ridges can either damage the \ac{UAV} directly or exceed its climb speed and thereby cause collision with terrain. Especially slow, fragile and low propulsion-power-to-weight vehicles such as solar-powered UAVs (\cref{fig:PL_Intro_Collage}) are susceptible to such weather effects.

Safe operations in such environments require three technologies that do not exist on today's small-scale UAVs: First, a \ac{SLAM} system to create a map of the environment, second, a method to predict the local 3D wind field around the \ac{UAV}, and third, a path planner to calculate an optimal collision-free path with respect to the terrain and 3D wind field. The focus of this paper is on wind field prediction and path planning. Such a system needs to run in real time to react to new terrain information. In addition, to guarantee flight safety at all times, including conditions with a degraded communication link, the system needs to run onboard and be independent from local infrastructure.

%%%%%%%%%%%%%%%%%%%%%%%%%%%%%%%%%%%%%%%%%%%%%%%%%%%%%%%%%%%%%%%%%
%\paragraph{State of the art}
%\label{sec:PL_Intro_StateOfTheArt}
%%%%%%%%%%%%%%%%%%%%%%%%%%%%%%%%%%%%%%%%%%%%%%%%%%%%%%%%%%%%%%%%%

The current literature \emph{does} consider path planning in wind. However, the wind fields are not calculated in real time but are pre-calculated using \ac{NWP} systems on ground-based computers or supercomputers. The \ac{UAV} can therefore neither react to weather forecast changes nor to terrain features that were not considered by the \ac{NWP} system. A comprehensive overview of the current state of the art is given in the respective sections on wind field prediction (\cref{sec:PL_WindPred_StateOfTheArt}) and path planning (\cref{sec:PL_Planning_StateOfTheArt}).

%%%%%%%%%%%%%%%%%%%%%%%%%%%%%%%%%%%%%%%%%%%%%%%%%%%%%%%%%%%%%%%%%
\paragraph{Contributions}
\label{sec:PL_Intro_Contributions}
%%%%%%%%%%%%%%%%%%%%%%%%%%%%%%%%%%%%%%%%%%%%%%%%%%%%%%%%%%%%%%%%%

This paper presents a path planning framework which integrates local 3D wind field predictions to guarantee the safe and efficient operation of fixed-wing Unmanned Aerial Vehicles in cluttered terrain. The framework runs in real time\footnote{We define real time as follows: Assume the UAV is flying and constantly mapping previously unknown terrain at a distance $d$ in front of it with a computer vision node. Both the wind prediction and path planning then need to be fast enough to allow the UAV to react to potential dangers (obstacles, dangerous winds) at distance $d$. For example, if $d=\unit[300]{m}$ and the flight speed is \unitfrac[10]{m}{s}, then calculation times $t\ll\unit[30]{s}$ are considered real time. Real time capability is easier to reach when flying far away from terrain at high altitude where $d$ is also larger.} on the limited computational resources of the UAV's onboard computer. The on-board implementation is chosen to guarantee that the framework can generate safe paths at all times, i.e. also in cluttered terrain and during remote missions where the telecommunications link is degraded. Two central research contributions are presented:
\begin{itemize}
\item \emph{Local 3D wind field prediction:} Our downscaling method receives low-resolution \ac{NWP} data as an input via low-bandwidth data links (e.g. satellite communication) and then leverages potential flow theory to generate a high-resolution 3D wind field close to the \ac{UAV}. The method is assessed in simulations and with LIDAR wind profiles from Swiss Alpine regions.
\item \emph{Real-time wind-aware path planning:} A sampling-based path planner that plans shortest- or time-optimal collision-free paths in the presence of the predicted non-uniform winds is presented. The planner is based on the \ac{OMPL}. We extend this planner to consider the vehicle dynamics through Dubins aircraft primitives and to speed up the planning via obstacle-aware sampling, heuristics for faster nearest neighbor search and fast collision checking on a pre-processed 2.5D map. The calculation of Dubins airplane paths in non-uniform wind is presented.
\end{itemize}
  
This paper is the first in the literature to implement \emph{real-time} 3D wind field predictions \emph{from onboard a \ac{UAV}}. Therefore, we consider this paper initial work on the path towards full onboard weather-awareness and path planning with UAVs. It is not expected that the simple wind field downscaling method can capture all the different flow patterns occuring in complex terrain. From a scientific viewpoint, the paper therefore seeks to assess, first, how accurately our simple method can predict the wind field when running onboard a \ac{UAV} in real time, and second, what path cost advantages and computational demands wind-aware path planning results in.
%F:\\
%always safe
%also in clut. terrain
%even when no LOS or CC connection to operator
%consider airplane dynamics 
%real time
%3d wind fields
%path planning

%SUPER IMPORTANT: STORYLINE: Central assumption: Flight path needs to be safe at all times. Therefore, need to have high-resolution forecasts available at all times, i.e. also in cluttered terrain and during remote missions where the telecommunications link is degraded. This excludes offboard computation because that would require transmitting vasts amount of data (DEM to GCS, and weather data backwards!) => Goal: Develop/investigate onboard 3D wind field prediction method with only limited telecommunication requirements to ground station.

The remainder of this paper is organized as follows. \Cref{sec:PL_WindPrediction} presents the design, implementation and assessment of the 3D wind field prediction method. \Cref{sec:PL_Planning} presents the wind-aware path planning framework and assesses it in synthetic and real-terrain test cases with wind fields generated by our wind downscaling method. \Cref{sec:PL_Conclusion} presents a summary and concluding remarks.

%%%%%%%%%%%%%%%%%%%%%%%%%%%%%%%%%%%%%%%%%%%%%%%%%%%%%%%
%%%%%%%%%%%%%%%%%%%%%%%%%%%%%%%%%%%%%%%%%%%%%%%%%%%%%%%
\section{Prediction of 3D Wind Fields from a UAV}
\label{sec:PL_WindPrediction}
%%%%%%%%%%%%%%%%%%%%%%%%%%%%%%%%%%%%%%%%%%%%%%%%%%%%%%%
%%%%%%%%%%%%%%%%%%%%%%%%%%%%%%%%%%%%%%%%%%%%%%%%%%%%%%%
 
This chapter investigates the feasibility of predicting the local 3D wind field in high resolution and real time on the limited computational resources of small UAVs. Given that large-scale weather models are too computationally expensive, we instead trade-off, implement and assess \emph{downscaling}-based methods. It is assumed that the model has access to low-resolution \acl{NWP} data and terrain information as inputs. Its output is used by the path planner of \cref{sec:PL_Planning}.

%%%%%%%%%%%%%%%%%%%%%%%%%%%%%%%%%%%%%%%%%%%%%%%%%%%%%%%%%%%%%%%%%
\subsection{State of the Art}
\label{sec:PL_WindPred_StateOfTheArt}
%%%%%%%%%%%%%%%%%%%%%%%%%%%%%%%%%%%%%%%%%%%%%%%%%%%%%%%%%%%%%%%%%

%\paragraph{Prediction of wind fields from UAVs}

The current \ac{UAV} literature does not discuss UAV-based wind predictions, but only \emph{in-situ} wind measurements: Filtering approaches that fuse the information of an \ac{IMU}, \ac{GPS} and airspeed sensor are mostly used~\cite{Petrich2011WindEstimation,Langelaan2011WindEstimation}. Approaches that estimate the wind vector without an airspeed sensor also exist~\cite{palanthandalam2008WindEstimationWithoutAirspeedSensor}. In addition, formations of UAVs have been used to cooperatively measure wind fields~\cite{Larrabee2014WindEstimationCooperative}. To find suitable methods for the \emph{online prediction} of 3D wind \emph{fields} from UAVs, \emph{downscaling} methods that historically come from the fields of meteorology and flow simulation are therefore investigated. In general, these downscaling methods are of three different types:
%\paragraph{Downscaling Methods}
%In general, downscaling methods for meteorological data are of three different types:
\begin{itemize}
\item \emph{Statistical methods} use previously measured wind data in a certain region and thereof derive wind expectations for the same region using statistical methods. Given that we aim to predict the wind over unvisited terrain, these methods are not suitable for our application.
\item \emph{Prognostic models} calculate the approximate solution of differential equations describing the \ac{ABL}. They can reproduce small-scale weather effects, but are computationally expensive. ARPS, the Advanced Regional Prediction System~\cite{Xue2000, Xue2001}, has been applied to complex terrain \cite{Mott2010, Mott2010b, Hug2005, Raderschall2008} and is freely available\footnote{\url{http://www.caps.ou.edu/ARPS/arpsdown.html}}. GRAMM, the Graz Mesoscale Model~\cite{GRAMM}, considers mass, momentum, potential temperature and humidity conservation and is often used together with the dispersion model GRAL (Graz Lagrangian Model), e.g. in \citet{Oettl2015}. It is free\footnote{\url{http://lampx.tugraz.at/~gral/index.php/download}} and provides a GUI.
\item \emph{Diagnostic methods} do not have a proper physical background but return a physically consistent wind field with respect to mass or momentum conservation. They are less computationally demanding than prognostic models but cannot predict local effects such as slope winds \cite{Ratto1994} or the time-dependency of the flow. MATHEW~\cite{Sherman1978a, Sherman1978b} is a diagnostic mass consistent model which adds a potential field to an initial wind field to make it divergence free. The source code of a modified model is available~\cite{Walt2016}. CALMET~\cite{CALMET} additionally models kinematic terrain effects, slope flows and blocking effects and can incorporate observational data at the cost of increased computational cost. The usage of CALMET coupled to NWP data showed a significant improvement with respect to NWP outputs \cite{Truhetz2007, Morales2012}. Software licences are available for free\footnote{\url{http://www.src.com/calpuff/calpuff_eula.htm}}. WindNinja~\cite{Forthofer2014} provides two different models: A mass-conserving model which resembles MATHEW, and a mass- and momentum-conserving model that uses commercial computational fluid dynamics software. Near surface winds are improved especially for high-wind events~\cite{Wagenbrenner2016}. The software with GUI is available for free\footnote{\url{https://www.firelab.org/document/windninja-software}}.
\end{itemize}

%%%%%%%%%%%%%%%%%%%%%%%%%%%%%%%%%%%%%%%%%%%%%%%%%%%%%%%%%%%%%%%%%
\subsection{Method Selection}
\label{sec:PL_WindPred_MethodSelection}
%%%%%%%%%%%%%%%%%%%%%%%%%%%%%%%%%%%%%%%%%%%%%%%%%%%%%%%%%%%%%%%%%

\paragraph{Requirements}

Based on the overall goal postulated in \cref{sec:PL_Introduction}, the following specific requirements are derived for the wind prediction framework: It shall

\begin{itemize}
\item be able to calculate a sufficiently accurate 3D wind field for an area of $\unit[1]{km^3}$ (with the \ac{UAV} at its center) and \unit[25]{m} grid resolution in less than \unit[30]{s} calculation time on the UAV's onboard computer\footnote{Such as an INTEL UP Board, \url{http://www.up-board.org/up/}}, which we assume corresponds to \unit[10]{s} on a standard laptop computer. 
\item have a robust and fully automated execution that does not require any human intervention.
\item be usable on UAVs operating remotely or in cluttered terrain where the communication bandwidth to the ground station is limited.
\item be modular and easily modifiable.
\end{itemize}

\paragraph{Model selection}

Five different models were traded off against the requirements presented above. First, both ARPS and GRAMM/GRAL were discarded because they violate the computation time constraint, i.e. usually require far more time to guarantee that a numerically stable solution is reached. The mass- and momentum-conserving version of WindNinja is discarded due to computation time disadvantages, too. Of the remaining models, CALMET is considered slower and has a less modular code base. The two remaining models are MATHEW and the mass-conserving WindNinja, which essentially solve the same mathematical problem. Finally, MATHEW is selected because it has a slim and modular source code and can, as shown by Walt~\cite{Walt2016}, solve similar meteorological downscaling problems within the calculation time requirements. In addition, it solves the internal \ac{FEM} using FEniCS, an open-source library that already provides many different solvers and preconditioners, has both Python and C++ bindings, and supports various grid types.

%%%%%%%%%%%%%%%%%%%%%%%%%%%%%%%%%%%%%%%%%%%%%%%%%%%%%%%%%%%%%%%%%
\subsection{Fundamentals}
\label{sec:PL_WindPred_Fundamentals}
%%%%%%%%%%%%%%%%%%%%%%%%%%%%%%%%%%%%%%%%%%%%%%%%%%%%%%%%%%%%%%%%%
 
%+? In contrast to \citet{Sherman1978a}, Walt uses \gls{FEM} instead of \gls{FDM}. This has the main advantage that no transformations are required when working with a terrain-following grid with variable cell size. \\

\paragraph{Mathematical derivation}
\label{sec:PL_WindPred_mathematical_derivation}

\begin{figure}[b]
	\centering
	\includegraphics[width=.4\columnwidth]{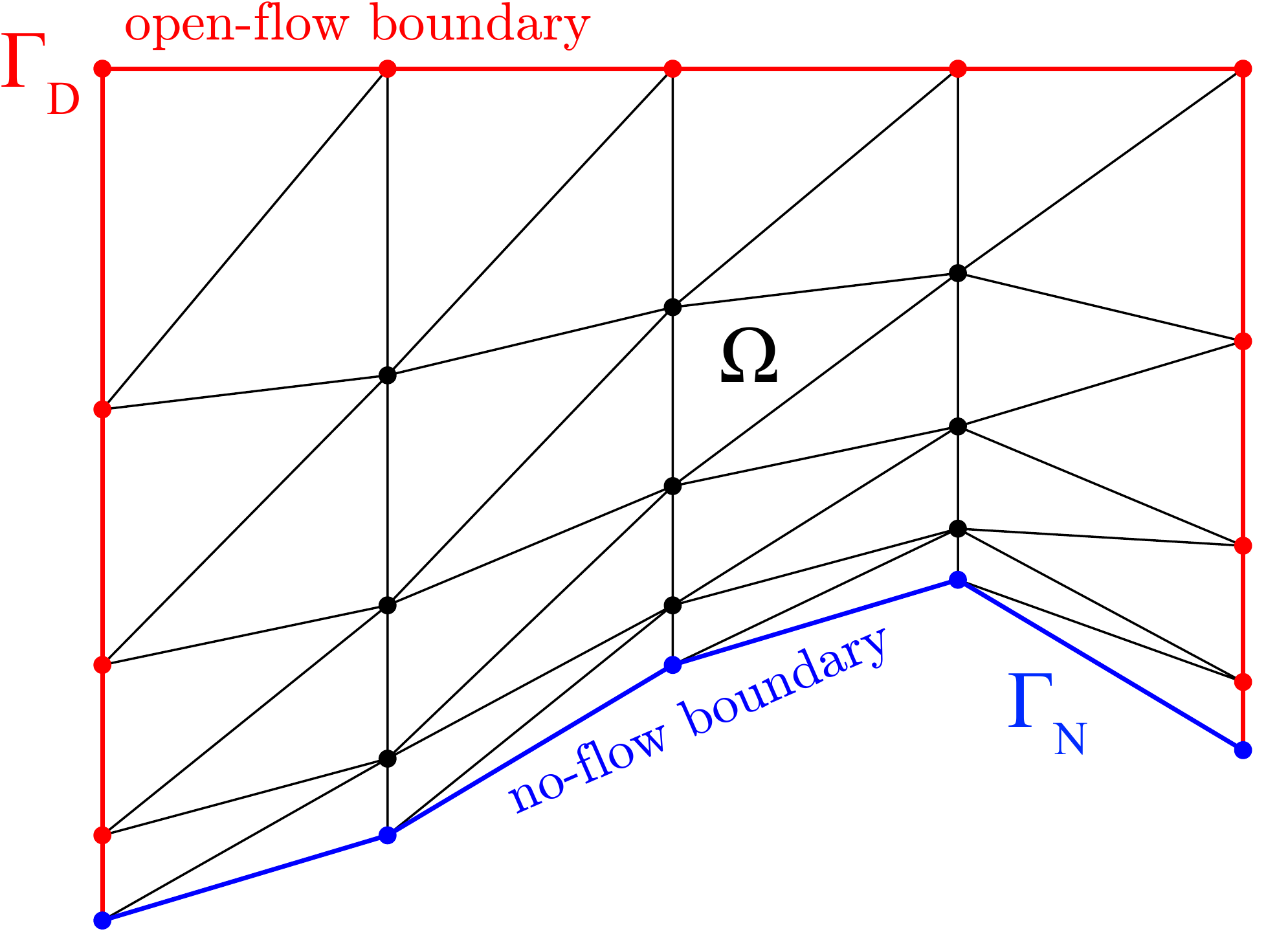}
	\caption[2D grid]{Computational domain and boundaries of MATHEW on 2D grid: The red lines and dots represent the open-flow boundary whereas blue is used for the terrain. The black dots are the grid points within the domain $\Omega$ and the black lines illustrate the cells.}
	\label{fig:PL_WindPred_Fund_Grid2D}
	\end{figure}

MATHEW calculates the divergence-free wind field $\vec{u}$ from an initial wind field $\vec{u}^\text{I}$ (retrieved from a NWP system), a \ac{DEM} and the stability matrix $\matr{S}$. The problem domain $\Omega$ is an open set in 3-dimensional space and its boundaries are the Dirichlet boundary $\Gamma_\text{D}$, which covers the open-flow boundary, and the Neumann boundary $\Gamma_\text{N}$ defining the terrain (\cref{fig:PL_WindPred_Fund_Grid2D}). 
The central \emph{hard} constraint enforced by MATHEW is that the resulting $\vec{u}$ needs to be divergence free over the domain $\Omega$, i.e.
	\begin{equation}
	\nabla \cdot \vec{u} = 0\; .
	\label{eqn:PL_WindPred_wind_div_free}
	\end{equation}
This essentially means that the flow is incompressible. The main \emph{soft} constraint is that the adjustments between the initial wind field $\vec{u}^\text{I}$ and the final wind field $\vec{u}$ shall be minimal. This expresses the assumption that the initial wind field observation $\vec{u}^\text{I}$ contains valuable information collected through sensing or a prior model --- and is thus \emph{to some degree} correct (see \cref{sec:PL_WindPred_PrelResults_Synthetic} for a violation of that assumption). Mathematically, MATHEW reduces the accumulated error between $\vec{u}$ and $\vec{u}^\text{I}$ over the domain in a least-squares sense by minimizing the functional
	\begin{equation}
	\pazocal{L}(\vec{u}; \lambda) = \int\limits_{\Omega} \left( \frac{1}{2} \matr{S} \left( \vec{u} - \vec{u}^\text{I} \right) \cdot \left( \vec{u} - \vec{u}^\text{I} \right) + \lambda \nabla \cdot \vec{u} \right)dV \; , 
	\label{eqn:PL_WindPred_functional}
	\end{equation}
with velocity potential $\lambda$ and stability matrix $\matr{S}$ (explained in the following subsection). The Euler-Lagrange equations for vector functions 
\begin{equation}
\frac{\partial\pazocal{L}}{\partial f_i}-\frac{d}{dx} \Big( \frac{\partial\pazocal{L}}{\partial f_i^\prime}\Big) = 0
\end{equation}
are applied to \cref{eqn:PL_WindPred_functional} in order to minimize the functional. The problem definition now has the form 
\begin{equation}
	\vec{u} = \vec{u}^\text{I} + \matr{S}^{-1} \nabla \lambda \; .
	\label{eqn:PL_WindPred_wind_pred}
\end{equation}
Note that while the added potential flow field $S^{-1}\nabla\lambda$ in \cref{eqn:PL_WindPred_wind_pred} is vorticity-free, the final wind field $\vec{u}$ can still contain vorticity (for example recirculation behind a ridge) if it is already present in $\vec{u}^\text{I}$. Two boundary conditions are now applied: The no-flow through terrain assumption, which translates to $\vec{u}\cdot\vec{n}=0$ where $\vec{n}$ is the normal vector at the terrain, and the assumption of no adjustment of the normal velocities at the open-flow boundaries, which translates to $\lambda=0$. The resulting system of equations is
	\begin{equation}
	\begin{aligned}
	\vec{u} &= \vec{u}^\text{I} + \matr{S}^{-1} \nabla \lambda && \text{in } \Omega \\
	\nabla \cdot \vec{u} &= 0 && \text{in } \Omega \\
	\lambda &= 0 && \text{on } \Gamma_\text{D} \\
	\vec{u} \cdot \vec{n} &= 0 && \text{on } \Gamma_\text{N} \; .
	\label{eqn:PL_WindPred_problem_orig}
	\end{aligned}
	\end{equation}
These equations can be expressed as a \ac{PDE} for the velocity potential $\lambda$ by taking the scalar product with the gradient operator on both sides of the equations of \cref{eqn:PL_WindPred_problem_orig}:
	\begin{equation}
	\begin{aligned}
	-\nabla \cdot \left(\matr{S}^{-1} \nabla \lambda\right) & = \nabla \cdot \vec{u}^\text{I} && \text{in } \Omega \\
	\lambda &= 0 && \text{on } \Gamma_\text{D} \\
	\left(-\matr{S}^{-1} \nabla \lambda\right) \cdot \vec{n} &= \vec{u}^\text{I} \cdot \vec{n} && \text{on } \Gamma_\text{N} \; .
	\label{eqn:PL_WindPred_problem_pdf}
	\end{aligned}
	\end{equation}
\Cref{eqn:PL_WindPred_problem_pdf} is an elliptic \ac{PDE}. For the transformation into a \ac{FEM} problem, the first line in \cref{eqn:PL_WindPred_problem_pdf} needs to be post-multiplied with a test function $\pazocal{F}$ satisfying its boundary condition ($\pazocal{F} = 0$ on $\Gamma_\text{D}$) and integrated over the problem domain $\Omega$:
	\begin{align}
	- \int\limits_{\Omega} \nabla \cdot \left(\matr{S}^{-1} \nabla \lambda\right) \pazocal{F} \, dV = \int\limits_{\Omega} \left(\nabla \cdot \vec{u}^\text{I} \right) \pazocal{F} \, dV
	\label{eqn:PL_WindPred_pde_trial}
	\end{align}
With the use of Green's Theorem the left-hand side of \cref{eqn:PL_WindPred_pde_trial} evolves to:
	\begin{equation}
	\begin{aligned}
	- \int\limits_{\Omega} \nabla \cdot \left(\matr{S}^{-1} \nabla \lambda\right) \pazocal{F} \, dV &= \int\limits_{\Omega} \left( \matr{S}^{-1} \nabla \lambda \right) \cdot \nabla \pazocal{F} \, dV + \int\limits_{\Gamma_\text{D} \cup \Gamma_\text{N}} \underbrace{\left( \left( - \matr{S}^{-1} \nabla \lambda \right) \cdot \vec{n} \right)}_{= \vec{u}^\text{I} \cdot \vec{n}} \pazocal{F} \, dS \\
	&= \int\limits_{\Omega} \left( \matr{S}^{-1} \nabla \lambda \right) \cdot \nabla \pazocal{F} \, dV + \int\limits_{\Gamma_\text{N}} \left( \vec{u}^\text{I} \cdot \vec{n} \right) \pazocal{F} \, dS \; . 	\label{eqn:PL_WindPred_pde_trial_green}
	\end{aligned}
	\end{equation}
When plugging the result of \cref{eqn:PL_WindPred_pde_trial_green} into \cref{eqn:PL_WindPred_pde_trial} and rearranging the terms, the so called weak formulation of the problem is derived:
\begin{equation}
	\underbrace{\int\limits_{\Omega} \left( \matr{S}^{-1} \nabla \lambda \right) \cdot \nabla \pazocal{F} \, dV}_{a(\lambda,\pazocal{F})} = \underbrace{\int\limits_{\Omega} \left( \nabla \cdot \vec{u}^\text{I} \right) \pazocal{F} \, dV - \int\limits_{\Gamma_\text{N}} \left( \vec{u}^\text{I} \cdot \vec{n} \right) \pazocal{F} \, dS}_{l(\pazocal{F})} \; ,
	\label{eqn:PL_WindPred_weak_form}
	\end{equation}
with bilinear form $a(\lambda,\pazocal{F})$ and linear form $l(\pazocal{F})$. The procedure is to now solve for the velocity potential $\lambda$ using the Ritz-Galerkin discretization (for details see Walt~\cite{Walt2016}), and to finally re-insert $\lambda$ into \cref{eqn:PL_WindPred_wind_pred} to retrieve the final 3D wind field $\vec{u}$.

\paragraph{Stability matrix}
The stability matrix is defined as
	\begin{equation}
	\matr{S} = \begin{pmatrix}
	\alpha_\text{h}^2 & 0 & 0 \\ 0 & \alpha_\text{h}^2 & 0 \\ 0 & 0 & \alpha_\text{v}^2
	\end{pmatrix} \; ,
	\label{eqn:PL_WindPred_stab_mat}
	\end{equation}
where $\alpha_\text{h}$ and $\alpha_\text{v}$ are weights for the horizontal and vertical component. In contrast to the work by Sherman~\cite{Sherman1978a}, they are herein defined as $\alpha_i^2=1/\sigma_i^2$\note{TBC}, where $\sigma_i$ are the observation errors or deviations of the observed and desired wind field. The weights $\alpha_\text{h}$ and $\alpha_\text{v}$ are used to define the stability parameter $\alpha$ as 
	\begin{equation}
	\alpha = \frac{\alpha_\text{h}^2}{\alpha_\text{v}^2} = \left( \frac{w}{u} \right)^2 \; ,
	\end{equation}
where $u$ and $w$ are the magnitudes of the expected horizontal and vertical wind speeds, respectively. Inserting $\alpha$ into the definition of the stability matrix, its inverse becomes
	\begin{equation}
	\matr{S}^{-1} = c\cdot\begin{pmatrix}
	1 & 0 & 0 \\ 0 & 1 & 0 \\ 0 & 0 & \alpha
	\end{pmatrix} \; .
	\end{equation}
Note that the constant factor $c$ can be and is neglected ($c=1$) in \cite{Walt2016,Mueller_MT_WindPrediction} because it, first, does not have an effect on the final wind field $\vec{u}$ as during the solution of \cref{eqn:PL_WindPred_weak_form} $\lambda$ would automatically be scaled by 1/c, and second, it cannot be used to adjust the atmospheric stability. This is however possible using the stability parameter $\alpha$. Sherman~\cite{Sherman1978a} assesses multiple test cases and determines a typical value of $\alpha \approx 10^{-4}$. A larger $\alpha$ implies that the wind field is mainly corrected in vertical direction, a smaller one that the correction is principally applied to the horizontal direction.

%%%%%%%%%%%%%%%%%%%%%%%%%%%%%%%%%%%%%%%%%%%%%%%%%%%%%%%%%%%%%%%%%
\subsection{Implementation}
\label{sec:PL_WindPred_Implementation}
%%%%%%%%%%%%%%%%%%%%%%%%%%%%%%%%%%%%%%%%%%%%%%%%%%%%%%%%%%%%%%%%%

\paragraph{Architecture}

\begin{figure}[htb]
\centering
\tikzstyle{base} = [draw, text width=5em, minimum height=3.5em]
\tikzstyle{process} = [base, rectangle, text centered, text width=16em, minimum height=3em, fill=blue!15]
\tikzstyle{extprocess} = [base, rectangle, text centered, text width=16em, minimum height=3em, fill=green!15]
\tikzstyle{inout} = [base, trapezium, trapezium left angle=70, trapezium right angle=-70, text badly centered, trapezium stretches=true, fill=orange!15]
\tikzstyle{line} = [draw, -latex']

\begin{tikzpicture}[node distance=1.2cm, scale=0.75, every node/.style={transform shape}]
\node [inout] (nwp) {NWP \\ data};
\node [inout, left of=nwp, node distance=2.2cm] (dem) {DEM \\ data};
\node [inout, right of=nwp, node distance=2.2cm] (simparam) {simulation parameters};

\coordinate [below of=nwp, node distance=0.8cm] (inputs);

\node [extprocess, below of=inputs, node distance=0.8cm] (updated) {updated?};
\coordinate [right of=updated, node distance=3.5cm] (updatedDecision);
\coordinate [below of=updatedDecision,node distance=1cm] (updatedDecisionNo1);
\coordinate [below of=updated,node distance=1cm] (updatedDecisionNo2);

\node [process, right of=updatedDecision, node distance=3.5cm] (init) {initialize model with DEM, NWP and simulation parameters};
\node [process, below of=init] (grid) {create grid and interpolate NWP data to its nodes};
\node [process, below of=grid] (fem_setup) {set up weak formulation of FEM problem \eqref{eqn:PL_WindPred_weak_form}};
\node [process, below of=fem_setup] (vel_pot) {solve for velocity potential $\lambda$};
\node [process, below of=vel_pot] (pred) {add potential field to initial wind field \eqref{eqn:PL_WindPred_wind_pred}};
\node [inout, left of=pred, node distance=7cm] (windfield) {adjusted wind field};
\node [extprocess, below of=windfield,node distance=1.4cm] (planner) {flight planner};

\path [line] (dem) |- (inputs) -- (updated);
\path [line] (nwp) -- (updated);
\path [line] (simparam) |- (inputs) -- (updated);
\path [draw] (updated) -- (updatedDecision);
\path [line] (updatedDecision) |- (updatedDecisionNo2) -- (updated);
\path (updatedDecision) edge node[right]{No} (updatedDecisionNo1);
\path (updatedDecisionNo1) edge node [below,color=gray]{0.5 Hz} (updatedDecisionNo2);

\path (updatedDecision) edge[-latex'] node[above]{Yes} (init);
\path [line] (init) -- (grid);
\path [line] (grid) -- (fem_setup);
\path [line] (fem_setup) -- (vel_pot);
\path [line] (vel_pot) -- (pred);
\path [line] (pred) -- (windfield);
\path [line] (windfield) -- (planner);

%Legend %0.35
\node (legend_data) at ( 9.06,-7.42) [text width=2em, minimum height=1em, fill=orange!15,label=left:Data] {};
\node (legend_process) [text width=2em, minimum height=1em, fill=blue!15,below of=legend_data, node distance=0.35cm, label=left:Wind model calculation] {};
\node (legend_extprocess) [text width=2em, minimum height=1em, fill=green!15,below of=legend_process, node distance=0.35cm, label=left:External/administrative process] {};

\end{tikzpicture}

%\path [line] (updatedDecisionNo2) -- (updated);
%\path (updatedDecision) | (updated);
%\path [line] (updatedDecision) (init);
%edge["Yes"']
%\path [line] (updatedDecision) -- (init);
\caption[Architecture and interfaces]{Wind downscaling architecture: Input data changes trigger the wind field downscaling calculation, which then publishes its outputs to \ac{ROS} and the flight planner.}
\label{fig:PL_WindPred_Impl_Architecture}
\end{figure}
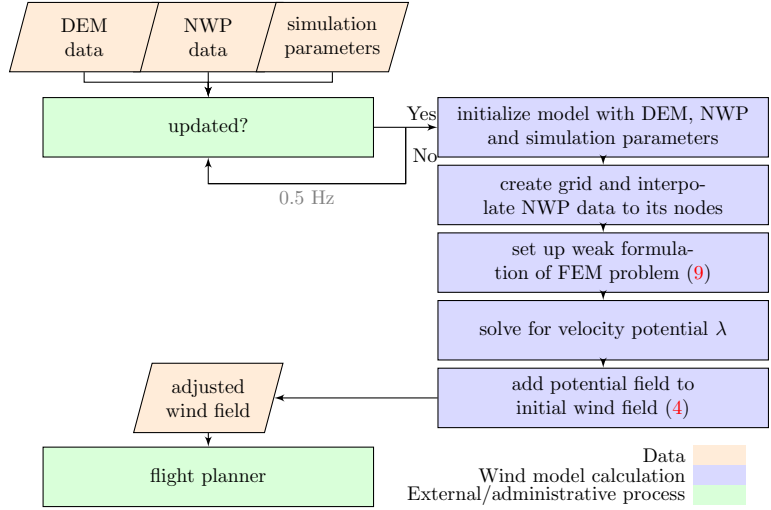

\Cref{fig:PL_WindPred_Impl_Architecture} presents the data sources and steps required by our implementation of MATHEW. The framework is implemented in Python with \ac{ROS} support. The input data are the simulation parameters, a \ac{DEM} and coarse \ac{NWP} data (wind profiles around and within the region of interest). The input data is monitored at \unit[0.5]{Hz} and the model is initialized when the inputs have changed. The grid is generated and the initial wind field $\vec{u}^\text{I}$ is calculated by linearly interpolating the coarse \ac{NWP} data to the vertical layer heights of the high-resolution grid. Afterwards bilinear interpolation is used within each vertical layer to calculate the wind components at each grid point. The mathematical operations described in \cref{sec:PL_WindPred_mathematical_derivation} are then executed to solve for the velocity potential $\lambda$, which allows to calculate the adjusted wind field $\vec{u}$ by using \cref{eqn:PL_WindPred_wind_pred}. This final wind field is published to \ac{ROS} nodes such as the flight planner (\cref{sec:PL_Planning}).

The decision to implement the architecture of \cref{fig:PL_WindPred_Impl_Architecture} onboard a \ac{UAV} instead of a ground-based computer is motivated by the requirement to have wind field forecasts available at all times, i.e. even when flying remotely with reduced communication bandwidth to the ground station. The transmitted data volume therefore needs to be minimal. However, even the $\unit[1]{km^3}$ high-resolution wind field defined in \cref{sec:PL_WindPred_MethodSelection} contains 1681 horizontal grid points with multiple vertical layers. Calculating this wind field on a ground-based computer and transmitting the results via low-speed telemetry links (for example degraded line-of-sight based telemetry or satellite communication) is not possible in real-time. Therefore, an onboard implementation  is chosen. The NWP data is provided by the European \ac{COSMO} model~\cite{COSMOMeteoSwiss} at \unit[1--7]{km} horizontal resolution and up to 80 vertical layers. A ground-based client computer only extracts the NWP data at the specific time and region of interest. Assuming COSMO-1 \ac{NWP} data, the $\unit[1]{km^3}$ region of interest then consists of only 9 horizontal grid points with different vertical levels that need to be transmitted via telemetry. Even low-bandwidth links such as the satellite communication link on \emph{AtlantikSolar}~\cite{Oettershagen_JFR2017} can manage such a bandwidth. The \ac{DEM} data comes from an octomap~\cite{Hornung2013Octomap} that is either preloaded or generated directly onboard the \ac{UAV} by a vision-based \ac{SLAM} node~\cite{Hinzmann2016}. %The framework can therefore provide wind predictions when they are most crucial: When operating remotely, in cluttered terrain or without local communications infrastructure.

\paragraph{Parameter optimization}

\begin{figure}[bth]
\centering
\includegraphics[width=.62\columnwidth]{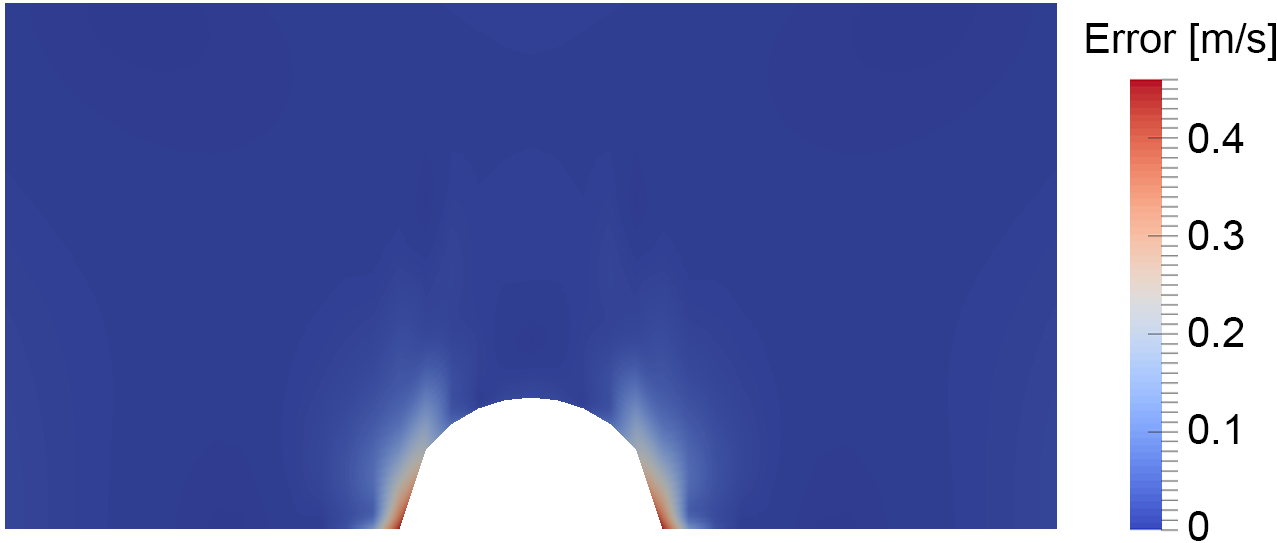}
\caption[Hemisphere error]{Non-weighted local errors, i.e. the magnitude of the difference between the wind field vectors predicted by the model and the analytical solution, for the hemisphere. The linear vertical spacing, ILU preconditioner and CG solver are used. The largest errors occur in a small region with abrupt terrain change at the bottom of the hemisphere.}
\label{fig:PL_WindPred_Impl_HemiError}
\end{figure}

The framework's accuracy and required computation time depend on three critical parameters: The vertical grid point spacing and the preconditioner and solver used by FEniCS. These are optimized by comparing the model to the analytical results for flow around a hemisphere with radius \unit[0.25]{m} (\cref{fig:PL_WindPred_Impl_HemiError}). It is placed at the origin of the domain $[-1, -1, 0] \times [1, 1, 1]$ (in m) with \unitfrac[1]{m}{s} inflow in positive $x$-direction. $N_x=N_y=41$ horizontal grid points are used. The stability parameter is $\alpha=1$. While the horizontal spacing is constant, the vertical spacing is not: The vertical position $z$ of a grid point at vertical index $n$ and position $x,y$ is defined as
\begin{equation}
z(x, y, n) = h(x, y) + \big(t(x, y) - h(x, y) \big) \cdot \gamma(n) \; ,
\end{equation}
where $h(x, y)$ is the terrain height and $t(x, y)$ is the top vertical domain boundary at the $x,y$ position. The discrete function $\gamma(n)$ needs to be strictly monotonically increasing and fulfill $\gamma(0) = 0$ and $\gamma(N_z) = 1$, where $N_z$ is the number of vertical grid points. Equidistant, square-root, linear and squared spacing functions $\gamma(n)$ are tested. The error at a location is defined as the length of the difference vector between the model output and the analytic solution. The weighed total error sums up the product of each local error with the ratio between the height of the respective cell and the average cell height of the entire domain. The median weighted error for all vertical spacings is nearly the same, i.e. about \unitfrac[0.005]{m}{s}. For the linear vertical spacing, the weighted errors show a low standard deviation and a low maximum weighted error of only \unitfrac[0.14]{m}{s}. The linear vertical spacing is therefore selected. 

The elliptic \ac{PDE} of \cref{eqn:PL_WindPred_problem_pdf} was previously solved~\cite{Walt2016} using a direct solver. In order to fulfill the stringent computation time requirements, different FEniCS preconditioners and iterative solvers were assessed using the aforementioned hemisphere test case. The best combination is found to be the FEniCS Incomplete LU-decomposition (ILU) preconditioner with the Conjugate Gradient (CG) solver. Averaged over 10 runs, this combination reduces the computation time for $\lambda$ from \unit[81.8]{s} for the direct solver to only \unit[6.3]{s}. The calculation time requirements of \cref{sec:PL_WindPred_MethodSelection} are thus fulfilled. The additional errors introduced into the wind field have a norm of less than $\unitfrac[10^{-4}]{m}{s}$ and are deemed acceptable given that a time reduction of \unit[92]{\%} is achieved. \Cref{fig:PL_WindPred_Impl_HemiError} shows the overall non-weighted errors of the framework. More detailed information on the implementation and parameter optimization is presented by Müller~\cite{Mueller_MT_WindPrediction}.

 %%%%%%%%%%%%%%%%%%%%%%%%%%%%%%%%%%%%%%%%%%%%%%%%%%%%%%%%%%%%%%%%%
\subsection{Results}
\label{sec:PL_WindPred_PreliminaryResults}
%%%%%%%%%%%%%%%%%%%%%%%%%%%%%%%%%%%%%%%%%%%%%%%%%%%%%%%%%%%%%%%%%

%%%%%%%%%%%%%%%%%%%%%%%%%%%%%%%%%%%%%%%%%%%%%%%%%%%%%%%%%%%%%%%%%
\subsubsection{Synthetic Test Cases}
\label{sec:PL_WindPred_PrelResults_Synthetic}
%%%%%%%%%%%%%%%%%%%%%%%%%%%%%%%%%%%%%%%%%%%%%%%%%%%%%%%%%%%%%%%%%

\paragraph{Steep and narrow valley}
\label{sec:PL_WindPred_valley}

\begin{figure}[htbp]
\centering
\subfloat{\includegraphics[height=6.5cm]{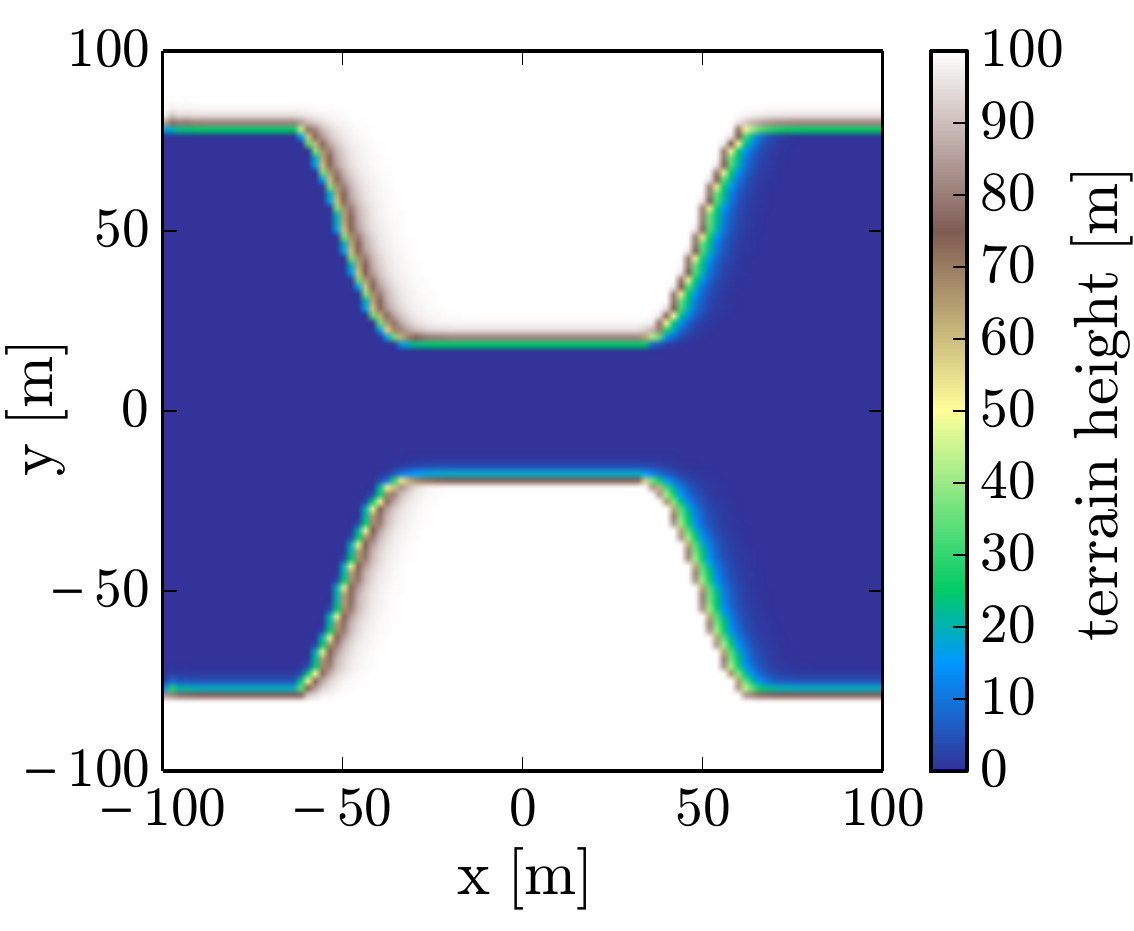}\label{fig:PL_WindPred_terrain_valley}}
\hspace*{\fill} % separation between the subfigures
\subfloat{\includegraphics[height=6.5cm]{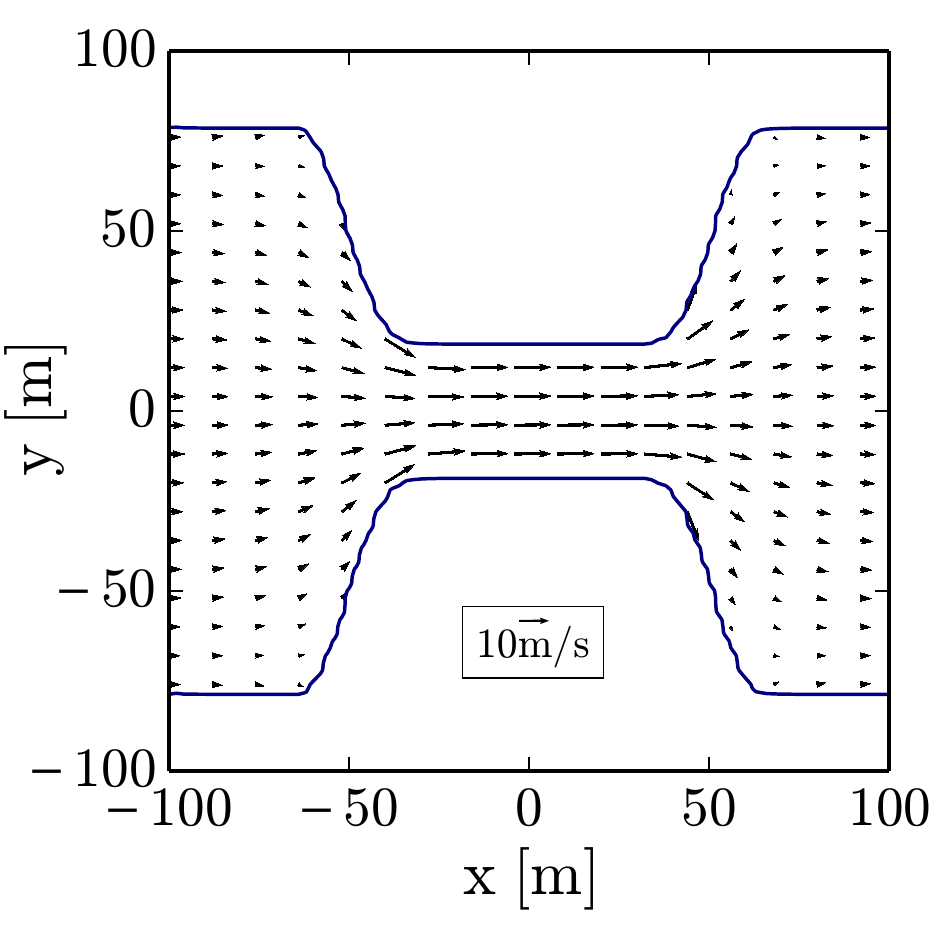}
\label{fig:PL_WindPred_hor_flow_valley}}
\caption{Synthetic valley. Left: Terrain. Right: Adjusted flow (black arrows) at {$z=\unit[40]{m}$} and $\alpha = 1.00$. As expected, the wind speed in the valley is roughly four times 
as high as at the in and outflow.
}
\label{fig:PL_WindPred_valley_testcase}
\end{figure}

\begin{figure}[htbp]
\centering
\subfloat{\includegraphics[height=5.70cm]{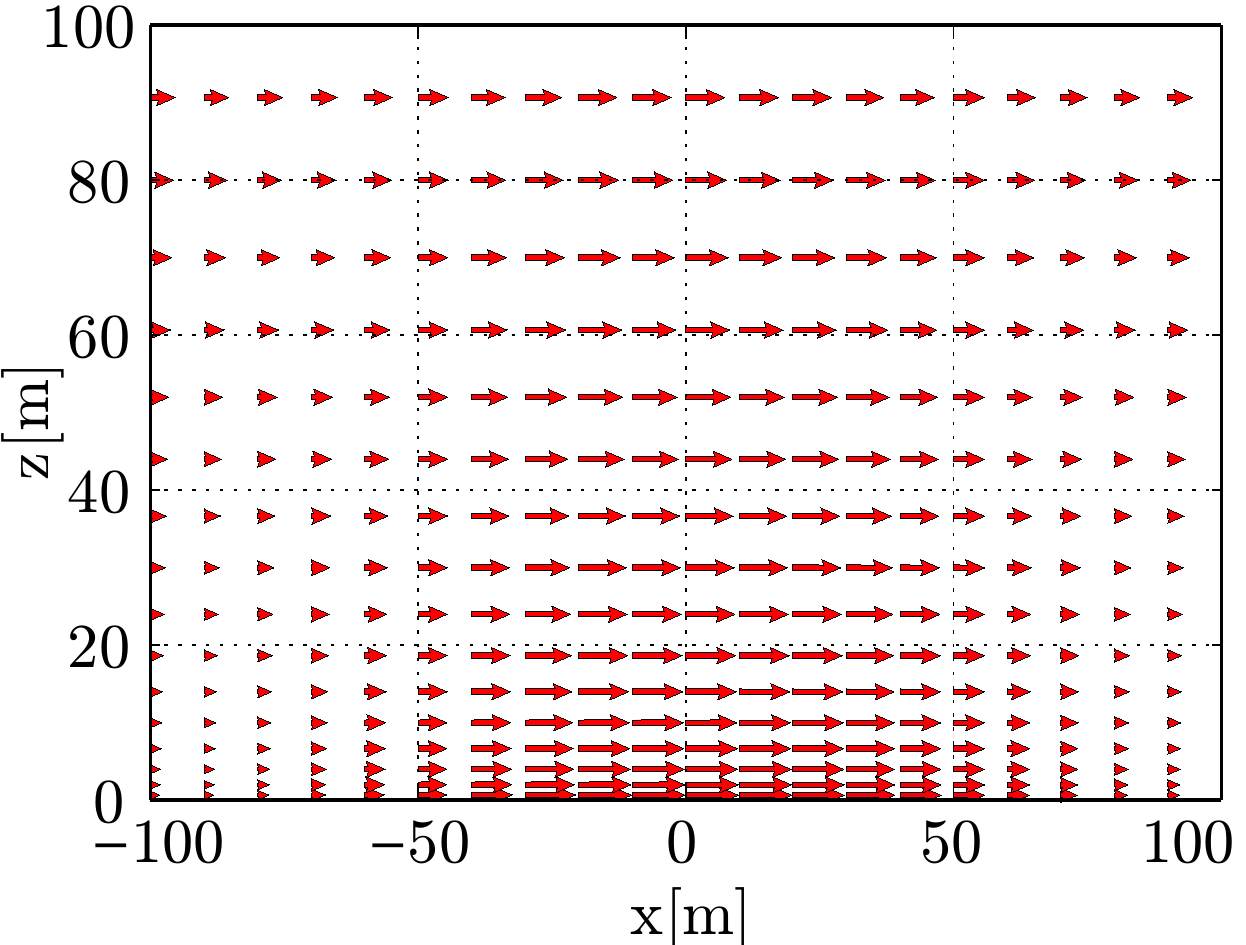}}
\hspace*{\fill} % separation between the subfigures
\subfloat{\includegraphics[height=5.70cm]{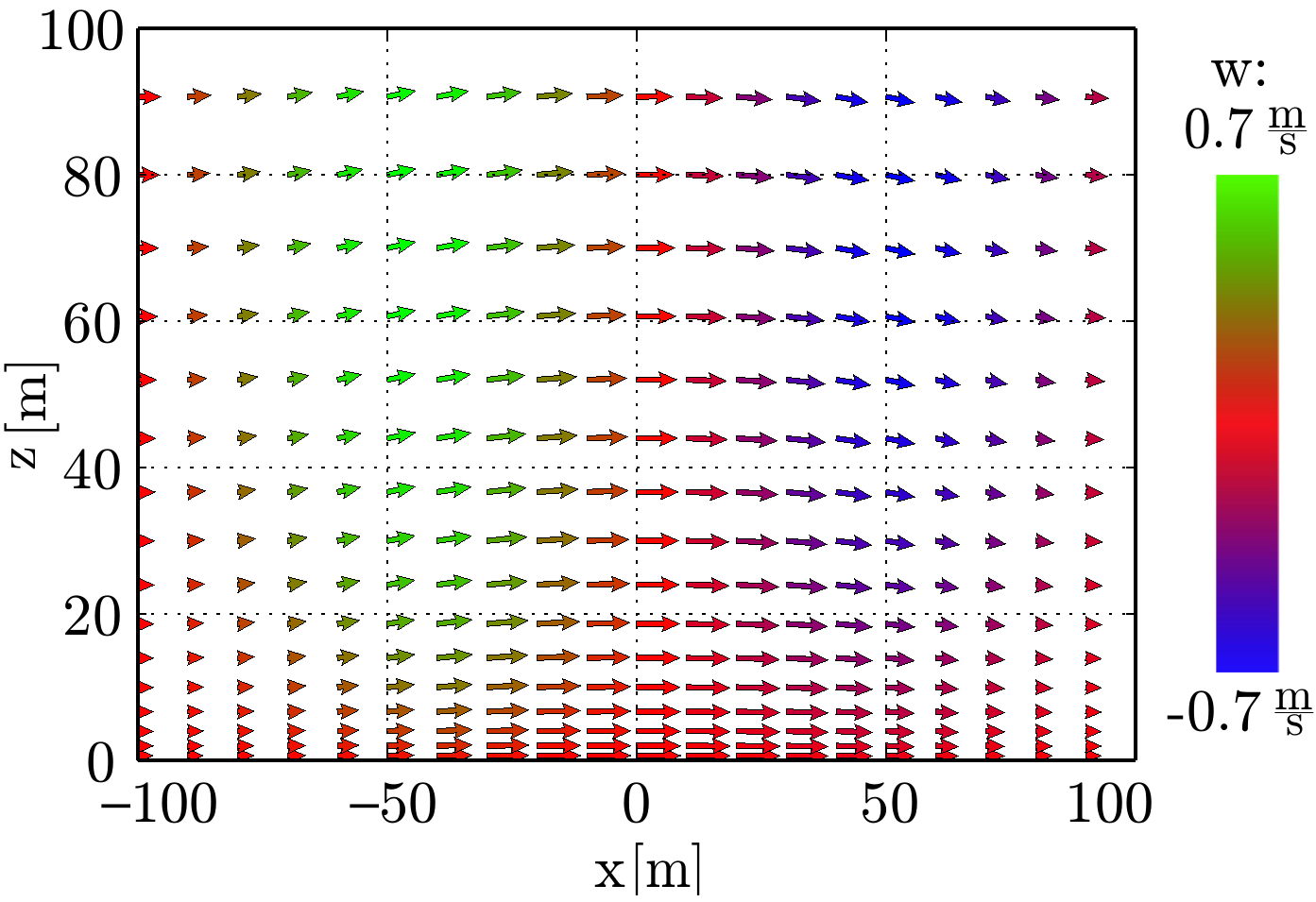}}
%width=.485\textwidth
\caption{Adjusted wind fields $\vec{u}$ for $\alpha = 0.01$ (left) and $\alpha = 1.00$ (right) for the synthetic valley test case in the $x$-$z$ cross section at $y = 0$. The arrow color represents the vertical component $w$. As expected, the wind speed is higher inside the valley. The overall wind magnitude is decreased in both cases. For $\alpha = 1.00$, the flow is also adjusted vertically.}
%Note: For the final journal paper, take out at least subfig1, probably also even subfig2, and maybe even this whole figure!
\label{fig:PL_wind_fields_valley}
\end{figure}

The steep and narrow valley (\cref{fig:PL_WindPred_valley_testcase}) tests how mass conservation is achieved for different stability parameters. At the center of the valley ($x = 0$, $y = 0$) the width is one fourth of the inflow width. The initial wind field $\vec{u}^\text{I}=[5,0,0]\unitfrac[]{m}{s}$ is constant over the entire domain. While $\vec{u}^\text{I}$ is divergence free over $\Omega$, it violates the terrain boundary conditions. The model output $\vec{u}$ is shown in \cref{fig:PL_WindPred_hor_flow_valley,fig:PL_wind_fields_valley}. The ratio of the average inflow speed to the average valley wind speed is roughly 1/4, thus mass conservation is fulfilled. However, the average wind speed in the domain is reduced (\cref{tab:PL_WindPred_wind_valley}) and the expected \unitfrac[20]{m}{s} within the valley are not reached. For $\alpha=1.00$, the flow is also able to avoid the orifice represented by the valley by rising in front of the valley and sinking again afterwards. Given the open flow boundary above the valley this behavior is expected. Overall, the test case shows that $\vec{u}$ is made divergence free, terrain following and mass consistent. However, it also shows a major drawback of the model: As indicated in \cref{sec:PL_WindPred_mathematical_derivation}, the initial wind field $\vec{u}^\text{I}$ is assumed to be a rather close representation of the final flow field. In this case $\vec{u}^\text{I}$ inside the valley was however too low by a factor of four. The model does not have a physically-intuitive way to deal with this (e.g. inflow boundary conditions that constrain the problem). Instead it strictly applies \cref{eqn:PL_WindPred_functional} to minimize the total least squares error between $\vec{u}$ and $\vec{u}^\text{I}$, which is smaller if $\vec{u}$ is decreased in the inflow than if it is kept constant at the inflow but significantly increased inside the valley. 
%It is remarkable that the wind magnitude grows with increasing heights at the in- and outflow, while the flow in the valley has its highest speeds near to the ground. Yes, this is true, but WHY is this the case? If we cannot explain it, we should not mention it?!

\begin{table}[!h]
\caption{Characteristic wind speeds for the valley test and two stability parameters $\alpha$.}
\centering
\begin{tabular}{l r r}
\toprule
Stability parameter & $\alpha = 0.01$ & $\alpha = 1.00$ \\
\midrule
Mean valley inflow [m/s] & 2.32 & 2.75 \\
Maximal wind speed [m/s] & 11.86 & 11.55 \\
Average domain wind [m/s] & 4.48 & 4.62 \\
\bottomrule
\end{tabular}
\label{tab:PL_WindPred_wind_valley}
\end{table}

\paragraph{Ramp}
\label{sec:PL_WindPred_ramp}

\begin{figure}[htbp]
\centering
\subfloat[Domain height {\unit[160]{m}}: {\unitfrac[3.85]{m}{s}} average inflow, {\unitfrac[5.59]{m}{s}} average outflow, {\unitfrac[1.26]{m}{s}} minimal wind, {\unitfrac[9.74]{m}{s}} maximal wind.]{\includegraphics[width=.45\columnwidth]{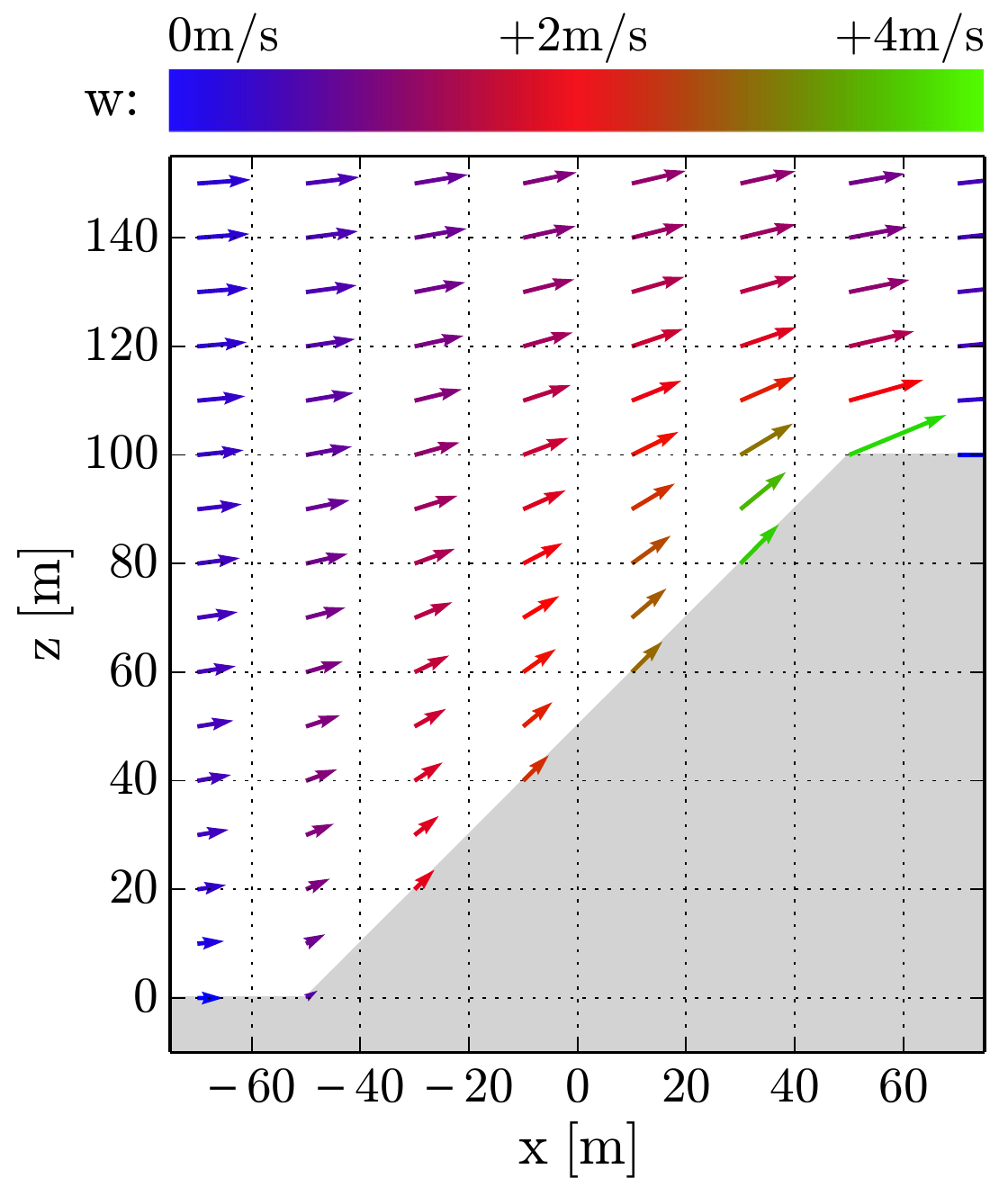}}
\hspace*{.05\columnwidth} % separation between the subfigures
\subfloat[Domain height {\unit[600]{m}}: {\unitfrac[3.79]{m}{s}} average inflow, {\unitfrac[5.79]{m}{s}} average outflow, {\unitfrac[1.23]{m}{s}} minimal wind, {\unitfrac[9.69]{m}{s}} maximal wind.]{\includegraphics[width=0.45\columnwidth]{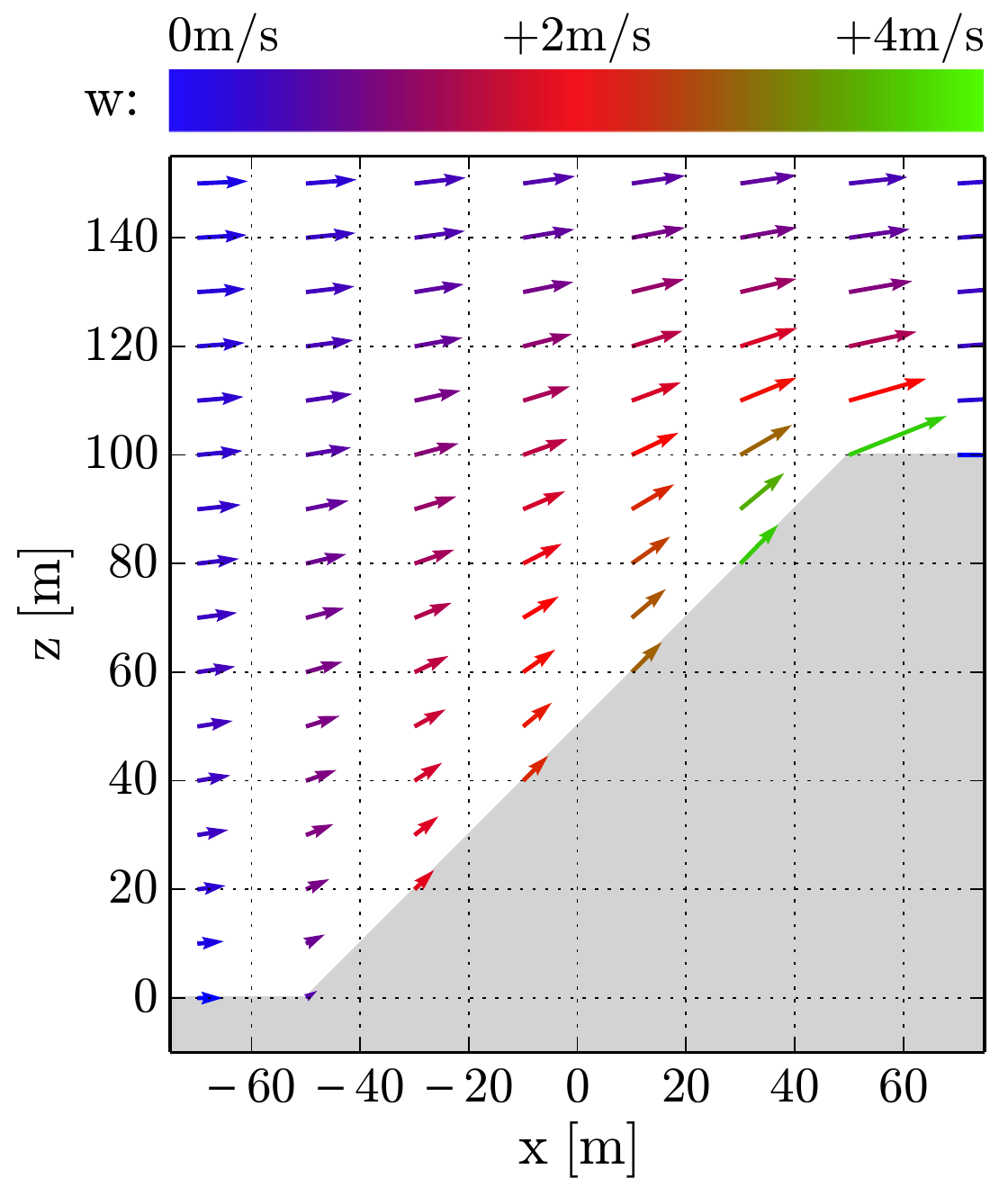}}
\caption[Results for the synthetic ramp test case]{Results for the synthetic ramp test case: Both the observed flow acceleration and the vertical components $w$ correspond well to an intuitive solution. The arrows scale with the magnitude and the color visualizes $w$.}
\label{fig:PL_WindPred_ramp_example}
\end{figure}

The synthetic ramp test case is also initialized with $\vec{u}^\text{I}=[5,0,0]\unitfrac[]{m}{s}$ over the entire domain. The results for the domain heights \unit[160]{m} and \unit[600]{m} are visualized in \cref{fig:PL_WindPred_ramp_example}. The geometry again represents an orifice for the flow, and the overall wind magnitudes are therefore decreased at the inflow and increased at the outflow. The ratio between inflow and outflow velocities is however lower than in the valley test because the geometry narrows less and the top domain boundary is an open flow boundary where some mass is expelled. The vertical flow field and the regions of maximum wind speed close to the top of the ramp --- where soaring flight would be optimal --- are represented well. \note{This next explanation on the diff. between the 160m and 600m test case is a bit shady: 1) Ask Benny 2) remove for journal}However, differences in the vertical component $w$ above \unit[130]{m} can be observed: For the domain size of \unit[160]{m} they are higher and less realistic as the influence of the obstacle should decrease with increasing distance. This is attributable to the fact that the relative sink size (the ratio outflow to inflow) of the initial wind field is much higher for the first case. MATHEW therefore requires a sufficiently large domain height. For the remaining analysis the domain factor, i.e. the ratio of domain height to the height of the tallest obstacle, is therefore set to 3.5. Overall, the results again emphasize that a good initial wind field $\vec{u}^\text{I}$ is required. However, in a qualitative sense the results agree very well with the intuitive solution.

%This comment line needs to be kept here to have the next figure align properly - dont know why. The line cannot even be empty but needs to contain this comment.
%\begin{figure}[!h]
%\centering
%\subfloat[Domain height {\unit[160]{m}}: {\unitfrac[3.85]{m}{s}} average inflow, {\unitfrac[5.59]{m}{s}} average outflow, {\unitfrac[1.26]{m}{s}} minimal wind, {\unitfrac[9.74]{m}{s}} maximal wind.]{\includegraphics[width=.45\columnwidth]{images/WindPred/Results/u_pred_ramp_df01_60.pdf}}
%\hspace*{.05\columnwidth} % separation between the subfigures
%\subfloat[Domain height {\unit[600]{m}}: {\unitfrac[3.79]{m}{s}} average inflow, {\unitfrac[5.79]{m}{s}} average outflow, {\unitfrac[1.23]{m}{s}} minimal wind, {\unitfrac[9.69]{m}{s}} maximal wind.]{\includegraphics[width=0.45\columnwidth]{images/WindPred/Results/u_pred_ramp_df06_00.pdf}}
%\caption[Results for the synthetic ramp test case]{Results for the synthetic ramp test case: Both the observed flow acceleration and the vertical components $w$ correspond well to an intuitive solution. The arrows scale with the magnitude and the color visualizes $w$.}
%\label{fig:PL_WindPred_ramp_example}
%\end{figure}

%%%%%%%%%%%%%%%%%%%%%%%%%%%%%%%%%%%%%%%%%%%%%%%%%%%%%%%%%%%%%%%%%
\subsubsection{Experimental Validation}
\label{sec:PL_WindPred_PrelResults_Experimental}
%%%%%%%%%%%%%%%%%%%%%%%%%%%%%%%%%%%%%%%%%%%%%%%%%%%%%%%%%%%%%%%%%

\paragraph{Measurement and simulation setup}
\label{sec:PL_WindPred_meas_setup}

\begin{figure}[htbp]
\centering
\includegraphics[width=.72\textwidth]{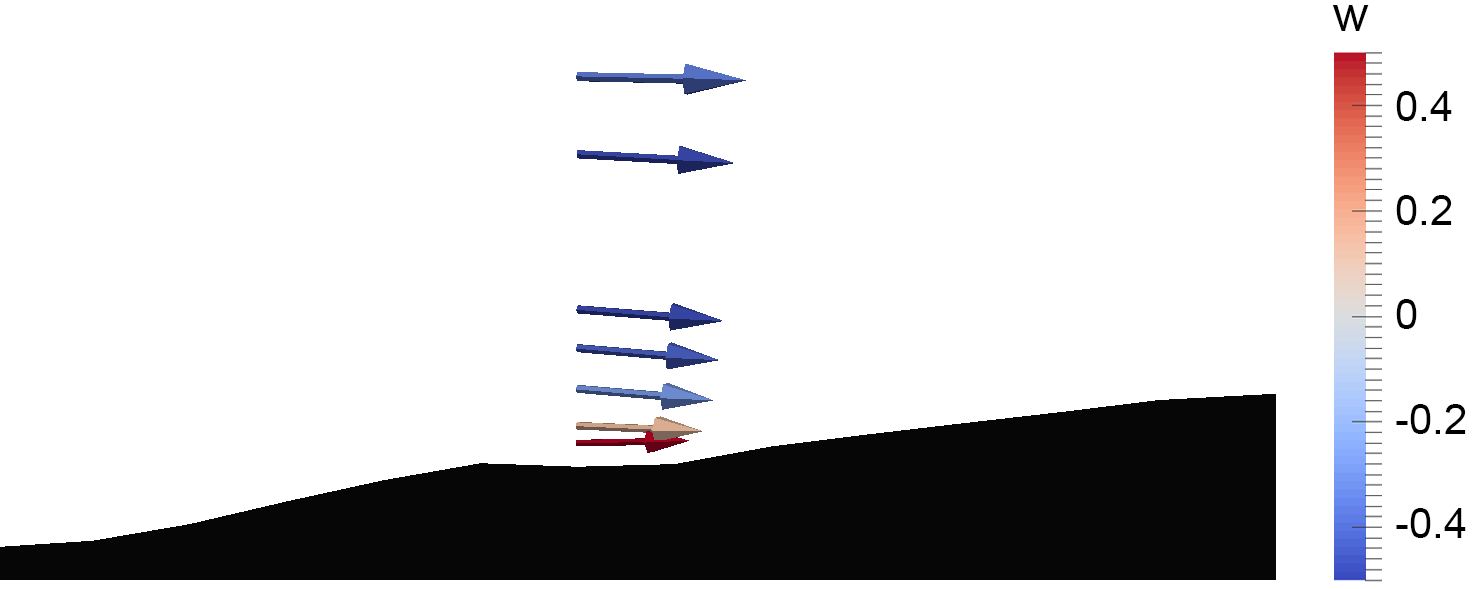}
\caption[LIDAR profile Vorab]{An exemplary LIDAR profile from Vorab, with the arrows representing wind vectors at different heights. While we'd expect $w>0$ due to the left-to-right winds and the increasing terrain altitude in that direction, the LIDAR measurements show $w<0$ for most altitudes. Such \emph{unintuitive} LIDAR measurements occur regularly. They are often caused by terrain or effects that lie outside the computational domain (e.g. large-scale mountain-waves), which the downscaling model can of course not take into account.}
\label{fig:PL_WindPred_vorab_lidar}
\end{figure}

To compare the model output to actual wind data, an extensive literature review for measured multi-dimensional wind fields in complex terrain and at typical flying altitudes of a \ac{UAV} was performed. Unfortunately, even after multiple weeks of searching no such measurements could be found. Therefore, 1-D LIDAR measurements of wind vectors at multiple altitude levels are used. \Cref{fig:PL_WindPred_vorab_lidar} visualizes an exemplary LIDAR profile. The LIDAR data was recorded by the Zurich University of Applied Sciences at three different locations and time periods (\cref{tab:PL_WindPred_campaign_details}). It was filtered and averaged to exclude invalid raw data, missing measurements, or excessive changes within consecutive time steps. More details about the exact instrumentation and measurements can be found in \cite{Kuratle2015} and \cite{Hasler2016} for San Bernardino/Bivio and Vorab, respectively. 

\begin{table}[htb]
\centering
\caption{LIDAR measurement sites in Switzerland, time periods and amount of valid data (in days and valid LIDAR profiles). The days with valid data are less than the time period because data had to be filtered out e.g. when clouds decreased the data quality.}
\begin{tabular}{l l l l l}
\toprule 
Site & Lat.~$^\circ$N/ Lon.~$^\circ$E & Year & Period & Valid data\\ 
\midrule
San Bernardino & 46.46357 / 9.18465 & 2015 & Jun.2--Jun.17 & 2~days/30~prof.\\ 
Bivio & 46.46251 / 9.66864 & 2015 & Jun.17--Jul.21 & 2~days/36~prof.\\ 
Vorab & 46.87398 / 9.18186 & 2016 & Mar.22--Apr.4 & 6~days/70~prof.\\
\bottomrule
\end{tabular}
\label{tab:PL_WindPred_campaign_details}
\end{table}

\begin{figure}[htbp]
\centering
\subfloat[San Bernardino]{\includegraphics[width=.32\columnwidth]{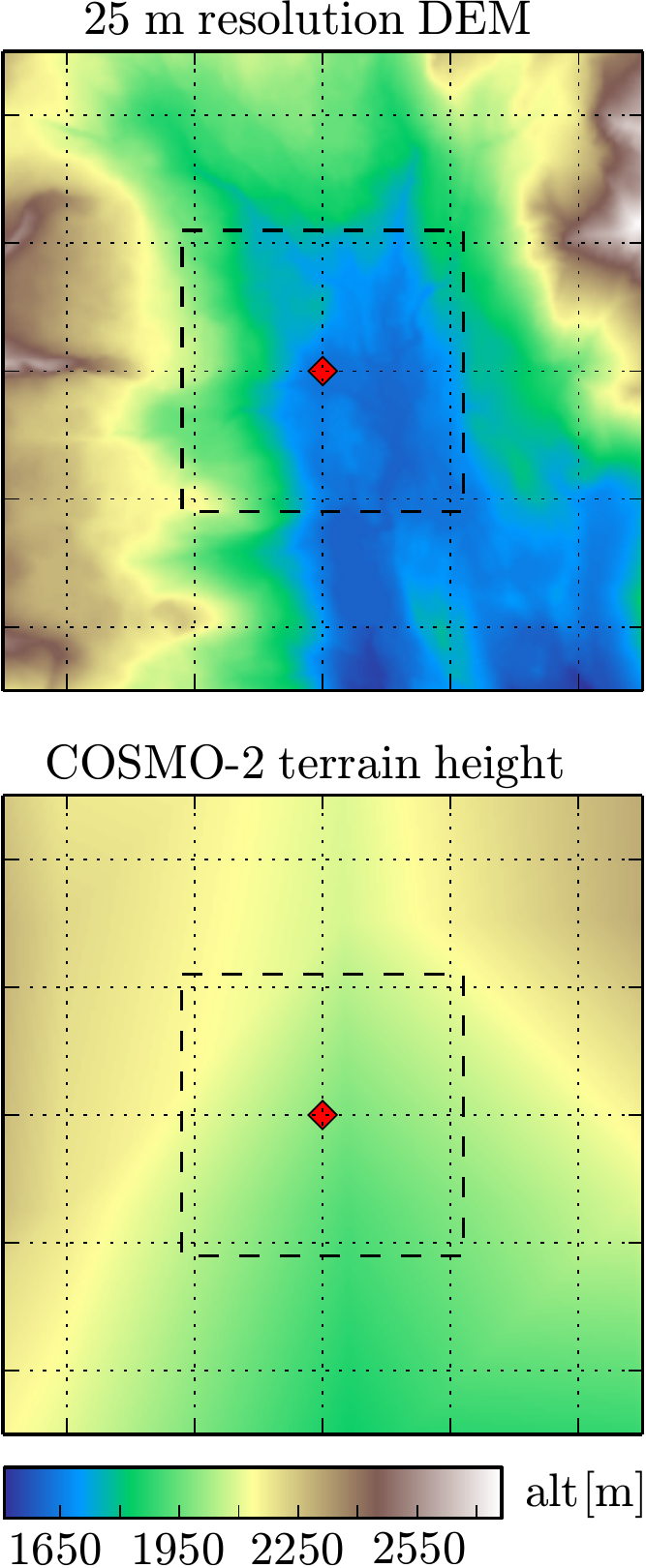}\label{fig:PL_WindPred_terrain_sanbernardino}}
\hspace*{\fill} % separation between the subfigures
\subfloat[Bivio]{\includegraphics[width=.32\columnwidth]{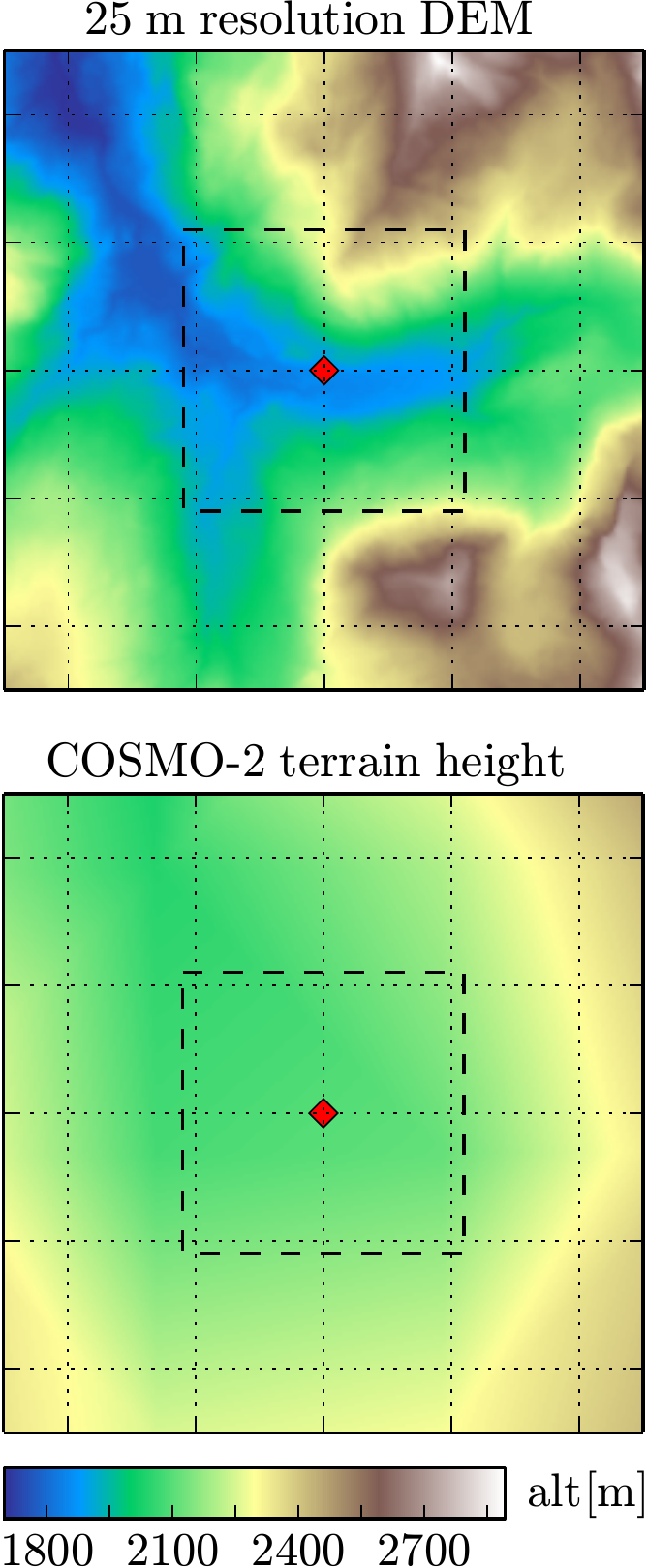}\label{fig:PL_WindPred_terrain_bivio}}
\hspace*{\fill} % separation between the subfigures
\subfloat[Vorab]{\includegraphics[width=.32\columnwidth]{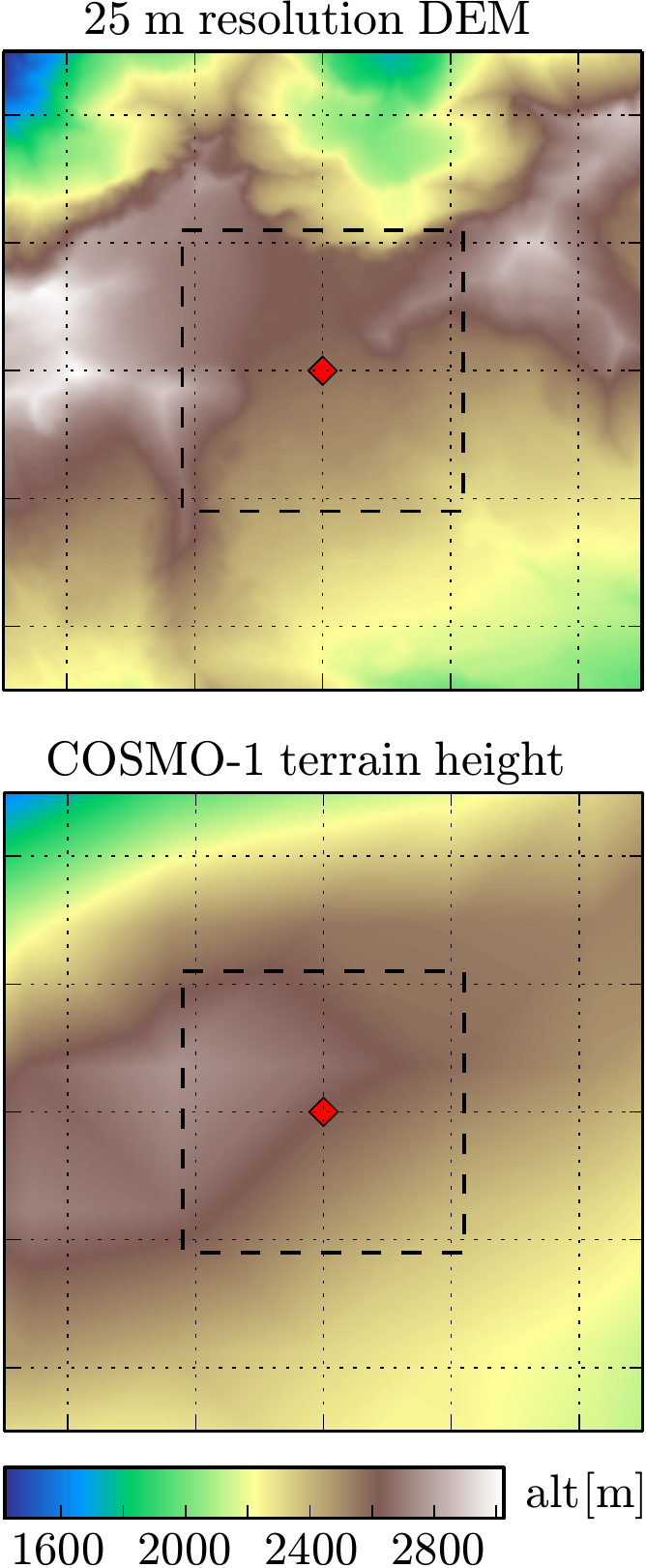}\label{fig:PL_WindPred_terrain_vorab}}
\caption[Terrain measurement locations]{Terrain (\unit[5x5]{km}), downscaling domain (\unit[2.2x2.2]{km}, dashed line) and LIDAR locations (red diamond). North is up. The COSMO-2 terrain is not able to resolve the cluttered terrain in San Bernardino and Bivio. The initial wind field $\vec{u}^\text{I}$ provided by COSMO-2 will thus not be a good approximation for $\vec{u}$. The COSMO-1 terrain represents the terrain at Vorab better. DEM data: \copyright{} 2016 swisstopo (JD100042), free for educational purposes.}
\label{fig:PL_WindPred_terrains}
\end{figure}

\Cref{fig:PL_WindPred_terrains} visualizes the terrain, the computational domain and the LIDAR, which is always positioned at the center of the computational domain with side length \unit[2.2]{km}. A grid with 20 vertical layers and \unit[50]{m} horizontal resolution is used. Note that for San Bernardino and Bivio only COSMO-2 weather data is available, while the higher-resolution COSMO-1 data can be used for Vorab.

%\footnotetext{\url{http://geodata4edu.ethz.ch/}}

\paragraph{Measured and estimated wind fields}
\label{sec:PL_WindPred_phenomena}

\begin{figure}[htbp]
\centering
\subfloat{\includegraphics[width=0.9\columnwidth]{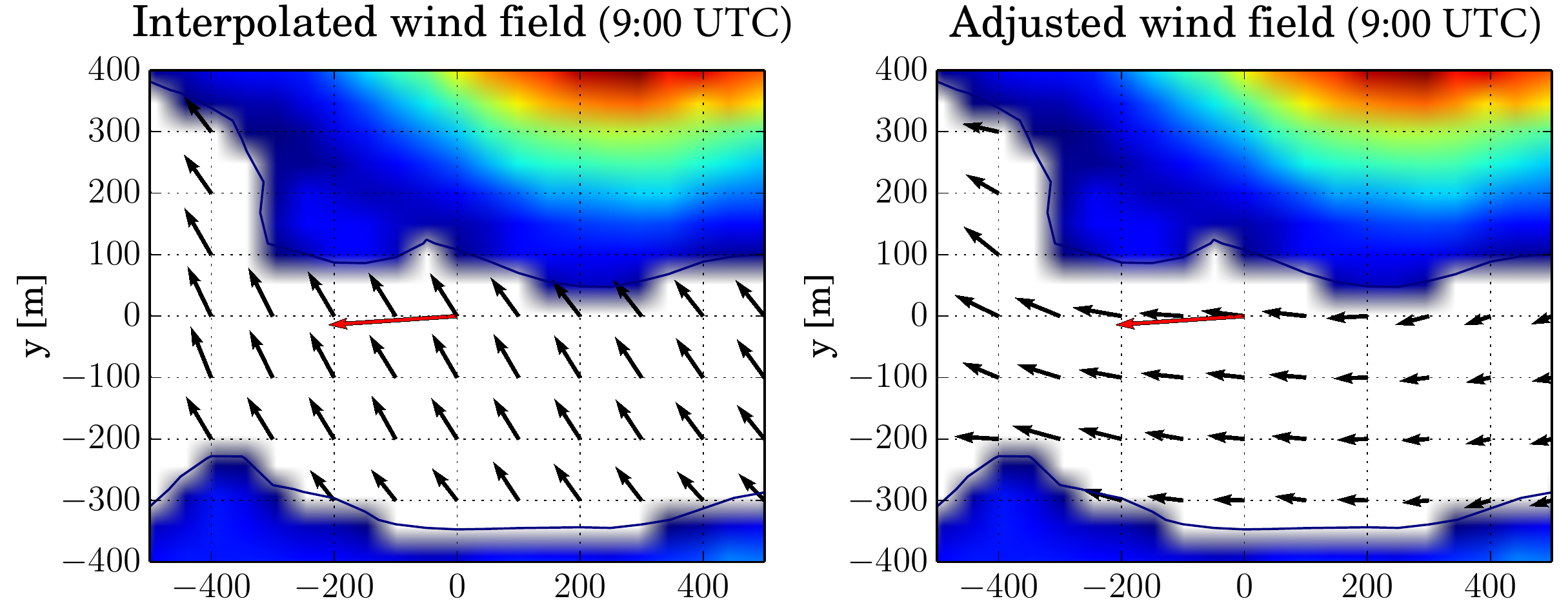}}\\[0.7ex]%[-2ex]
\subfloat{\includegraphics[width=0.9\columnwidth]{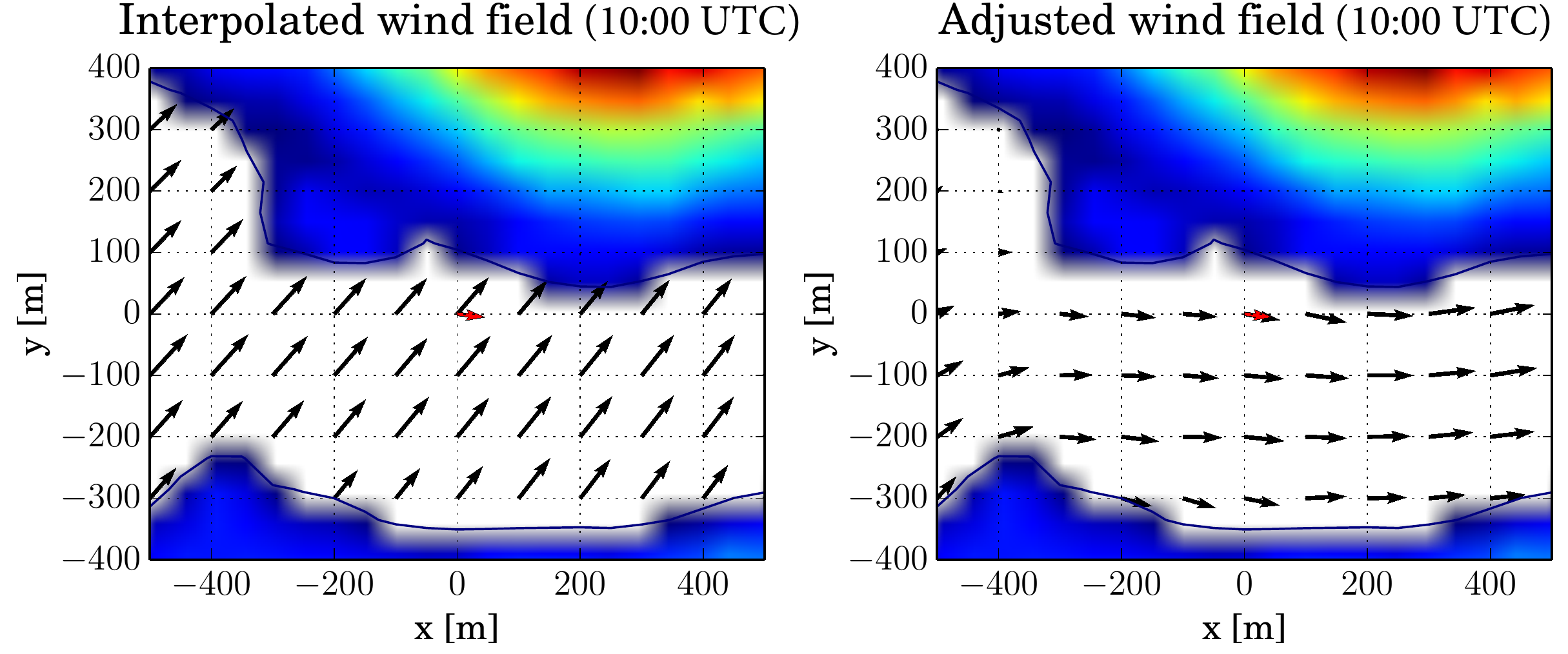}}
\caption[Valley breeze Bivio]{Initial and adjusted wind fields $\vec{u}$ (black arrows) at Bivio for 09:00UTC (top) and 10:00UTC (bottom) on July 8th 2015. The wind is shown at \unit[1900]{m} AMSL. The red arrow is the wind measured by the LIDAR.}
\label{fig:PL_WindPred_valley_breeze}
\end{figure}

To show how the downscaling model can improve the wind field estimate, but to also show its limitations with respect to modelling small-scale thermally induced wind phenomena, this section exemplarily compares the model output and measured winds for Bivio. As visible in \cref{fig:PL_WindPred_terrain_bivio} the LIDAR is located in a steep east-west valley but the corresponding COSMO-2 terrain is basically flat. As a result, on July 8th 9:00 UTC the interpolated COSMO-2 wind field indicates flow coming from south-east (\cref{fig:PL_WindPred_valley_breeze}) and therefore violates the assumption of no flow through the terrain boundaries. The downscaling model is able to rotate $\vec{u}$ in the correct direction such that the winds satisfy the terrain boundary conditions. The measured wind direction agrees well with the downscaled wind field. However, the wind speed is significantly underestimated both by the initial wind field and the adjusted wind field. The overall observed situation is a so called mountain breeze, i.e. a cold air mass that moves from the mountain into the valley during the night and early morning hours. 

One hour later, at 10:00 UTC, the wind direction has changed and wind is flowing from the valley to the mountains. This \emph{valley breeze} situation is shown in \cref{fig:PL_WindPred_valley_breeze}. The interpolated wind direction again does not respect the terrain boundaries and also differs significantly from the measured wind direction. The downscaling model is again able to predict the correct wind direction. The wind magnitude is overestimated by both approaches, but the error for the downscaling model is much smaller.

\begin{figure}[htbp]
\centering
\subfloat{\includegraphics[width=.7\columnwidth]{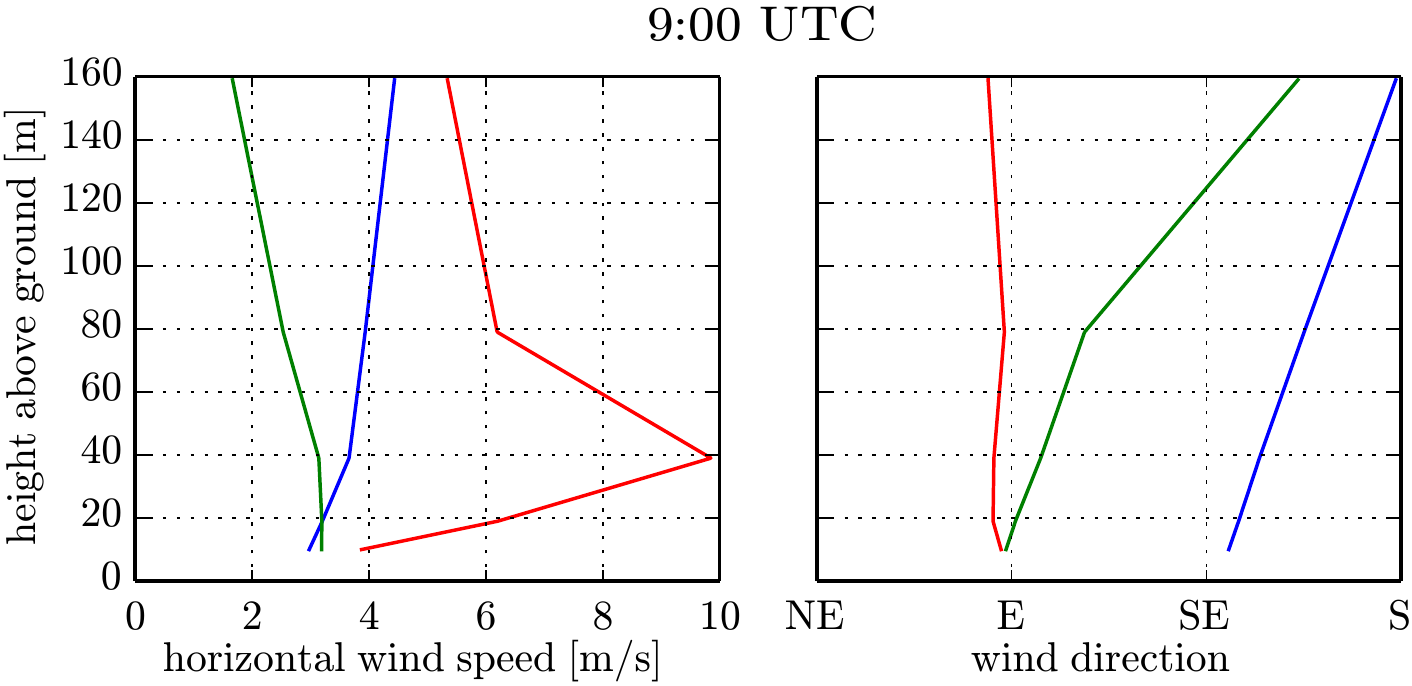}}\\[1.2ex]
\subfloat{\includegraphics[width=.7\columnwidth]{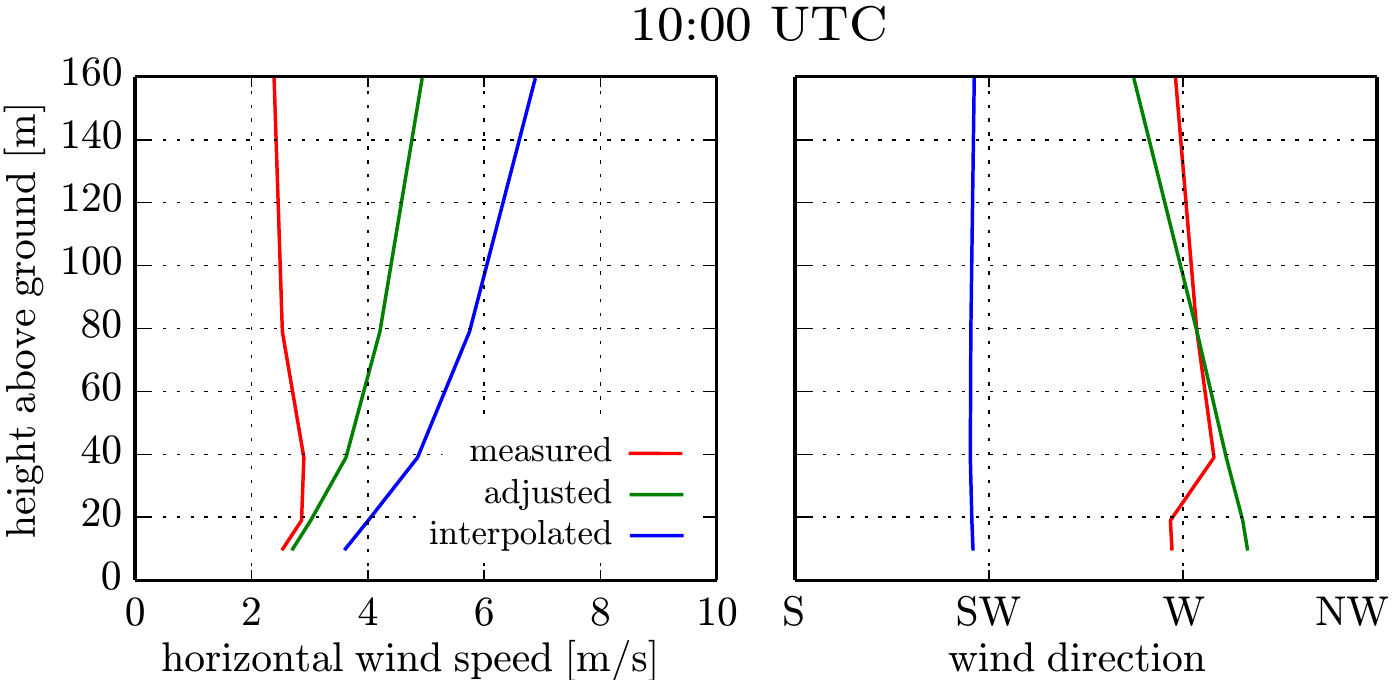}}%\\
%\subfloat{
%\small
%\vspace{10pt}
%\definecolor{darkgreen}{rgb}{0,0.5,0}
%\textcolor{red}{\textbf{---}} measured profile \hspace{10pt} \textcolor{blue}{\textbf{---}} interpolated profile \hspace{10pt} \textcolor{darkgreen}{\textbf{---}} adjusted profile}
\caption{Wind profiles at Bivio on July 8th 2015 at 9:00 UTC (top) and 10:00 UTC (bottom). In the top plot, the LIDAR measures a so-called \emph{low-level jet}. The downscaling model significantly improves the wind direction and tends to improve the wind speed.}
\label{fig:PL_WindPred_valley_breeze_profile}
\end{figure}

To analyze the reason for the underestimated wind speed at 09:00 UTC, \cref{fig:PL_WindPred_valley_breeze_profile} plots the measured, interpolated and modeled wind profiles. The measured horizontal wind profile shows a distinct maximum \unit[40]{m} above ground. This small-scale phenomenon is called \emph{low-level jet} and is not represented in the interpolated wind profile. Given that the downscaling model has no deeper physical understanding of thermally-induced effects, it also cannot predict this behavior. At 10:00 UTC, the interpolated wind is again too high but the downscaling model can partially correct for this. In both cases, the downscaling model clearly improves the wind direction such that the overall agreement between the modeled and measured wind vectors is improved. 

\paragraph{Statistical evaluation}
\label{sec:PL_WindPred_evaluation}

To quantitatively assess the wind field improvement, a comparison of the downscaling model against LIDAR measurements is performed over all test cases. Given that we only want to investigate modeling errors but not errors due to a badly selected stability parameter, we always first run the model for $\alpha = [10^{-6},10^1]$ and select the run with the \emph{optimal} $\alpha$, i.e. the one which yields the minimum \ac{RMSE} between the modeled and measured wind profile. This approach of course has a certain risk of overfitting such that the following results have to be considered best-case results. In the future, the wind field downscaling model will automatically select the best $\alpha$ by assessing the current atmospheric stratification using the temperature profiles in the \ac{NWP} data.

Given that only 1D measurement data exists, all wind fields can only be compared locally at the LIDAR's position. In addition to the downscaled wind field $\vec{u}$, we also compare the initial wind field $\vec{u}^\text{I}$ and a zero-wind field $\vec{u}^0=0$. The LIDAR measurements serve as ground-truth. The comparison metric is the root mean square error. It is calculated for two newly introduced variables: First, the weighted horizontal vector error\footnote{It would make more sense to compare the horizontal wind speed and direction instead of combining them in one variable. But given that the zero wind field $\vec{u}_0$ has no proper wind direction, the above introduced variable seems to be the best solution.} is
	\begin{equation}
	e_\text{hor} = \frac{\sqrt{(u_\text{p} - u_\text{m})^2 + (v_\text{p} - v_\text{m})^2}}{v_\text{air}} \; . 
	\end{equation}
Here, the subscripts $\text{p}$ and $\text{m}$ stand for the predicted and measured components, and $v_\text{air}$ is the nominal airspeed of \unitfrac[9]{m}{s} for \emph{AtlantikSolar}. The variable $e_\text{hor}$ therefore normalizes the horizontal wind estimate error with the aircraft speed. If $e_\text{hor}>1$, the wind is too strong to overcome the wind and collision with terrain can occur. The second variable is the weighted vertical error
	\begin{align}
	e_\text{ver} = \begin{cases} \frac{w_\text{p} - w_\text{m}}{w_\text{sr}} & \textnormal{if } w_\text{p} - w_\text{m} \leq 0, \\ \frac{w_\text{p} - w_\text{m}}{w_\text{cr}} & \textnormal{if } w_\text{p} - w_\text{m} > 0 \; , \end{cases}
	\end{align}
where $w_\text{sr}$ is the maximum sink rate (positive) and $w_\text{cr}$ the maximum climb rate. For \emph{AtlantikSolar} these are \unitfrac[3]{m}{s} and \unitfrac[1.5]{m}{s} respectively. This error is especially important for fragile or low-power aircraft: For example, if the predicted vertical wind $w_\text{p}=\unitfrac[0]{m}{s}$ and $e_\text{ver}<-1$, then the aircraft sink rate is exceeded when altitude hold is required and the aircraft may get damaged. Equivalently, when $w_\text{p}=\unitfrac[0]{m}{s}$ but $e_\text{ver}>1$, then the actual downwards wind magnitude is higher than the aircraft climb rate and the aircraft is in risk of colliding with terrain. Overall, if any of the error magnitudes exceed unity, the aircraft is in danger. 

\Cref{tab:PL_WindPred_rmse} shows the final results. The root mean square horizontal errors, vertical errors, and --- given that they were already properly normalized with the aircraft nominal speed, climb rate and sink rate --- their sum are presented. In Bivio, the vertical error is reduced by neither the initial nor the downscaled wind field but the horizontal error is reduced by \unit[38]{\%} by the downscaling model. In San Bernardino, the vertical error is slightly reduced by both $\vec{u}^\text{I}$ and $\vec{u}$. The horizontal error is reduced by \unit[27]{\%} by $\vec{u}^\text{I}$ and by \unit[31]{\%} by $\vec{u}$. In Vorab, the initial wind field yields a much larger vertical error than the zero-wind assumption and the downscaling model can only marginally compensate for that. The horizontal error is however reduced by \unit[46]{\%} by $\vec{u}^\text{I}$ and by \unit[48]{\%} by $\vec{u}$. Averaged over all three test cases (with equal weight for each), it becomes clear that both COSMO-2 and the higher-resolution COSMO-1 model provide initial wind fields which have similar or higher vertical errors than the zero-wind assumption. The low-quality terrain representation (\cref{fig:PL_WindPred_terrains}) is certainly contributing to this issue. The downscaling model then does not have the leverage to significantly improve this: On average, $\vec{u}^\text{I}$ increases the vertical error by \unit[30]{\%}, and the error increase after downscaling is still \unit[26]{\%}. However, the horizontal error is reduced by \unit[31]{\%} by the initial wind field and by \unit[41]{\%} for the downscaled wind field. The sum of both errors reduces by \unit[15]{\%} with the initial wind field and by \unit[23]{\%} with the presented downscaling method.

\begin{figure}[htbp]
\centering
\includegraphics[width=0.8\columnwidth]{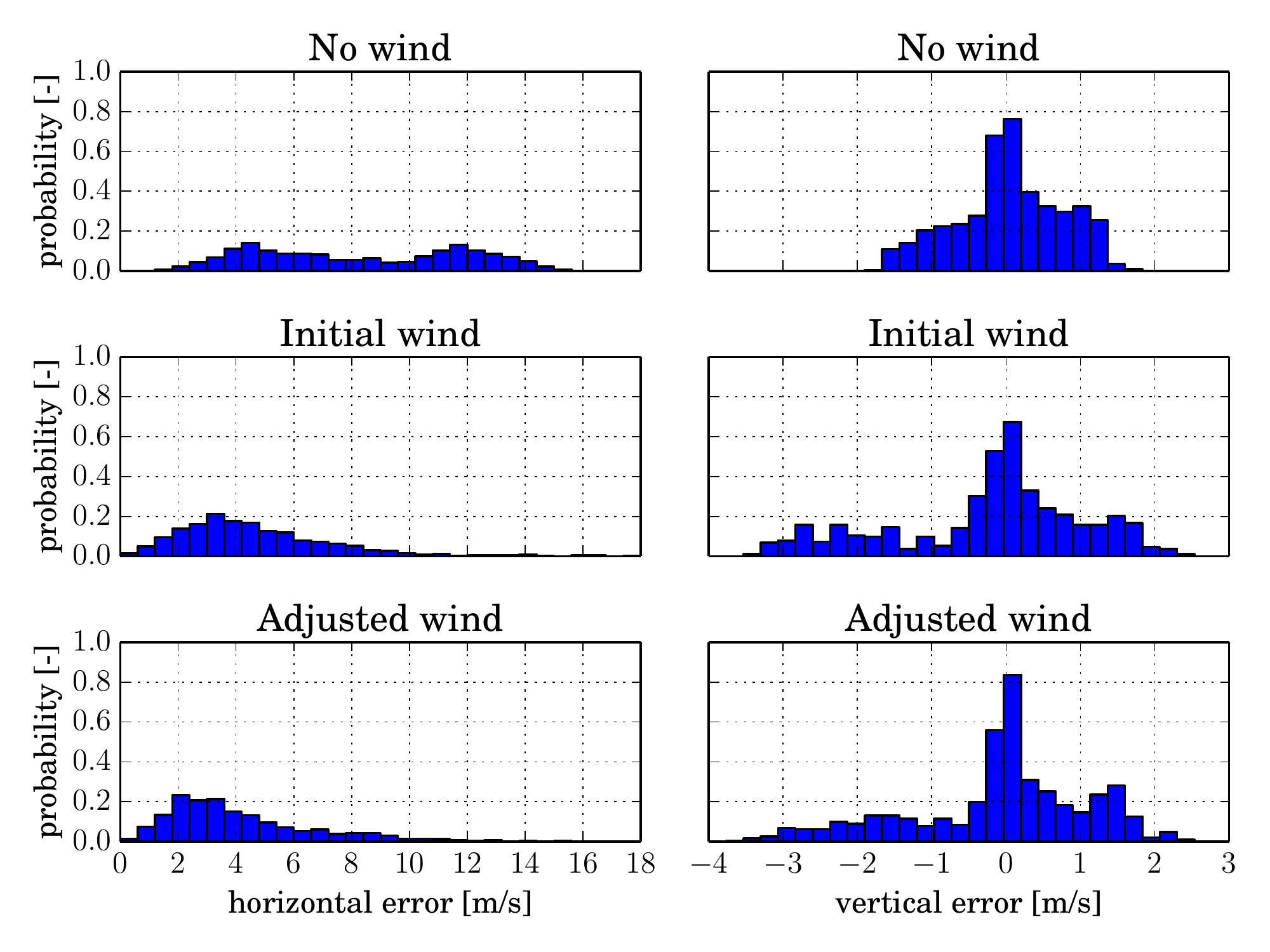}
\caption[Error histograms]{Error histograms: The horizontal error distribution is clearly improved by the downscaling model, but the vertical component is worse than the no wind assumption. }
\label{fig:PL_WindPred_hist_unweighted}
\end{figure}

The error histograms are displayed in \cref{fig:PL_WindPred_hist_unweighted}. In a first step, only the horizontal error distribution is analysed: The no wind assumption shows a wide distribution with two peaks at approximately \unitfrac[4]{m}{s} and \unitfrac[12]{m}{s}. The interpolation of \ac{NWP} data already performs better and the respective error distribution has only one peak at slightly below \unitfrac[4]{m}{s}. The output for the model with optimal stability parameter performs even better, the respective peak is around \unitfrac[3]{m}{s}. Looking at the vertical error distributions, the zero wind assumption leads to the smallest \ac{RMSE}. The interpolated wind field shows a much higher vertical RMSE and thus more spread in the error distribution. The adjusted wind field shows a very similar distribution. As discussed before, this mainly comes from the measurements in Vorab ($e_\text{ver}=$0.713/0.272/0.142 for Vorab/San Bernardino/Bivio respectively), which show vertical winds that cannot be described by simple phenomena (see for example \cref{fig:PL_WindPred_vorab_lidar}) and are therefore not covered in \ac{COSMO} data. It should be noted that modeling local anomalies such as mountain waves or thermal flows is also very challenging for more complex (e.g. \emph{prognostic}) downscaling models. In addition it is also possible that the LIDAR data is error-prone because of clouds, fog or precipitation. All in all, the adjusted wind field however has a 23\% and 15\% smaller \ac{RMSE} than the zero-wind and interpolated wind fields respectively.

\begin{table}[htb]
\centering
\caption[Root mean square errors]{Horizontal and vertical wind estimate root mean square errors for the zero-wind assumption $\vec{u}^0$, the \ac{NWP}-based interpolation $\vec{u}^\text{I}$ and the adjusted wind field $\vec{u}$ with optimal stability parameter. The LIDAR measurement serves as wind ground truth. The percentage values are the relative changes with respect to the zero-wind assumption.}
\begin{tabular}{l l l l}
\toprule
Type & RMSE($e_\text{ver}$) & RMSE($e_\text{hor}$) & Sum \\
\midrule \\[-1.8ex]
\emph{Bivio} &&&\\\addlinespace[-0.1ex]\cmidrule(l{6pt}){1-1}
Zero wind & 0.130 & 0.541 & 0.671 \\
Interpolated wind & 0.159 (+22\%) & 0.522 (-4\%)& 0.681 (+1\%)\\ 
Optimal adjusted wind & 0.142 (+9\%)& 0.339 (-38\%)& 0.481 (-28\%)\\[1.2ex]%\cmidrule{1-1} 
\emph{San Bernardino} &&&\\\addlinespace[-0.1ex]\cmidrule(l{6pt}){1-1}
Zero wind & 0.299 & 0.713 & 1.012 \\
Interpolated wind & 0.268 (-10\%)& 0.517 (-27\%)& 0.785 (-22\%)\\ 
Optimal adjusted wind & 0.272 (-9\%)& 0.493 (-31\%)& 0.765 (-24\%)\\[1.2ex]
\emph{Vorab} &&&\\\addlinespace[-0.1ex]\cmidrule(l{6pt}){1-1}
Zero wind & 0.466 & 1.200 & 1.666 \\
Interpolated wind & 0.737 (+58\%)& 0.643 (-46\%)& 1.380 (-17\%)\\ 
Optimal adjusted wind & 0.713 (+53\%)& 0.627 (-48\%)& 1.340 (-20\%)\\[1.2ex] 
%\emph{All test cases combined} &&&\\\addlinespace[-0.1ex]\cmidrule(l{6pt}){1-1}
%Zero wind & 0.387 & 1.009 & 1.396 \\
%Interpolated wind & 0.586 (+51\%)& 0.596 (-41\%)& 1.182 (-15\%)\\ 
%Optimal adjusted wind & 0.567 (+47\%)& 0.552 (-45\%)& 1.119 (-20\%)\\
\emph{All test cases combined} &&&\\\addlinespace[-0.1ex]\cmidrule(l{6pt}){1-1}
Zero wind & 0.298 & 0.818 & 1.116 \\
Interpolated wind & 0.388 (+30\%)& 0.561 (-31\%)& 0.949 (-15\%)\\ 
Optimal adjusted wind & 0.376 (+26\%)& 0.486 (-41\%)& 0.862 (-23\%)\\ 
\bottomrule
\end{tabular}
\label{tab:PL_WindPred_rmse}
\end{table}

%%%%%%%%%%%%%%%%%%%%%%%%%%%%%%%%%%%%%%%%%%%%%%%%%%%%%%%%%%%%%%%%%
%\subsection{Conclusion}
%\label{sec:PL_WindPred_Conclusion}
%%%%%%%%%%%%%%%%%%%%%%%%%%%%%%%%%%%%%%%%%%%%%%%%%%%%%%%%%%%%%%%%%

%%%%%%%%%%%%%%%%%%%%%%%%%%%%%%%%%%%%%%%%%%%%%%%%%%%%%%%%%%%%%%%%%
%%%%%%%%%%%%%%%%%%%%%%%%%%%%%%%%%%%%%%%%%%%%%%%%%%%%%%%%%%%%%%%%%
\section{Real-Time Wind-Aware Path Planning for UAVs}
\label{sec:PL_Planning}
%%%%%%%%%%%%%%%%%%%%%%%%%%%%%%%%%%%%%%%%%%%%%%%%%%%%%%%%%%%%%%%%%
%%%%%%%%%%%%%%%%%%%%%%%%%%%%%%%%%%%%%%%%%%%%%%%%%%%%%%%%%%%%%%%%%

This section presents a wind-aware path planner that runs in real time on a UAV's onboard computer and incorporates the real-time 3D wind field predictions from \cref{sec:PL_WindPrediction}. It uses near-optimal Dubins aircraft paths to represent the vehicle dynamics and terrain information from a \ac{SLAM} system to generate safe and efficient paths through wind and cluttered terrain. This section presents the state of the art, fundamentals, improvements for fast shortest-path planning (\cref{sec:PL_Planning_Improvements}), the wind-aware time-optimal planning (\cref{sec:PL_Planning_TimeOptimal}), and overall results (\cref{sec:PL_Planning_Results}).

%%%%%%%%%%%%%%%%%%%%%%%%%%%%%%%%%%%%%%%%%%%%%%%%%%%%%%%%%%%%%%%%%
\subsection{State of the Art}
\label{sec:PL_Planning_StateOfTheArt}
%%%%%%%%%%%%%%%%%%%%%%%%%%%%%%%%%%%%%%%%%%%%%%%%%%%%%%%%%%%%%%%%%

Motion planning can be performed using grid-based approaches~\cite{Hart1968Astar}, Artificial Potential Fields~\cite{Khatib1986PotentialFieldPathPlanning} or sampling-based methods. Rapidly-exploring Random Trees (RRT)~\cite{LaValle1998RRT} are a well-known non-optimal sampling-based planning technique for high dimensional problems or vehicles with nonholonomic constraints. The non-optimal RRT has been extended with the asymptotically optimal RRT*~\cite{Karaman2010RRTstarKinodynamic}. Other optimal sampling-based planners include BIT*~\cite{Gammell2015BITstar}, PRM*~\cite{Karaman2011RRTstarPRMstar}, or FMT*~\cite{Janson2014FMTstar}. \citet{Gammell2014IRRTstar} introduce IRRT*, a planner that uses information from already known solutions to improve the convergence to the optimal path. Sampling-based planners have been applied to time-optimal planning in uniform wind by \citet{Ceccarelli2007}, \citet{Hota2009}, and \citet{Schopferer2015}. The extension, time-optimal planning in non-uniform wind, is presented by \citet{Lawrance2011} and \citet{Otte2016}. The approach presented by \citet{Chakrabarty2013PlanningDynWindFields} is able to compute the time-optimal path in non-uniform time-varying wind fields. In previous work~\cite{Oettershagen2018Metpass} we have presented a similar approach that plans optimal paths considering extensive non-uniform time-varying weather data, but does not run in real time. To our knowledge, no approach that can plan cost-optimal aircraft paths in non-uniform 3D wind fields in real time has been demonstrated.

%%%%%%%%%%%%%%%%%%%%%%%%%%%%%%%%%%%%%%%%%%%%%%%%%%%%%%%%%%%%%%%%%
\subsection{Fundamentals}
\label{sec:PL_Planning_Fundamentals}
%%%%%%%%%%%%%%%%%%%%%%%%%%%%%%%%%%%%%%%%%%%%%%%%%%%%%%%%%%%%%%%%%

%Note 1: For a journal paper, the fundamentals section needs to be shortened significantly. We don't need to explain everything in such detail in a journal.
%Note 2: For a journal paper: We talk a lot about IRRT* and informed subsets, but then don't use it at all in the results. This does not fit for a journal! -> Maybe remove it there?

Motion planning is the problem of finding a path from start state $q_\text{start}\in Q_\text{free}$ to the goal configuration $q_\text{goal}\in Q_\text{free}$ through the free space $Q_\text{free}={Q\setminus Q_\text{obs}}$ while obeying additional constraints (e.g. vehicle dynamics). Here, $Q$ is the set of all possible robot configurations and $Q_\text{obs}$ is the set of robot states that are infeasible e.g. due to obstacles. While the states can be selected using grid-based approaches~\cite{Hart1968Astar}, this paper focuses on sampling-based planning methods and more specifically the asymptotically optimal RRT*~\cite{Karaman2010RRTstarKinodynamic} and its informed counterpart IRRT*~\cite{Gammell2014IRRTstar}.

%%%%%%%%%%%%%%%%%%%%%%%%%%%%%%%%%%%%%%%%%%%%%%%%%%%%%%%%%%%%%%%%%
\subsubsection{Sampling-Based Path Planning Methods}
\label{sec:PL_Planning_Fundamentals_SamplingBased}
%%%%%%%%%%%%%%%%%%%%%%%%%%%%%%%%%%%%%%%%%%%%%%%%%%%%%%%%%%%%%%%%%

%%%%%%%%%%%%%%%%%%%%%%%%%%%%%%%%%%%%%%%%%%%%%%%%%%%%%%%%%%%%%%%%%
\paragraph{RRT*}
%%%%%%%%%%%%%%%%%%%%%%%%%%%%%%%%%%%%%%%%%%%%%%%%%%%%%%%%%%%%%%%%%
Sampling-based planners are advantageous because their computational complexity stays manageable for higher-dimensional problems and $Q_\text{free}$ does not have to be constructed explicitly. Such planners are \emph{probabilistically complete}, i.e. are guaranteed to find a solution if one exists and infinite time is available. RRT* is also \emph{asymptotically optimal}, i.e. the path converges to the optimal path over time. \Cref{alg:PL_Planning_RRTstar} contains pseudo-code for RRT*. After initialization of the motion tree $T$, such planners randomly select a configuration $q_\text{rand}$ from $Q$. The current configuration $q_\text{nearest}\in T$ which has the smallest distance to $q_\text{rand}$ is then found. The actual new state $q_\text{new}$ is then selected such that it lies on the path between $q_\text{rand}$ and $q_\text{nearest}$ and is no further than $d_\text{max}$ away from $q_\text{nearest}$. This capping step in the so-called steering function avoids excessively long path segments in $T$ because they have a higher probability of collision and therefore yield only few valid motions. Next the motion from $q_\text{nearest}$ to $q_\text{new}$ is checked for validity, meaning that it does not collide with any obstacle. If that is the case, then either the $k$-nearest neighbors of $q_{\text{new}}$ or all states within a distance $r$ from $q_{\text{new}}$ are determined and stored in $Q_\text{near}$. The values of $k$ and $r$ are dependent on the number of nodes in the motion tree $n$:
\begin{align}
k &= k_{\text{rrg}} \ln \left( n \right)\label{eqn:PL_Planning_k_nearest}\\
r &= \sqrt[\text{dim}]{r_{\text{rrg}} \frac{\ln(n)}{n}} \label{eqn:PL_Planning_r_nearest}\; ,
\end{align}
where $k_{\text{rrg}}$ and $r_{\text{rrg}}$ are variable parameters and $\text{dim}$ is the dimension of the state space. The neighbor in $Q_\text{near}$ with the lowest cost-to-go to $q_\text{new}$ (with respect to a cost function $f_\text{cost}$) is selected as parent $q_\text{parent}$. The edge from $q_\text{parent}$ to $q_\text{new}$, which is part of the currently best-known path from $q_\text{start}$ to $q_\text{new}$, is added to the motion tree $T$ and $q_\text{new}$ is added as a vertex. The so-called \emph{rewiring} step is one of the main differences to RRT as it provides the optimality-properties of RRT*: It checks whether any of the paths from $q_\text{new}$ to the elements of $Q_\text{near}$ is more optimal than the previously known respective path. If so, then that path is rewired, i.e. the parent of $q\in Q_\text{near}$ is set to $q_\text{new}$. The RRT* algorithm terminates if the termination condition, e.g. expressed through the number of iterations or a certain error metric, is fulfilled.

\begin{algorithm}
\caption{RRT* Algorithm: A motion tree $T$ is generated from the initial configuration $q_\text{start}$ until the \textsc{terminationCondition} is fulfilled.}
\label{alg:PL_Planning_RRTstar}
\begin{algorithmic} [1]
\Procedure{solve}{}
\State $T$.init$\left(q_\text{start}\right)$
\While {$\left( \textsc{terminationCondition} == \text{false} \right)$}
  \State $q_\text{rand} \gets \textsc{sampleRandomState} \left(\right)$
  \State $q_\text{nearest} \gets \textsc{getNearestNeighbor} \left(q_\text{rand}, T\right)$
  \State $q_{\text{new}} \gets \textsc{Steer} \left(q_\text{nearest}, q_\text{rand}, d_{\text{max}}\right)$
  \If {$\textsc{motionValid} \left(q_\text{nearest}, q_{\text{new}}\right)$}
    \State $Q_\text{near} \gets \textsc{getNeighbors} \left(q_{\text{new}}, T\right)$
    \State $q_\text{parent}, \text{cost} \gets \textsc{getBestParent} \left(q_{\text{new}}, Q_\text{near}, f_{\text{cost}}\right)$
    \State $T$.add\_edge$\left(q_\text{parent}, 
    q_{\text{new}}, \text{cost}\right)$
    \State $T$.add\_vertex$\left(q_{\text{new}}\right)$
    \State $\textsc{rewireTree}\left(q_{\text{new}}, Q_\text{near}, f_{\text{cost}}\right)$
  \EndIf
\EndWhile
\EndProcedure
\end{algorithmic}
\end{algorithm}

%%%%%%%%%%%%%%%%%%%%%%%%%%%%%%%%%%%%%%%%%%%%%%%%%%%%%%%%%%%%%%%%%
\paragraph{Informed RRT*}
%%%%%%%%%%%%%%%%%%%%%%%%%%%%%%%%%%%%%%%%%%%%%%%%%%%%%%%%%%%%%%%%%
The Informed RRT* algorithm~\cite{Gammell2014IRRTstar} proceeds exactly like RRT* until a first feasible path is found. After that, it leverages the information on the initial solution to speed up the convergence towards an optimal solution: First, it only samples the informed subset $Q_\text{inf}$, i.e. the solutions which are known to be able to improve the solution:
\begin{equation}
Q_\text{inf}=\{q\in Q|h(q)\leq c_\text{best}\} \; .
\end{equation}
Here, $c_\text{best}$ is the cost of the current best solution and $h(q)$ is a heuristic function which returns a lower limit for the cost of a path from start $q_\text{start}$ to goal $q_\text{goal}$ which is constrained to go through $q$. As an example, for shortest-path planning problems in $\mathbb{R}^n$ the Euclidean distance is a valid heuristic such that the informed subset becomes
\begin{equation}
Q_\text{inf}= \big\{ q\in Q \big| \Vert q_\text{start}-q\Vert_{_2}+\Vert q_\text{goal}-q\Vert_{_2} \leq c_\text{best} \big\} \; .
\end{equation}
This informed subset is an $n$-dimensional prolate hyperspheroid (a special form of a hyperellipsoid) with focal points $q_\text{start}$ and $q_\text{goal}$. For simple cases such as an ellipsoid, states from the informed subset are sampled \emph{directly} \cite{Gammell2014IRRTstar}. For more complicated cases such as general cost functions or differential constraints, \emph{indirect} sampling based on rejecting all samples $q\not\in Q_\text{inf}$ has to be used. Given that this process becomes very inefficient for high-dimensional problems, \citet{Kunz2016HRS} propose the hierarchical rejection sampling method. Second, Informed RRT* performs \emph{tree pruning}, i.e. after a new solution is found all states in the motion tree $T$ that cannot improve the solution anymore are removed. Overall, the combination of informed sampling and tree pruning significantly speeds up the planner's convergence.

%%%%%%%%%%%%%%%%%%%%%%%%%%%%%%%%%%%%%%%%%%%%%%%%%%%%%%%%%%%%%%%%%
\subsubsection{Dubins Airplane Paths Without Wind}
\label{sec:PL_Planning_Fundamentals_DubinsNoWind}
%%%%%%%%%%%%%%%%%%%%%%%%%%%%%%%%%%%%%%%%%%%%%%%%%%%%%%%%%%%%%%%%%

%%%%%%%%%%%%%%%%%%%%%%%%%%%%%%%%%%%%%%%%%%%%%%%%%%%%%%%%%%%%%%%%%
%\paragraph{Dubins Path}
%\label{sec:PL_dubinspath}
%%%%%%%%%%%%%%%%%%%%%%%%%%%%%%%%%%%%%%%%%%%%%%%%%%%%%%%%%%%%%%%%%
The Dubins path~\cite{Dubins1957DubinsCar} is the shortest path between two states in an obstacle-free world for a car which can only drive forward at constant velocity and turn at a bounded turning rate. The shortest path consists of three segments which are either straight (S), left turn (L), or right turn (R) with the maximum turn rate. Six optimal maneuvers exist: RSR, LSL, RSL, LSR, RLR, LRL. For certain cases, the classification scheme by \citet{Shkel2001DubinsClassif} can determine the optimal maneuver type without explicitly computing the length of all six maneuvers. The Dubins airplane~\cite{Chitsaz2007DubinsAirplane} is the extension of the Dubins car to 3D. The aircraft state $q$ is subject to the dynamics
\begin{equation}
\dot{q}
=
\begin{bmatrix}
\dot{x}\\\dot{y}\\\dot{z}\\\dot{\psi}
\end{bmatrix}
=
\begin{bmatrix}
v_\text{air} \cdot \cos(\gamma) \cdot \cos(\psi) \\
v_\text{air} \cdot \cos(\gamma) \cdot \sin(\psi) \\
v_\text{air} \cdot \sin(\gamma) \\
u_\psi
\end{bmatrix} \; ,
\end{equation}
where $x$, $y$, and $z$ represent the position and $\psi$ the heading. The aircraft airspeed $v_\text{air}$ is assumed to be constant. The two inputs of the system are the path angle (climb and sink angle) $\gamma$ and the turn rate $u_\psi$. The inputs are subject to the constraints
\begin{align}
	\gamma &\in \big[ -\gamma_{\text{max}}, \gamma_{\text{max}}  \big] \; ,
	\label{eqn:PL_climbinlim}\\[00pt]
	u_{\psi} &\in \big[ -\frac{g}{v_\text{air}} \cdot \tan\left(\phi_{\text{max}} \right), \frac{g}{v_\text{air}} \cdot \tan\left(\phi_{\text{max}} \right) \big] \; ,
	\label{eqn:PL_thetalim}
\end{align}
where $\gamma_{\text{max}}$ is the maximum allowed path angle, $g$ the gravitational acceleration, and $\phi_{\text{max}}$ the maximum allowed bank angle. The bank angle limit can also be expressed in terms of the aircraft's minimum turn radius
\begin{equation}
r_{\text{turn}} = \frac{v_\text{air}^2}{\tan \left(\phi_{\text{max}} \right) \cdot g} \; .\label{eqn:PL_rmin}
\end{equation}
The Dubins airplane path is the shortest path under these dynamics between two states in an obstacle-free environment without wind. The calculation method~\cite{Chitsaz2007DubinsAirplane} relies on projecting the start and goal states into a horizontal plane, solving the respective Dubins car problem, and then expanding the path back to three dimensions by computing the climbing angle and adding a climb/sink helix if necessary. The so called low- and high-altitude Dubins airplane paths are calculated optimally. However, no closed-form solution for the medium altitude case exists. To speed up the planning, this paper therefore uses the approximation described in \cite{Schneider2016MT} for the $<5\%$ medium altitude cases that occur in usual planning problems. The error with respect to the optimal solution is bounded, but strictly speaking, not all Dubins airplane paths are optimal. More generally, standard Dubins airplane paths have two main limitations: First, they are only a very rough approximation of the aircraft dynamics because the discontinuities in the turn rate make them dynamically infeasible. \citet{Scheuer1998}, \citet{Richter2013}, or \citet{Askari2015} therefore present the generation of curvature continuous paths. Second, they are only valid in zero-wind. However, \citet{McGee2005} extend them to uniform wind, and \Cref{sec:PL_Planning_dubinswithwind} extends McGee's work towards Dubins airplane paths for non-uniform wind. 

%%%%%%%%%%%%%%%%%%%%%%%%%%%%%%%%%%%%%%%%%%%%%%%%%%%%%%%
\subsection{Implementation and Testing Setup}
%\label{sec:PL_Planning_Implementation}
%%%%%%%%%%%%%%%%%%%%%%%%%%%%%%%%%%%%%%%%%%%%%%%%%%%%%%%

\begin{figure}[htbp]
\centering
\includegraphics[width=1.0\textwidth]{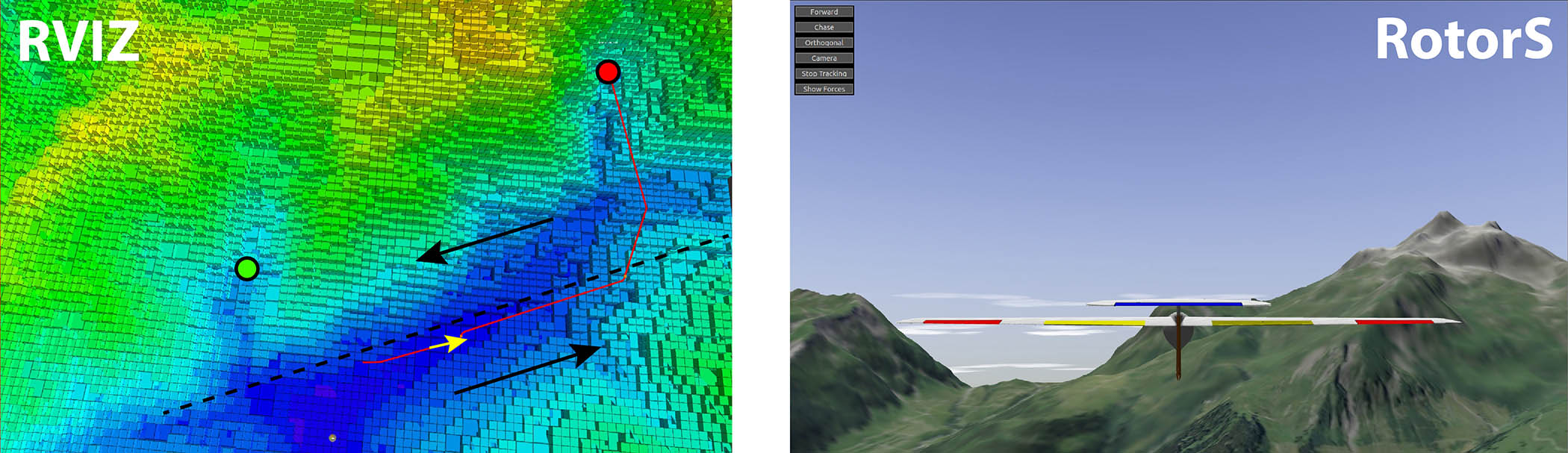}
\caption{The time-optimal planner running in real time in a \ac{HIL} simulation: RVIZ (left) and Gazebo/RotorS~\cite{Furrer2016rotors} (right) visualizations. The aircraft (yellow arrow) avoids the Alpine terrain and leverages favorable winds that, as indicated by the black arrows, blow from start (green circle) to goal (red circle) in the lower section of the valley.}
\label{fig:PL_Planning_hilsim2}
\end{figure}

The path planning framework is implemented in C++ and is deeply integrated with the \acf{ROS}, which provides the interfaces to load the octomap~\cite{Hornung2013Octomap} or \ac{DEM} and the wind field data from the downscaling method described in \cref{sec:PL_WindPrediction}. The planning framework is based on and extends \ac{OMPL}~\cite{OMPL}. A variety of planners, among them RRT*, FMT*, PRM* and the informed planners IRRT* and BIT*, is supported. The path's waypoints are exported to a MAVLink\footnote{\url{http://mavlink.org}} compatible autopilot via Mavros\footnote{\url{http://wiki.ros.org/mavros}}. The whole framework, including generation and processing of its inputs and waypoint outputs, can therefore be tested in a \acl{HIL} simulation using Gazebo/RotorS \cite{Furrer2016rotors} (\cref{fig:PL_Planning_hilsim2}). In the following sections, the planning performance is assessed with an Intel i7-6500U dual-core processor with \unit[16]{GB} RAM using the \ac{OMPL} benchmarking tools. If not mentioned otherwise, each test case is run 100 times with the aircraft and planning parameters in \cref{tab:PL_Planning_experimentsetupparameter}. More details on the implementation and parameters are presented by \citet{Achermann2017MT}.

\begin{table}[htb]
\centering
\caption{The default parameters used for the planner benchmarking.}
\begin{tabular}{l l l}
\toprule
\textbf{Parameter} & \textbf{Variable} & \textbf{Value} \\
\midrule
\\[-1.8ex]
\emph{Planner parameters}\\
Planner && RRT*\\ 
Maximum planning time && \unit[2]{s}/\unit[15]{s} \\ 
Terrain altitude source && 2.5D map (\emph{height checking})\\
Obstacle aware sampling && true \\  
Custom nearest neighbor search && true \\ 
Pre-evaluated height map && false \\ 
Bounding box side length & $s_\text{bb}$ & \unit[30]{m} (all sides)\\ 
Terrain height map resolution & $r_\text{hm}$ & \unit[30]{m}\\ 
Meteo map resolution & $r_\text{mm}$ & \unit[30]{m}\\
Max. Dubins motion length & $d_\text{max}$ & $0.2\cdot\text{map size}$\\[0.8ex]
\emph{Airplane parameters}\\
Minimum turn radius & $r_\text{min}$ &\unit[25]{m}\\ 
Maximum climbing/sinking angle & $\gamma_\text{max}$ & \unit[0.15]{rad}\\ 
Airspeed & $v_\text{air}$ & \unitfrac[9.0]{m}{s}\\
\bottomrule 
\end{tabular}
\label{tab:PL_Planning_experimentsetupparameter}
\end{table}
%+missing: $d_\text{icc}$?

%%%%%%%%%%%%%%%%%%%%%%%%%%%%%%%%%%%%%%%%%%%%%%%%%%%%%%%%%%%%%%%%%
\subsection{Improving Shortest-Distance Path Planning Performance}
\label{sec:PL_Planning_Improvements}
%%%%%%%%%%%%%%%%%%%%%%%%%%%%%%%%%%%%%%%%%%%%%%%%%%%%%%%%%%%%%%%%%

\begin{figure}[htbp]
\centering
\subfloat[\emph{Dom} mountain range scenario]{\includegraphics[width=0.38\textwidth]{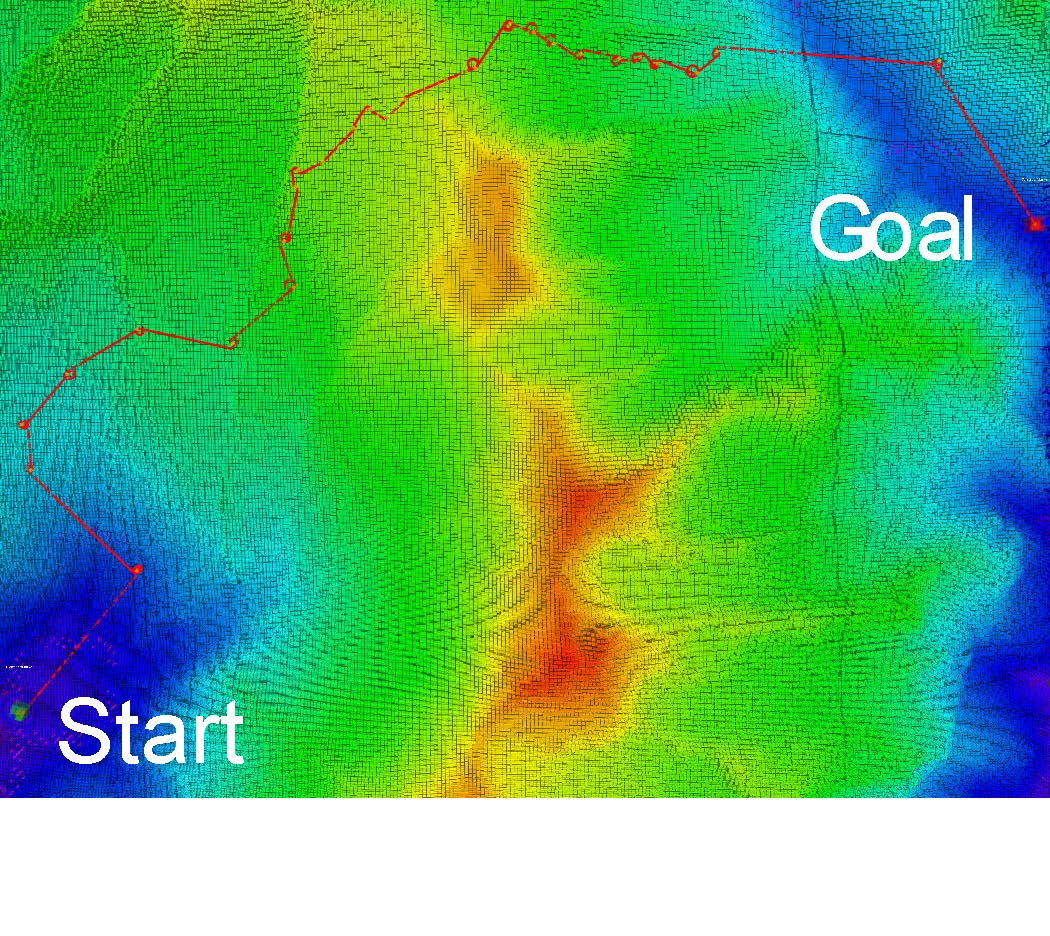}\label{fig:PL_Planning_Impr_TestcasesDomA}}
\hfill
\subfloat[\emph{Dom} mountain range scenario]{\includegraphics[width=0.57\textwidth]{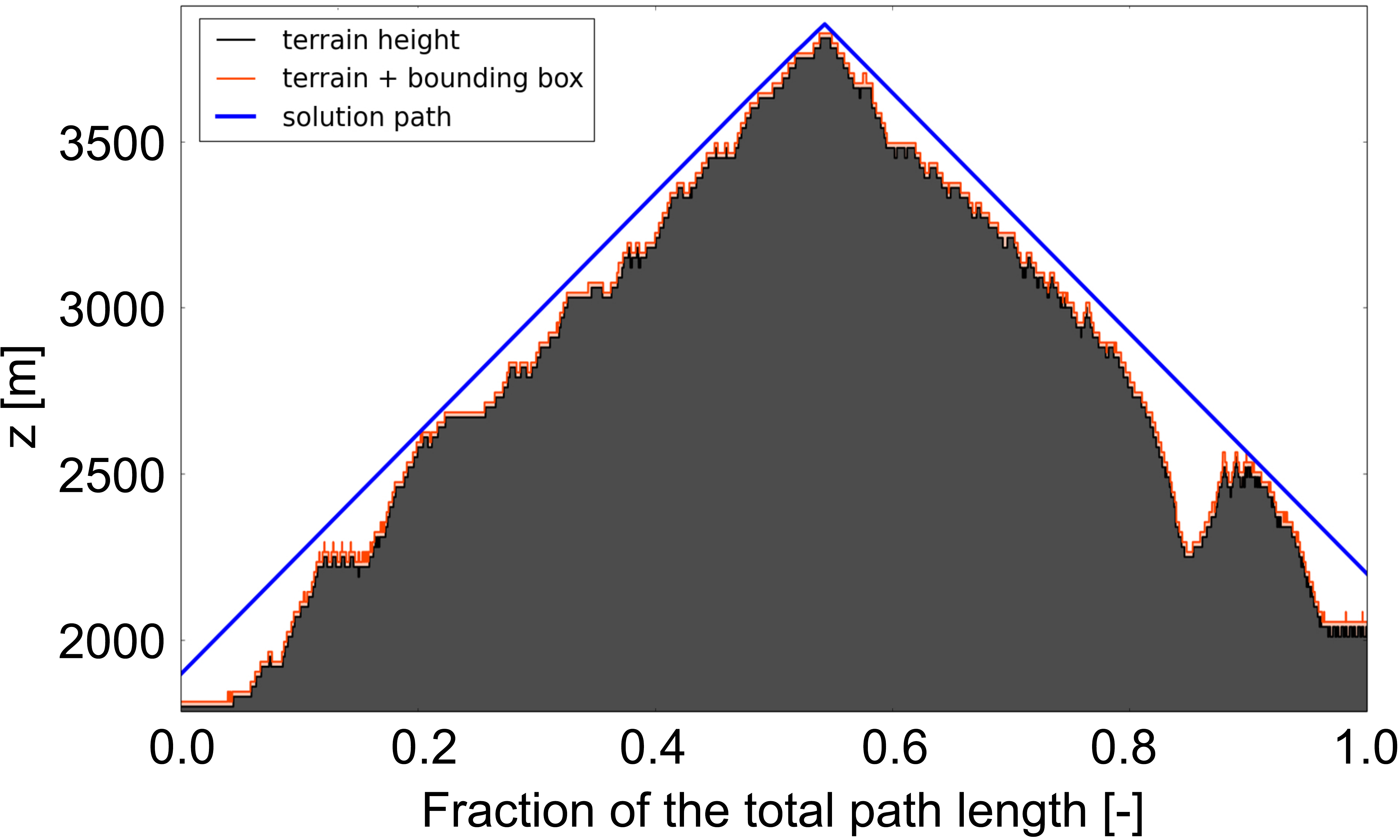}\label{fig:PL_Planning_Impr_TestcasesDomB}}

\subfloat[\textit{Kandertal Low} scenario]{\includegraphics[width=0.38\textwidth]{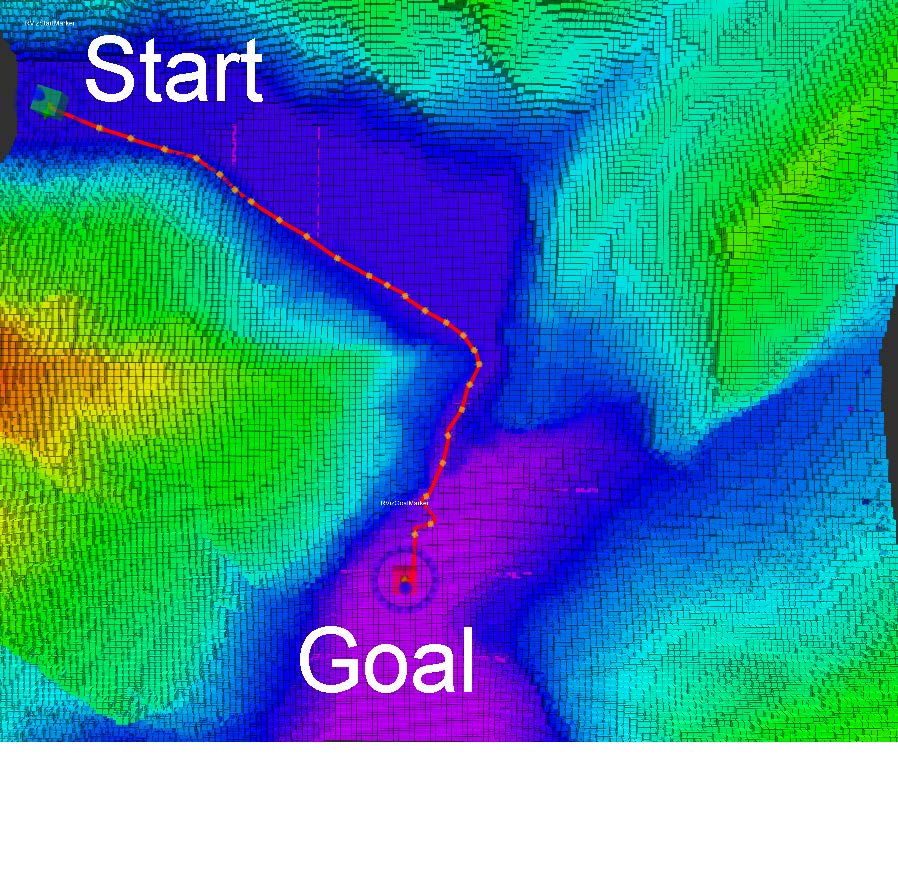}\label{fig:PL_Planning_Impr_TestcasesKandertalA}}
\hfill
\subfloat[\textit{Kandertal Low} scenario]{\includegraphics[width=0.57\textwidth]{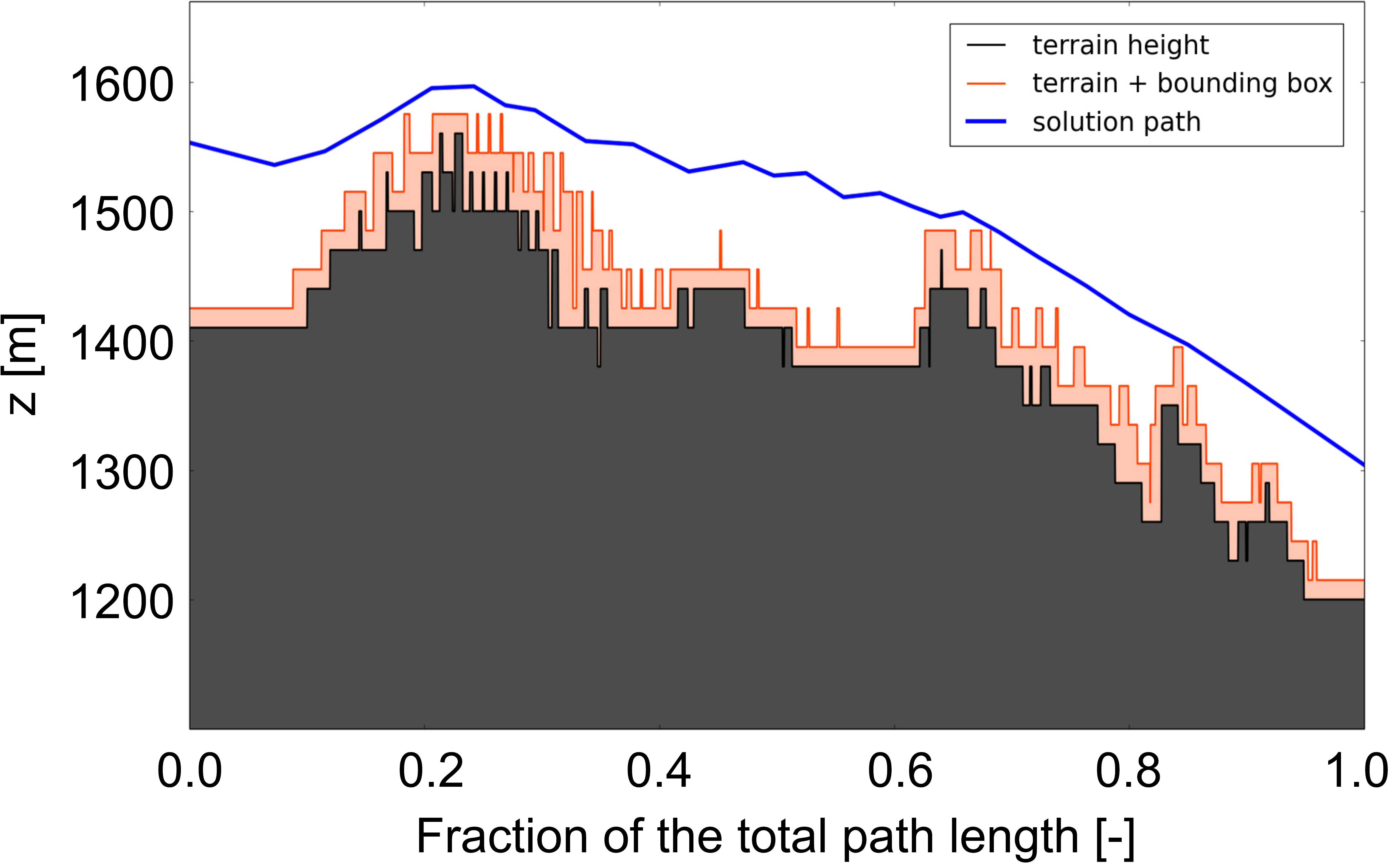}\label{fig:PL_Planning_Impr_TestcasesKandertalB}}
\caption{The test scenarios with the optimal path after 2 hours of planning. Left: Terrain height (blue equals low terrain, red high terrain) and the path (red). Right: Terrain and path along the total path length fraction.}
\label{fig:PL_Planning_Impr_Testcases}
\end{figure}

This section contributes computational performance improvements for shor\-test-dis\-tance aircraft motion planning with RRT*-like methods. The combined performance improvement is analyzed in \cref{sec:PL_Planning_Improvements_Results}. Two different test cases in Alpine terrain are used: The \emph{Dom} mountain range %\footnote{The highest mountain of Switzerland with its complete base lying on Swiss territory}
(\cref{fig:PL_Planning_Impr_TestcasesDomA}) with start and goal on opposite sides of the mountain and the \emph{Kandertal Low} test case (\cref{fig:PL_Planning_Impr_TestcasesKandertalA}).

%%%%%%%%%%%%%%%%%%%%%%%%%%%%%%%%%%%%%%%%%%%%%%%%%%%%%%%%%%%%%%%%%
\subsubsection{An Improved Informed Subset for a Dubins Airplane}
\label{sec:PL_Planning_Improvements_InformedSubset}
%%%%%%%%%%%%%%%%%%%%%%%%%%%%%%%%%%%%%%%%%%%%%%%%%%%%%%%%%%%%%%%%%

The Euclidean distance is always shorter than the Dubins aircraft~\cite{Chitsaz2007DubinsAirplane} path between two states. It is thus a valid heuristic $h(q)$. However, due to the constrained path angle (and thus climb and sink rate) the aircraft has to fly a helix or lengthen its horizontal path if the goal altitude is too high or low. This fact is used to clip the informed sampling ellipse in $z$-direction. Defining
\begin{align}
d_{\text{eucl}}(q_1, q_2) &= \Vert q_1-q_2 \Vert_{_2}
\label{eqn:PL_Planning_eucldist}\\%[10pt]
d_\text{climb}(q_1, q_2) &= \frac{|q_1.z -q_2.z|}{\sin\left|\gamma_{\text{max}} \right|} \; ,
\label{eqn:PL_Planning_dubapproxdist}
\end{align}
where $q_1,q_2\in Q_\text{free}$, then the resulting cost heuristic $h(q)$ for the airplane path length from $q_\text{start}$ through $q$ to $q_\text{goal}$ is
\begin{align}
h(q) &= \max \big( d_{\text{eucl}}(q_\text{start}, q), d_\text{climb}(q_\text{start}, q) \big) \nonumber \\
& + \max \big( d_{\text{eucl}}(q, q_\text{goal}), d_\text{climb}(q, q_\text{goal}) \big) \; .
\label{eqn:Planning_dubheuristic}
\end{align}

\begin{figure}[htbp]
\subfloat{\includegraphics[width=0.48\textwidth]{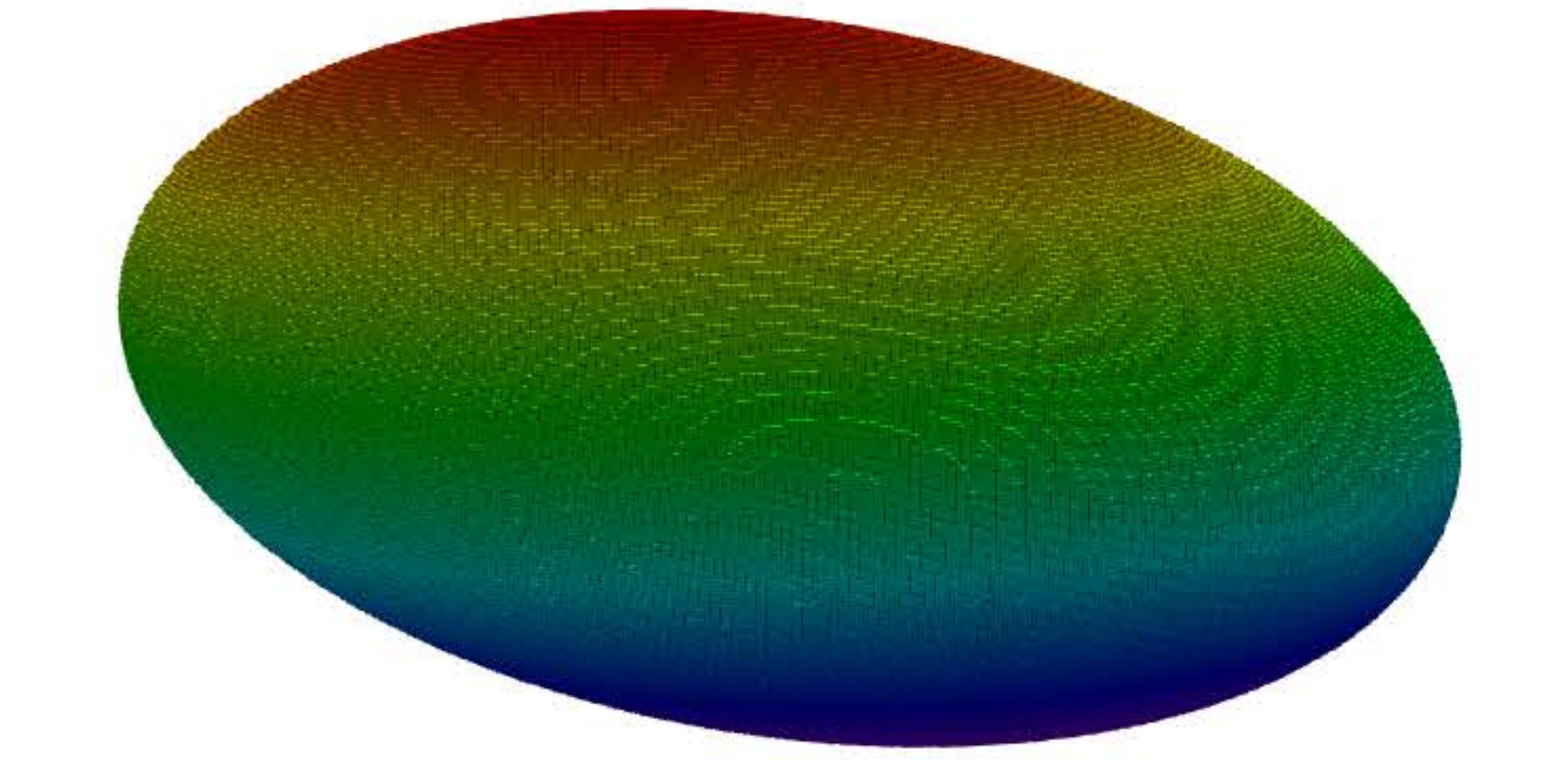}}
\hfill
\subfloat{\includegraphics[width=0.48\textwidth]{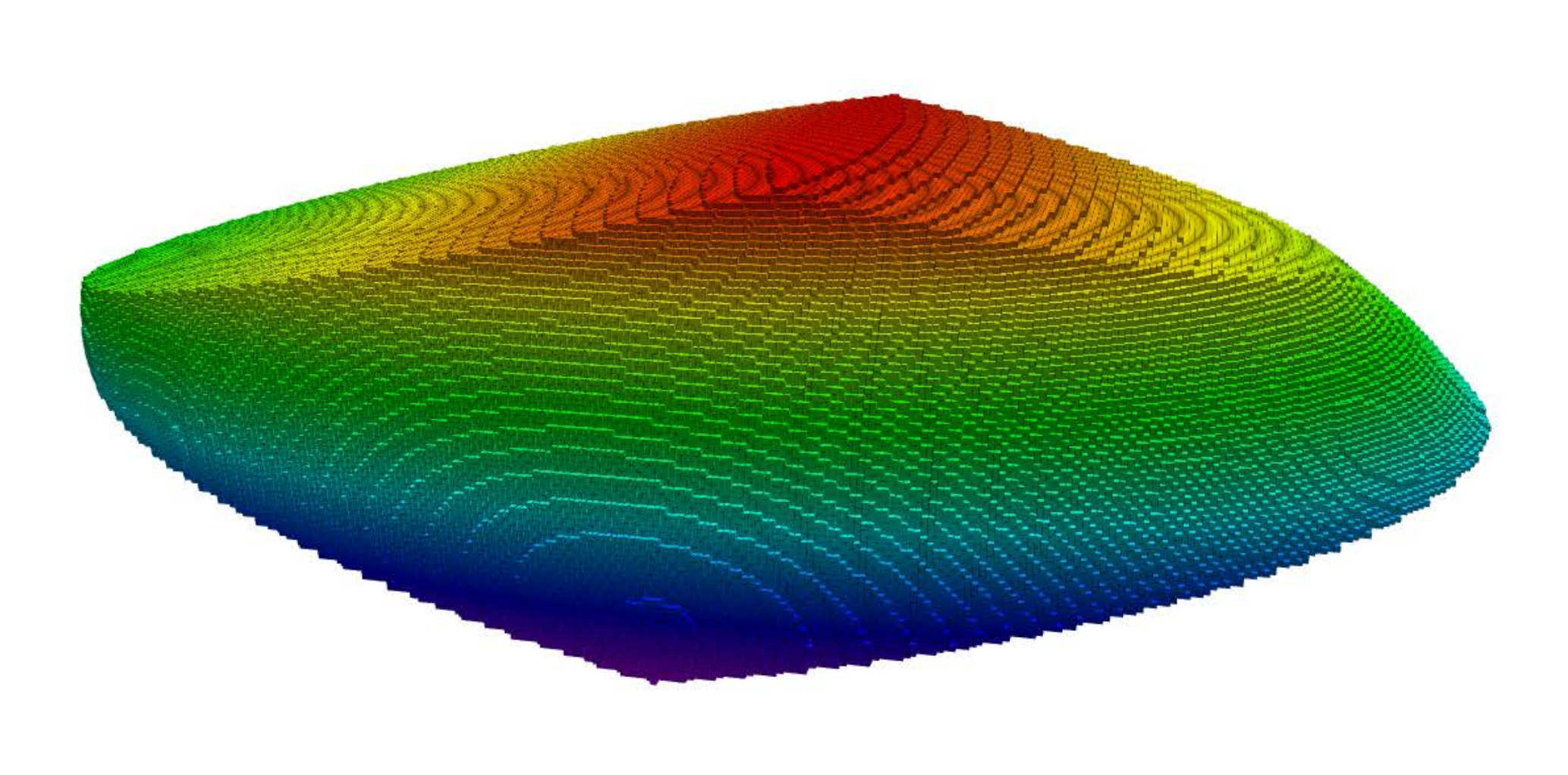}}
\caption{The left ellipsoid visualizes $Q_\text{inf}$ if $h(q)$ is the Euclidean distance only. The right ellipse is the heuristic of \cref{eqn:Planning_dubheuristic}, which is capped in positive and negative $z$-direction. Here, a maximum path angle of $\gamma_\text{max}=\unit[0.25]{rad}$ is used.}
\label{fig:PL_Planning_eucldubinformed}
\end{figure}

This formulation of $h(q)$ results in the informed subset $Q_\text{inf}$ shown in \cref{fig:PL_Planning_eucldubinformed}. The size of $Q_\text{inf}$ is reduced as a function of $\gamma_\text{max}$. Of course only informed planners (i.e. IRRT* and BIT* in this framework) benefit from this improved heuristic.

%%%%%%%%%%%%%%%%%%%%%%%%%%%%%%%%%%%%%%%%%%%%%%%%%%%%%%%%%%%%%%%%%
\subsubsection{Collision Checking}
\label{sec:PL_Planning_Improvements_CollisionChecking}
%%%%%%%%%%%%%%%%%%%%%%%%%%%%%%%%%%%%%%%%%%%%%%%%%%%%%%%%%%%%%%%%%

The \code{MotionValid} function in \cref{alg:PL_Planning_RRTstar} checks whether a motion lies in $Q_{\text{free}}$. In addition to the initial and final states $q_1,q_2$, all intermittent states are checked by discretizing the motion with the user-specified intermediate collision check distance $d_\text{icc}$. This distance should be small enough so that all collisions are detected but large enough to not slow down the motion checking excessively. Our framwork implements both the \ac{FCL}~\cite{Pan2012FCL} and 2.5D map height checking. For flight at medium to high altitude in outdoor environments, the full 3D collision checking capability of \ac{FCL} is however not needed. The overhanging geometry of certain obstacles (bridges, power lines, trees) can simply be modeled as 2.5D features to speed up the collision checking.

%\paragraph{Contributions}

\begin{figure}[htbp]
\centering
\tikzset{font={\fontsize{12pt}{12}\selectfont}}
\begin{tikzpicture}[auto, scale=0.7, every node/.style={transform shape}]
%orange/red markers
\draw [fill=orange,draw=none] (3.5,2.5) rectangle (5.5,4.5);
\draw [fill=red,draw=none] (4.0,3.0) rectangle (5.0,4.0);
\draw [fill=red,draw=none] (13.0,3.0) rectangle (14.0,4.0);
%grey markers. Left plot.
\draw [fill=lightgray,draw=none] (1.0,0.0) rectangle (3.0,1.0);
\draw [fill=lightgray,draw=none] (1.0,1.0) rectangle (3.0,2.0);
\draw [fill=lightgray,draw=none] (0.0,2.0) rectangle (3.0,3.0);
\draw [fill=lightgray,draw=none] (0.0,3.0) rectangle (2.0,4.0);
\draw [fill=lightgray,draw=none] (0.0,4.0) rectangle (2.0,5.0);
%grey markers. Right plot
\draw [fill=lightgray,draw=none] (10.0,0.0) rectangle (11.0,5.0);

\draw[step=1cm,gray,very thin] (0.0,0.0) grid (6	.0,5.0);
\draw[thick,->] (0,0) -- (6.3,0.0) node[anchor=north] {};
\draw[thick,->] (0,0) -- (0,5.3) node[anchor=south] {y-axis};
\node[draw=none] at (5.8,-0.3) {x-axis};
\node[draw=none] at (0.5,0.5) {10};
\node[draw=none] at (1.5,0.5) {12};
\node[draw=none] at (2.5,0.5) {13};
\node[draw=none] at (3.5,0.5) {8};
\node[draw=none] at (4.5,0.5) {6};
\node[draw=none] at (5.5,0.5) {4};

\node[draw=none] at (0.5,1.5) {19};
\node[draw=none] (v2) at (1.5,1.5) {11};
\node[draw=none] (v4) at (2.5,1.5) {15};
\node[draw=none] at (3.5,1.5) {12};
\node[draw=none] at (4.5,1.5) {10};
\node[draw=none] at (5.5,1.5) {8};

\node[draw=none] at (0.5,2.5) {25};
\node[draw=none] at (1.5,2.5) {20};
\node[draw=none] at (2.5,2.5) {18};
\node[draw=none] at (3.5,2.5) {15};
\node[draw=none] at (4.5,2.5) {13};
\node[draw=none] at (5.5,2.5) {11};

\node[draw=none] at (0.5,3.5) {30};
\node[draw=none] at (1.5,3.5) {18};
\node[draw=none] at (2.5,3.5) {15};
\node[draw=none] at (3.5,3.5) {10};
\node[draw=none] at (4.5,3.5) {8};
\node[draw=none] at (5.5,3.5) {6};

\node[draw=none] (v1) at (0.5,4.5) {28};
\node[draw=none] (v3) at (1.5,4.5) {16};
\node[draw=none] at (2.5,4.5) {13};
\node[draw=none] at (3.5,4.5) {8};
\node[draw=none] at (4.5,4.5) {6};
\node[draw=none] at (5.5,4.5) {4};

\draw[step=1cm,gray,very thin] (9.0,0.0) grid (15.0,5.0);
\draw[thick,->] (9,0) -- (15	.3,0) node[anchor=north] {};
\draw[thick,->] (9,0) -- (9,5.3) node[anchor=south] {y-axis};
\node[draw=none] at (14.8,-0.3) {x-axis};

\node[draw=none] at (9.5,0.5) {19};
\node[draw=none] at (10.5,0.5) {19};
\node[draw=none] at (11.5,0.5) {15};
\node[draw=none] at (12.5,0.5) {15};
\node[draw=none] at (13.5,0.5) {12};
\node[draw=none] at (14.5,0.5) {10};

\node[draw=none] at (9.5,1.5) {25};
\node[draw=none] at (10.5,1.5) {25};
\node[draw=none] at (11.5,1.5) {20};
\node[draw=none] at (12.5,1.5) {18};
\node[draw=none] at (13.5,1.5) {15};
\node[draw=none] at (14.5,1.5) {13};

\node[draw=none] at (9.5,2.5) {30};
\node[draw=none] at (10.5,2.5) {30};
\node[draw=none] at (11.5,2.5) {20};
\node[draw=none] at (12.5,2.5) {18};
\node[draw=none] at (13.5,2.5) {15};
\node[draw=none] at (14.5,2.5) {13};

\node[draw=none] at (9.5,3.5) {30};
\node[draw=none] at (10.5,3.5) {30};
\node[draw=none] at (11.5,3.5) {20};
\node[draw=none] at (12.5,3.5) {18};
\node[draw=none] at (13.5,3.5) {15};
\node[draw=none] at (14.5,3.5) {13};

\node[draw=none] at (9.5,4.5) {30};
\node[draw=none] at (10.5,4.5) {30};
\node[draw=none] at (11.5,4.5) {18};
\node[draw=none] at (12.5,4.5) {15};
\node[draw=none] at (13.5,4.5) {10};
\node[draw=none] at (14.5,4.5) {8};

\draw[->, ultra thick] (6.5, 2.0) -- (8.5, 2.0);

\node[draw=none] at (3.0,-1.0) {a) Original Height Map};
\node[draw=none] at (12.0,-1.0) {b) Pre-Evaluated Height Map};

%Path: Left
\draw[red,thick]  plot[smooth, tension=.7] coordinates {(1.1,5.0) (1.9,0)}; %main
\draw[red,dashed,thick]  plot[smooth, tension=.7] coordinates {(0.6,5.0) (1.4,0)}; %left
\draw[red,dashed,thick]  plot[smooth, tension=.7] coordinates {(1.6,5.0) (2.4,0)}; %right
%Path: Right
\draw[red,thick]  plot[smooth, tension=.7] coordinates {(10.1,5.0) (10.9,0)}; %main

\end{tikzpicture}
\caption{Left: Standard collision checking with exemplary cell (red) and relevant collision area (orange). Right: Pre-evaluated collision checking with corresponding pre-evaluated cell (red). The numbers represent the terrain height $h_\text{ter}$. The figure assumes $s_\text{bb}=r_\text{hm}$. An exemplary path (red solid line) is shown together with the bounding box swath (red dashed lines) and the cells that need to be checked (grey).}
\label{fig:PL_Planning_preprocessheightmap}
\end{figure}
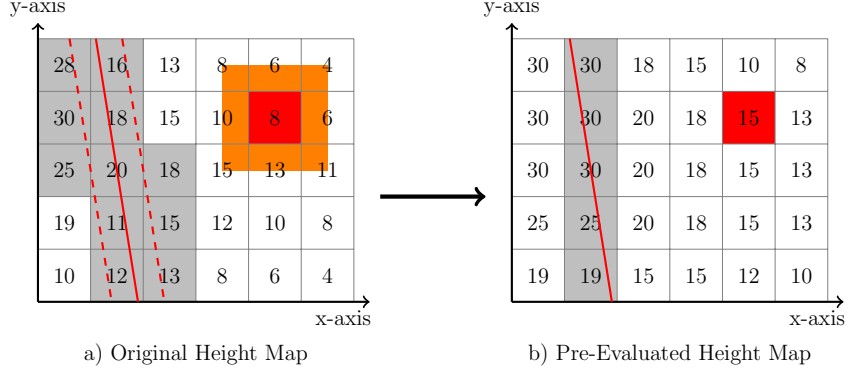

This paper therefore presents a pre-evaluation method for 2.5D-map based height checking. Standard 2.5D height checking defines the terrain altitude $h_\text{ter}(x,y)$ and considers the space below $h_\text{ter}$ obstacle space $Q_\text{obs}$. The map is then collision-checked against the bounding box of the aerial vehicle. The box size is chosen to fully contain the vehicle in all attitudes. A safety margin for the minimum desired distance to terrain is added. In our implementation, the 2.5D-map is internally always stored at the bounding box resolution. This approach is conservative but efficient because a maximum of four cells of the 2.5D map have to be checked. As shown in \cref{fig:PL_Planning_preprocessheightmap}a, the bounding box can generally intersect with the eight surrounding cells of the current aircraft position. The pre-evaluation calculates the relevant collision area for a cell with center $x_c,y_c$ and any arbitrary aircraft position inside that cell
\begin{align}
x_\text{c} - 0.5 \left( r_\text{hm} + s_\text{bb}\right) & \leq x \leq x_\text{c} + 0.5 \left( r_\text{hm} + s_\text{bb}\right) \\
y_\text{c} - 0.5 \left( r_\text{hm} + s_\text{bb}\right) & \leq y \leq y_\text{c} + 0.5 \left( r_\text{hm} + s_\text{bb}\right) \; .
\end{align}
Here, $r_\text{hm}$ is the height map resolution and $s_\text{bb}$ is the bounding box side length. The largest $h_\text{ter}$ in the relevant area is determined and filled into the corresponding cell of the pre-processed height map (\cref{fig:PL_Planning_preprocessheightmap}b). As a result, instead of checking four cells, only the single cell the aircraft actually is in has to be checked. %The computational complexity is also independent of the ratio $\nicefrac{s_\text{bb}}{r_\text{hm}}$. 
The downside is that the pre-evaluated collision checking is more conservative, i.e. it effectively increases the bounding box size with a \emph{virtual} safety margin $\Delta_\text{sm} \leq r_\text{hm}$.

\paragraph{Preliminary results}

\begin{figure}[htbp]
\centering
\subfloat[Number of iterations]{\includegraphics[width=0.48\textwidth]{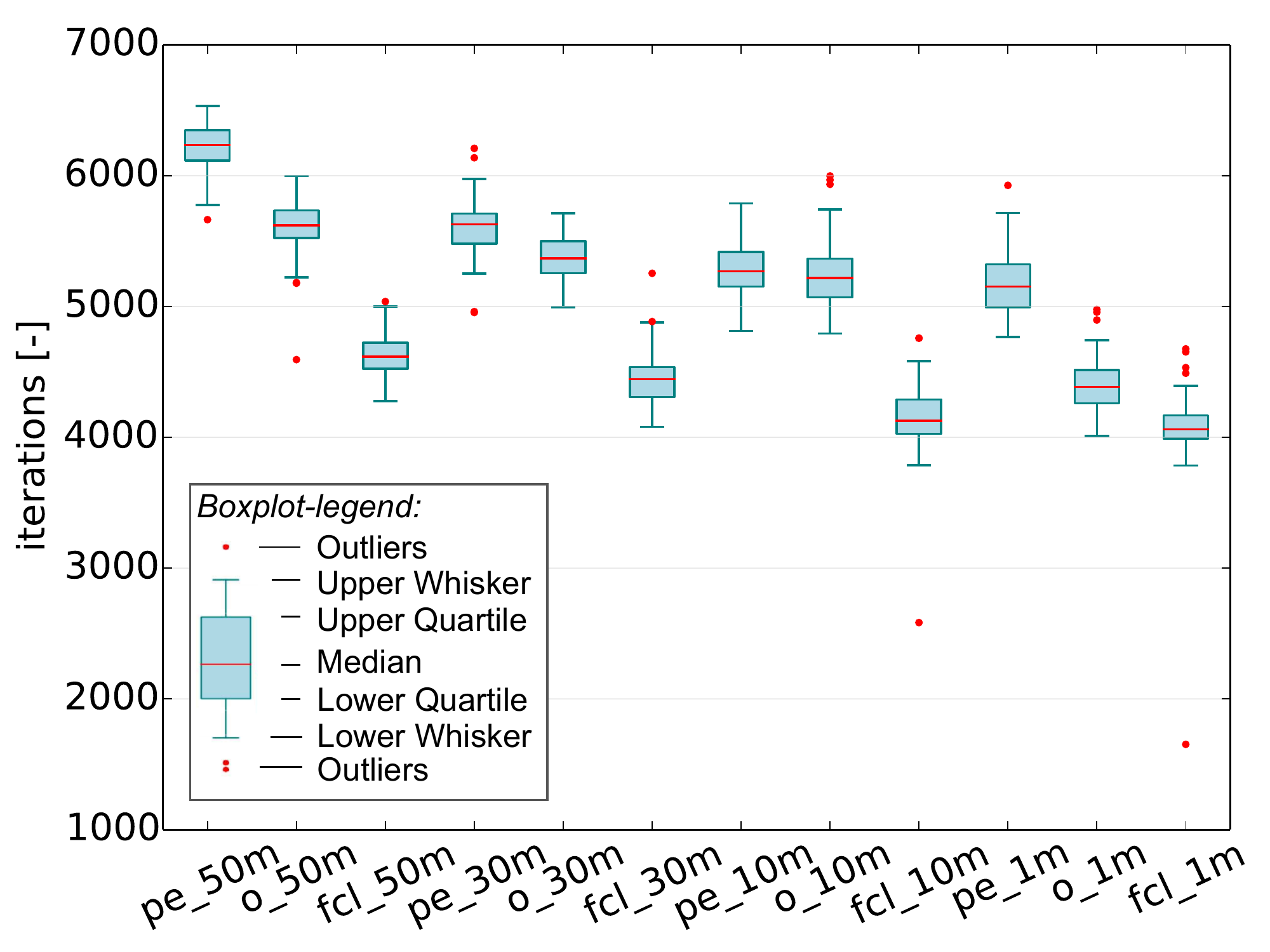}\label{fig:PL_Planning_Impr_CCIterations}}
\hfill
\subfloat[Path cost in meters]{\includegraphics[width=0.48\textwidth]{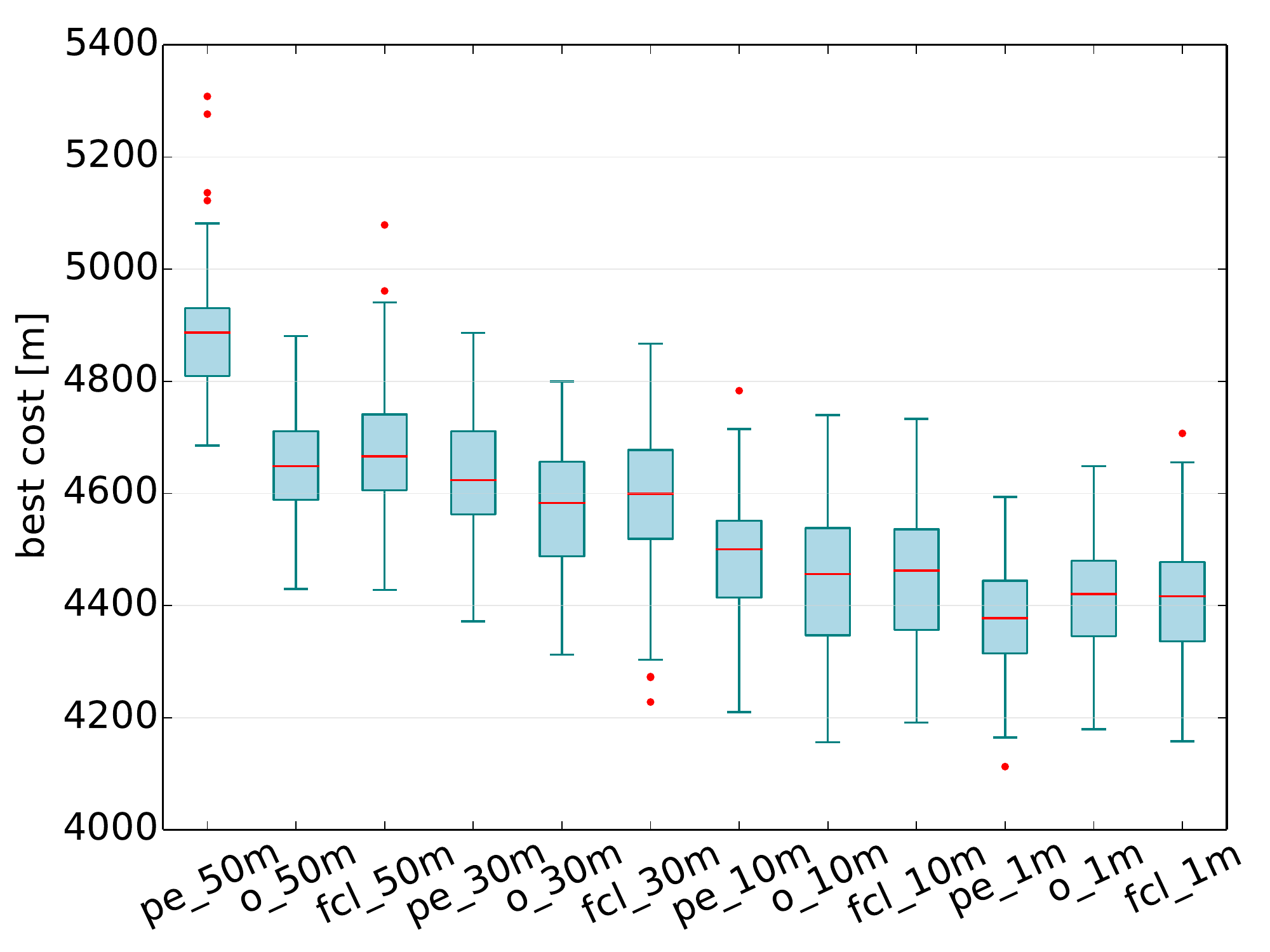}\label{fig:PL_Planning_Impr_CCPathCost}}
\caption{Comparison of original 2.5D-map (o\_), pre-evaluated 2.5D-map (pe\_) and \ac{FCL} (fcl\_) height checking with respect to number of iterations and path cost after \unit[2]{s} of planning for $r_\text{hm}=[\unit[1]{m},\unit[50]{m}]$ and $s_\text{bb}=\unit[30]{m}$.}
\label{fig:PL_Planning_Impr_CC}
\end{figure}

Whereas \ac{FCL}-based collision checking requires 20--30\% of the total planning time, both the original and pre-evaluated height map collision checking require $\leq \unit[0.1]{\%}$ of the planning time for $r_\text{hm}\approx s_\text{bb}$. \Cref{fig:PL_Planning_Impr_CCIterations} plots the number of completed iterations after \unit[2]{s} of planning in the \emph{Kandertal Low} scenario. \ac{FCL} clearly completes the lowest number of iterations. The pre-evaluated height map checking is always faster than the original approach because it only needs to check one cell per planning iteration. The difference is most significant around \unit[1]{m} map resolution. The result of the increased number of planning iterations on the path cost is shown in \cref{fig:PL_Planning_Impr_CCPathCost}. Interestingly, the pre-evaluation only yields better results for $r_\text{hm}=\unit[1]{m}$. Two separate reasons can be identified: First, the pre-evaluation \emph{virtually} increases the size of obstacles such that the path around such obstacles \emph{has} to be longer. While this might not be desired, it is no disadvantage per se because it also makes the path more conservative. Second, the increased obstacle size also leads to more of the sampled states being invalid such that, despite the same number of planning iterations, less valid states can be used for path planning. The overall conclusion is therefore that the pre-evaluated height map should only be used if the height map resolution is small (ca. $r_\text{hm}<\frac{1}{30}s_\text{bb}$) compared to the bounding box. 

%%%%%%%%%%%%%%%%%%%%%%%%%%%%%%%%%%%%%%%%%%%%%%%%%%%%%%%%%%%%%%%%%
%%%%%%%%%%%%%%%%%%%%%%%%%%%%%%%%%%%%%%%%%%%%%%%%%%%%%%%%%%%%%%%%%
\subsubsection{Obstacle-Aware Sampling}
\label{sec:PL_Planning_Improvements_Sampling}
%%%%%%%%%%%%%%%%%%%%%%%%%%%%%%%%%%%%%%%%%%%%%%%%%%%%%%%%%%%%%%%%%
%%%%%%%%%%%%%%%%%%%%%%%%%%%%%%%%%%%%%%%%%%%%%%%%%%%%%%%%%%%%%%%%%

In Euclidean space, an optimal path generally sticks as close as possible to the straight line path and passes obstacles at small distance. As a result, samples close to obstacles have a higher likelihood to improve the solution path than samples in large open space. \citet{Amato1998} presented the idea of obstacle-aware sampling for generic configuration spaces. In this paper we extend Amato's work by using the specific structure of the aircraft path planning problem, i.e. the fact that start and goal pose are known and the obstacles are represented by a 2.5D map. Let $z_\text{s,g}^\text{min}=\min(z_\text{start},z_\text{goal})$ and $z_\text{s,g}^\text{max}=\max(z_\text{start},z_\text{goal})$. Then for Dubins airplane shortest-path planning in a 2.5D map, the optimal path always stays above the plane $z=\smash{z_\text{s,g}^\text{min}}$. The algorithm for obstacle-aware sampling is:
\begin{enumerate}
\item Draw a uniform sample $q=(x,y,z,\psi)$ from the state space $Q$ (or from the state $Q_\text{inf}$ in the informed case).
\item If $z<z_\text{s,g}^\text{min}$, resample the $z$-component from a uniform distribution with the boundaries $[z_\text{s,g}^\text{min}, z_\text{s,g}^\text{max}]$.
\item If $q\in Q_\text{obs}$, shift the state $q$ in positive $z$-direction with $\text{dz}_\text{shift}$ until $q\in Q_\text{free}$. If $q$ is outside the state space $Q$ (for the uninformed case) or the informed subspace $Q_\text{inf}$ (for the informed case), return to Step 1.  
\end{enumerate}

\begin{figure}[htbp]
\begin{minipage}[t]{0.48\textwidth}
\includegraphics[width = \textwidth]{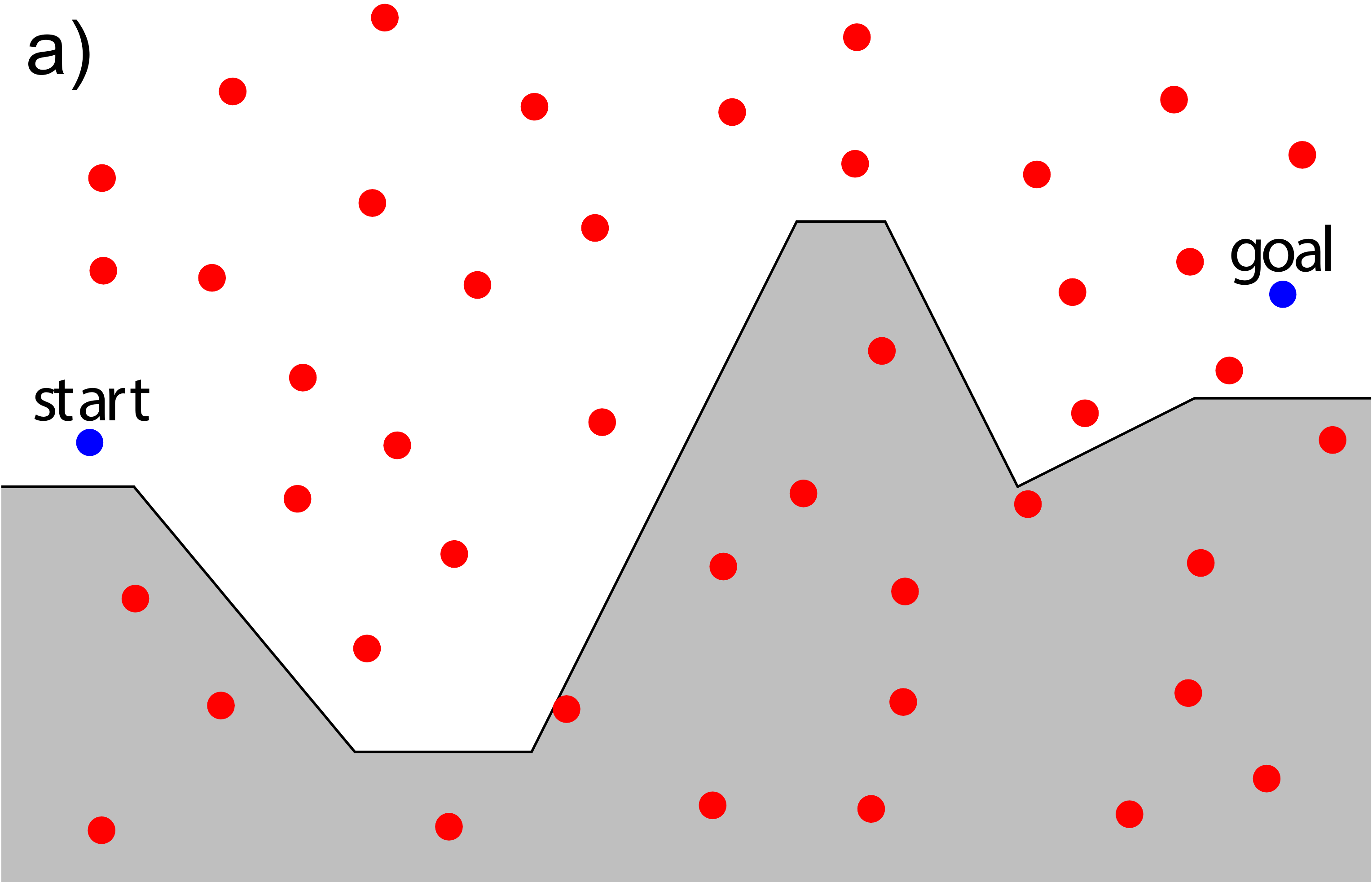}
\end{minipage}
\hfill
\begin{minipage}[t]{0.48\textwidth}
\includegraphics[width = \textwidth]{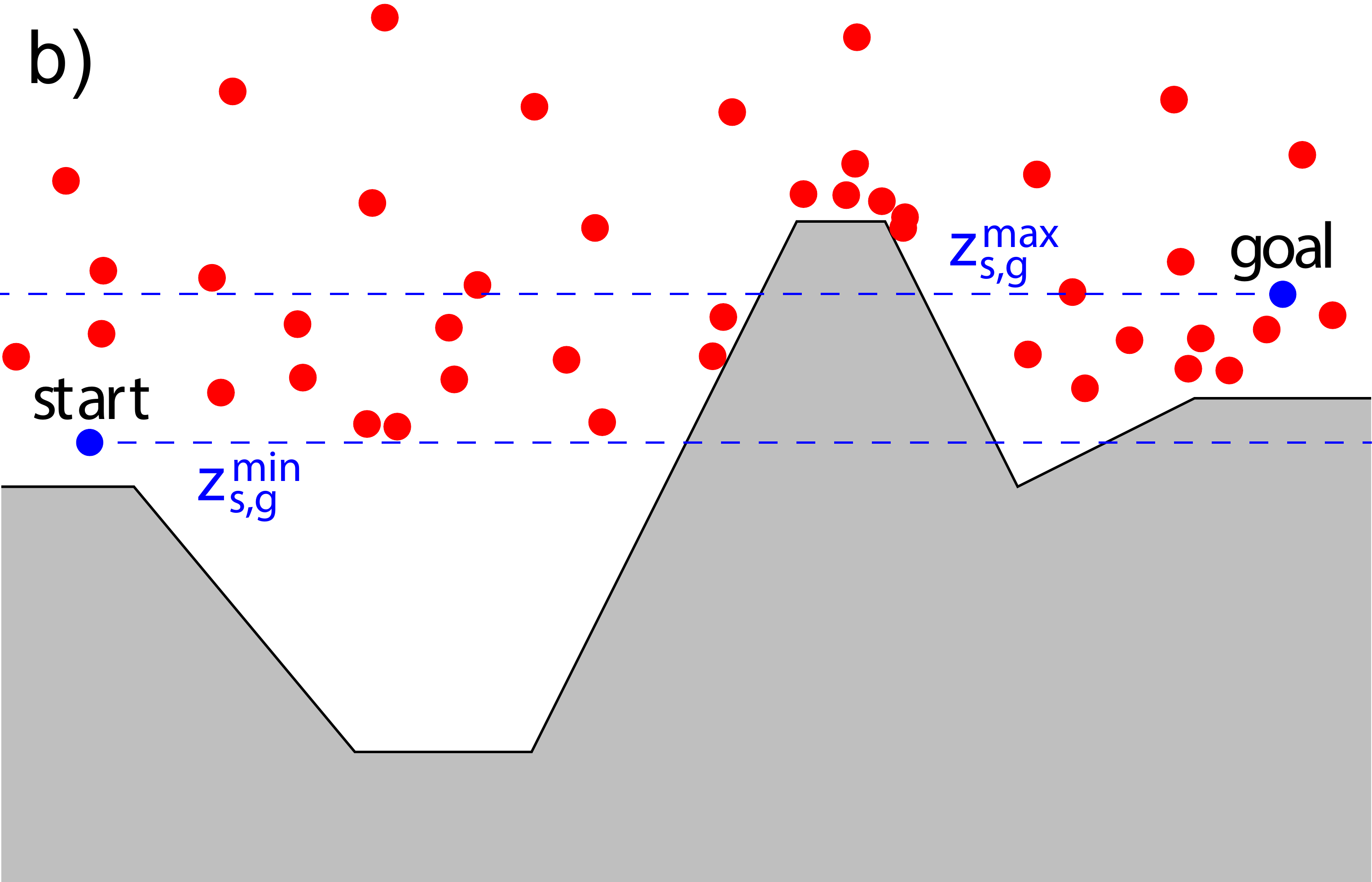}
\end{minipage}
\caption{Obstacle unaware (left) and aware (right) sampling. The obstacle-unaware samples are generated over the full map and also inside obstacles. Applying the obstacle aware shifting algorithm results in no states inside obstacles and more states close to terrain.}
\label{fig:PL_Planning_awarecomparedtounaware}
\end{figure}

This sampling approach avoids samples below $z_\text{s,g}^\text{min}$ which cannot improve the path cost. It increases the sampling density in the regions which contribute most to improving the path cost, i.e. the band between $z_\text{s,g}^\text{min}$ and $z_\text{s,g}^\text{max}$, or if this band is fully or partially occupied by terrain, the band with height $\text{dz}_\text{shift}$ above the terrain. For example, in \cref{fig:PL_Planning_awarecomparedtounaware}b there are no samples in the valley but many samples at the top of the hill crest where the optimal path from $q_\text{start}$ to $q_\text{goal}$ would pass.

%\begin{algorithm}
%\caption{Informed Obstacle Aware Sampling: While shifting the sample up in $z$-direction it needs to be checked if the sample is still in $C_\text{inf}$. If this is not the case a new sample is drawn and shifted again if necessary.}
%\label{alg:PL_obstacleawaresampling}
%\begin{algorithmic} [1]
%\Procedure{sample}{$c_\text{max}, \text{dz}_\text{shift}, z_\text{S,G}^\text{min}, z_\text{S,G}^\text{max}$}
%\While {$\left(\text{foundSample} == \text{false} \right)$}
%  \State $q_r \gets \textsc{sampleInformedRandomState} \left(c_\text{max}\right)$
%  \If {$\left(q_r.z < z_\text{S,G}^\text{min}\right)$}
%    \State $q_r.z \gets \textsc{resampleZ} \left(z_\text{S,G}^\text{min}, z_\text{S,G}^\text{max}\right)$
%  \EndIf
%  
%  \While {$\left(\textsc{isValid} \left(q_r \right) == \text{false} \textbf{ and } \textsc{isInCinf} \left(q_r \right) == \text{true}\right)$}
%    \State $q_r \gets \textsc{shiftState} \left(q_r, \text{dz}_\text{shift}\right)$
%    \State $\text{foundSample} \gets \textsc{isValid} \left(q_r \right) \textbf{ and } \textsc{isInCinf} \left(q_r \right)$
%  \EndWhile
%\EndWhile
%\State \Return $q_r$
%\EndProcedure
%\end{algorithmic}
%\end{algorithm}

\paragraph{Preliminary results}

\begin{figure}[htbp]
\centering
\subfloat[Number of states in the motion tree.]{\includegraphics[width=0.48\textwidth]{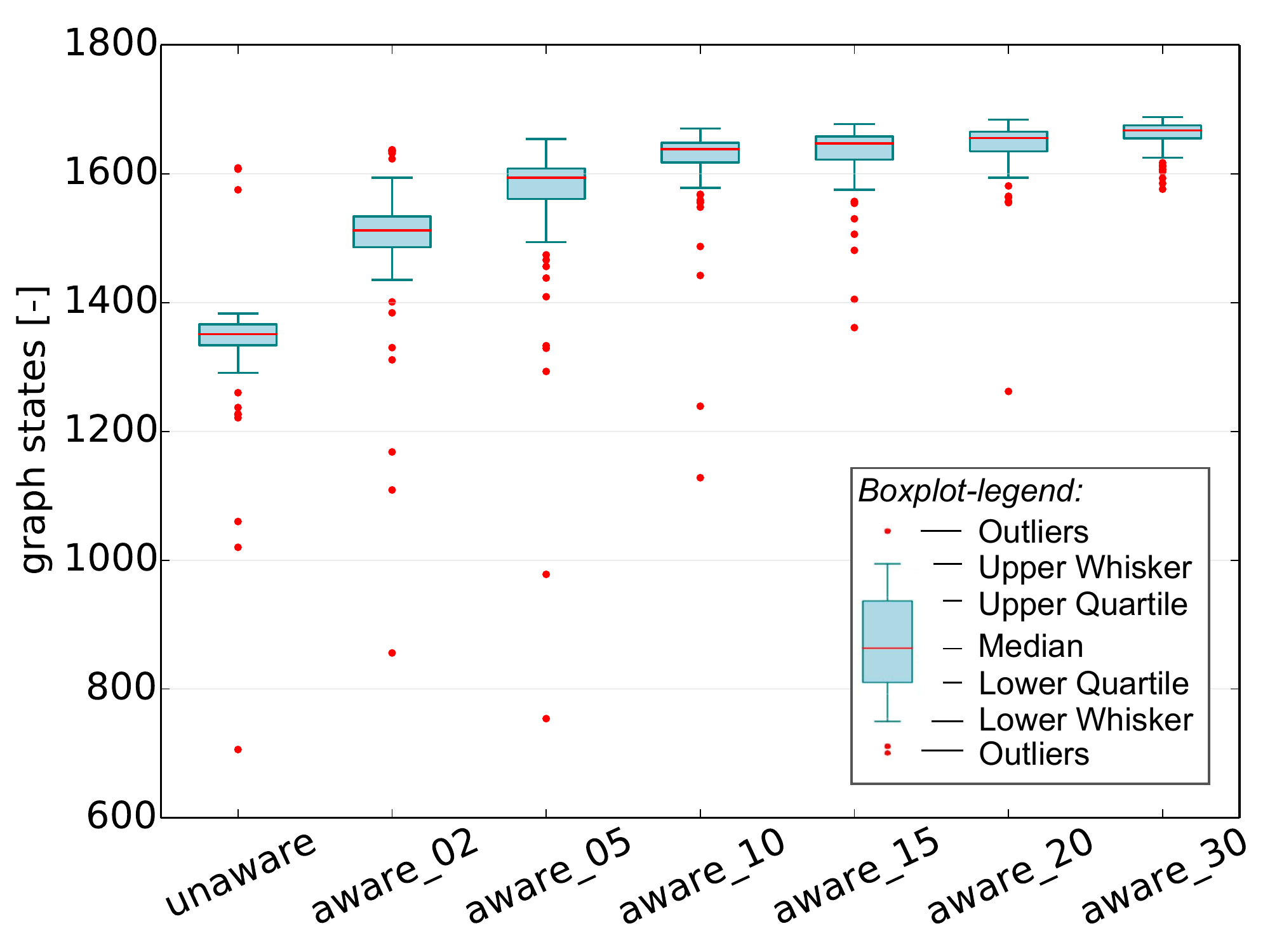}\label{fig:PL_Planning_samplingdomstates}}
\hfill
\subfloat[Path cost in meters.]{\includegraphics[width=0.48\textwidth]{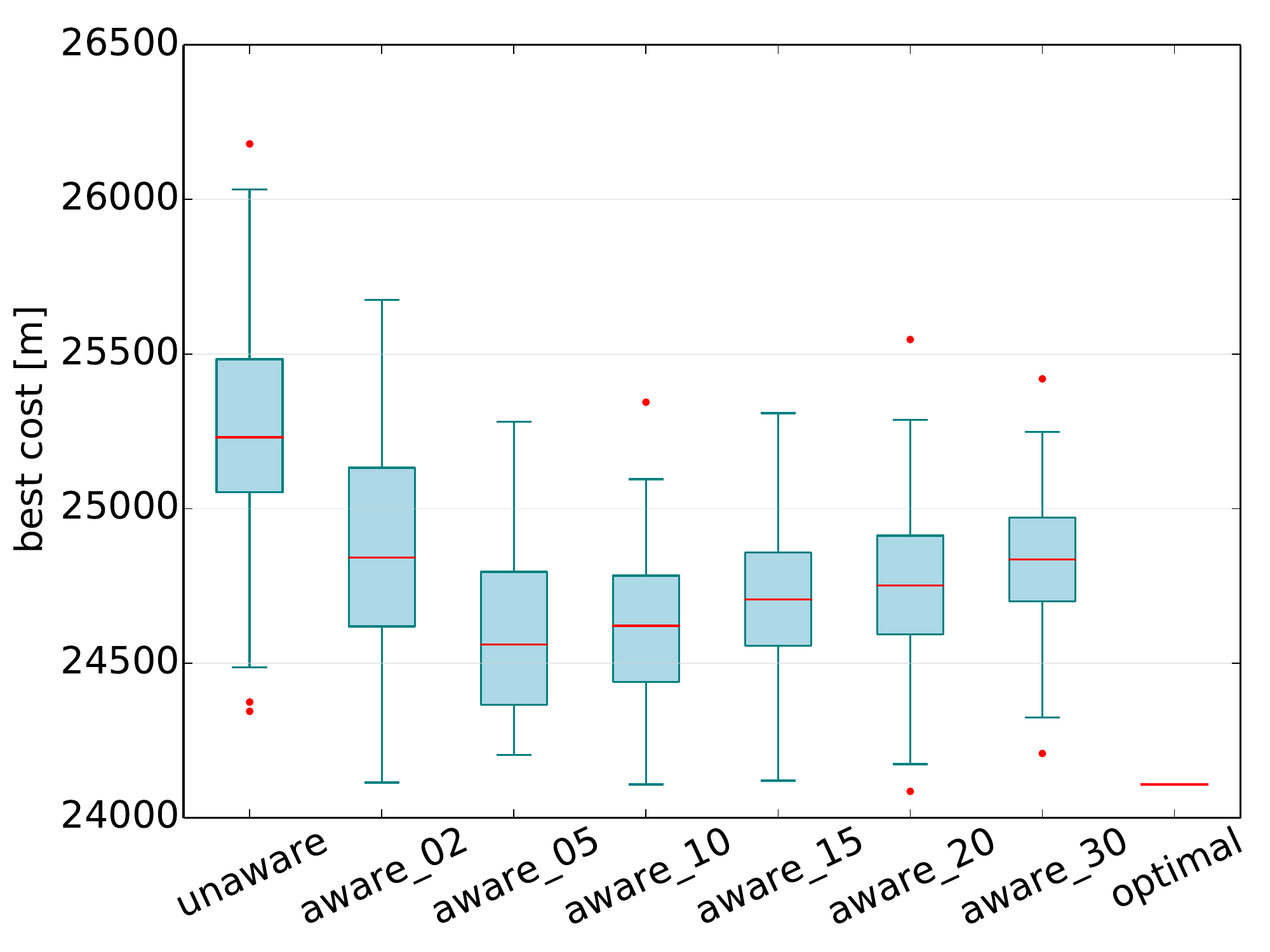}\label{fig:PL_Planning_samplingdomlength}}
\caption{Comparison of obstacle-unaware and obstacle-aware planning (with $k=[2,30]$ given as a postfix) in the \emph{Dom} scenario after two seconds of planning. The right plot includes the cost after 2 hours of planning (\emph{optimal}) as a reference.}
\end{figure}

The iterative $z$-shift is $\text{dz}_\text{shift} = k \cdot d_\text{icc}$, where $d_\text{icc}$ again denotes the intermediate collision check distance and $k$ is a design parameter that is varied for this analysis. The \emph{Dom} scenario is chosen. Note that due to the significant altitude changes in this scenario, the \emph{optimal} path consists exclusively of maximum climb or sink rate operation (\cref{fig:PL_Planning_Impr_TestcasesDomB}). As a result, the planner does \emph{not} chose the most direct path in the horizontal plane (\cref{fig:PL_Planning_Impr_TestcasesDomA}), but could instead choose any path that allows constant maximum climb and sink rate. \Cref{fig:PL_Planning_samplingdomstates} compares the number of valid states in the motion tree after two seconds of planning. The number of valid states is lowest for obstacle-unaware sampling. It increases asymptotically with $k$ for obstacle-aware sampling. The path costs in \cref{fig:PL_Planning_samplingdomlength} clearly show that due to more valid states the obstacle-aware sampling results in lower path costs. The cost errors of the worst configuration (obstacle-unaware sampling) and best configuration (obstacle-aware sampling with $k=5$) to the \emph{optimal} cost are \unit[1080]{m} (4.5\% error) and \unit[430]{m} (1.8\%) respectively. The error has thus been reduced by more than half, which due to the asymptotic convergence of the planner means that the obstacle-ware planning leads to significantly faster convergence of the solution. For larger $k$ and thus $dz_\text{shift}$, the path cost increases again because the shifted samples have a higher distance from the terrain and the path can thus not follow the terrain as closely. The obstacle-aware sampling benefits are high in the \emph{Dom} scenario because the terrain $Q_\text{obs}$ is a significant part of the state space $Q$. \citet{Achermann2017MT} presents further test cases with flatter terrain. Overall, the results suggest to choose $k=[5,10]$.

%%%%%%%%%%%%%%%%%%%%%%%%%%%%%%%%%%%%%%%%%%%%%%%%%%%%%%%
\subsubsection{Nearest Neighbor Search}
\label{sec:PL_Planning_Improvements_NeighborSearch}
%%%%%%%%%%%%%%%%%%%%%%%%%%%%%%%%%%%%%%%%%%%%%%%%%%%%%%%

The nearest neighbor search identifies the closest neighbor or the k-nearest neighbors of a state. RRT* performs each of these computations once per planning iteration. The neighbor search makes up \unit[90]{\%} of the planning time because it involves multiple Dubins path calculations. Two contributions are therefore outlined below. 

%%%%%%%%%%%%%%%%%%%%%%%%%%%%%%%%%%%%%%%%%%%%%%%%%%%%%%%
\paragraph{Speeding up the nearest neighbor search via a Dubins path approximation}
%%%%%%%%%%%%%%%%%%%%%%%%%%%%%%%%%%%%%%%%%%%%%%%%%%%%%%%
When reusing the definitions of $d_\text{eucl}$ from \cref{eqn:PL_Planning_eucldist} and $d_\text{climb}$ from \cref{eqn:PL_Planning_dubapproxdist}, one can define the Dubins distance approximation 
	\begin{equation}
	\hat{d}_\text{dubins}(q_1, q_2)= \max \big(d_\text{eucl}(q_1, q_2), d_\text{climb}(q_1, q_2) \big) \; .
	\label{eqn:PL_Planning_dubapprox}
	\end{equation}
Computing $\hat{d}_\text{dubins}$ is around ten times faster than computing the Dubins distance $d_\text{dubins}$ itself. The Dubins approximation can be used to define an upper bound for the actual Dubins distance:
\begin{equation}
d_\text{dubins} \leq \hat{d}_\text{dubins} + \frac{(4 \pi + 2) \cdot r_\text{turn}}{\cos(\gamma_\text{max})} \; ,
\label{eqn:PL_Planning_dubapprox2}
\end{equation}
where $\text{r}_\text{turn}$ is the aircraft minimum turn radius. Both $r_\text{turn}$ and $\gamma_\text{max}$ are fixed aircraft parameters and the term on the right hand side is thus constant. It is derived by considering the worst case difference between Dubins distance and Euclidean distance: When the start and goal position are equal but their heading $\psi$ is offset by 180\degree, then the path offset is $\left(2 \pi + 2 \right) \cdot \text{r}_\text{turn}$. The division by $\cos(\gamma_\text{max})$ accounts for climb or sink operations. An additional $2\pi r_\text{turn}/\cos(\gamma_\text{max})$ needs to be added to the offset because the sub-optimal Dubins airplane path is used for the medium altitude case~\cite{Schneider2016MT}. The proposed method to compute the nearest neighbors then is:

\begin{enumerate}
\item
Compute the approximation $\hat{d}_\text{dubins}$ from the input state to every state in the motion tree and order them from smallest to largest $\hat{d}_\text{dubins}$.
\item For the nearest neighbor search the threshold distance is set to the first distance in the ordered list, i.e. $d_\text{thres} = \hat{d}_\text{dubins}[1]$. For the k-nearest neighbor search it is set to $\smash{\hat{d}_\text{dubins}[k]}$, i.e. the k-th element.
\item Identify the index $i$ of the first element in the list for which
\begin{equation}
\hat{d}_\text{dubins}[i] > d_\text{thres} + \frac{(4\pi+2) \cdot r_\text{turn}}{\cos(\gamma_\text{max})} \; .
\end{equation}
\item Compute the actual Dubins distance $d_\text{dubins}$ from the input state to all states in the sorted list up to $i-1$ and return the nearest or k-nearest states, i.e. those with the smallest value $d_\text{dubins}$.
\end{enumerate}

We expect that the planner's convergence is sped up because the Dubins distance is only calculated for the first $i-1$ states and not all states in the motion tree. This should apply especially for the late planning stages when the motion tree is large.

\paragraph{Considering Dubins path asymmetry during the nearest neighbor calculation} 
The standard RRT* in \ac{OMPL} is only designed for symmetric paths. However, the Dubins airplane path is not symmetric, i.e. the distances $q_1\rightarrow q_2$ and $q_2\rightarrow q_1$ are generally not equal. Therefore, the k-nearest neighbors from input state to motion tree (required in the RRT* rewiring step) might differ from the k-nearest neighbors from motion tree to input state (required to initially connect the motion tree to the new input state). However, the standard RRT* planner~\cite{Karaman2010RRTstarKinodynamic} and its OMPL-implementation only compute a single set of nearest neighbors and thus use a potentially wrong neighborhood. We therefore adapt the OMPL RRT* to compute the k-nearest neighbors in both directions. While this is the correct way for asymmetric motions, it of course requires twice the amount of k-nearest neighbor calculations per iteration.

%%%%%%%%%%%%%%%%%%%%%%%%%%%%%%%%%%%%%%%%%%%%%%%%%%%%%%%
\paragraph{Preliminary results}
%%%%%%%%%%%%%%%%%%%%%%%%%%%%%%%%%%%%%%%%%%%%%%%%%%%%%%%

%This comment has to be here, otherwise text layout gets messed up
\begin{table}[b]
\caption{The nearest-neighbor search test configurations. \emph{Dub.} means the standard OMPL implementation, which calculates the Dubins path to \emph{all} neighbors, is used. \emph{Dub.Appx.} means the Dubins approximation of \cref{eqn:PL_Planning_dubapprox2} is used. The 1-way calculation in a Dubins case always represents the motion from the state to the motion tree.}
\centering
\begin{tabular}{l l l l l}
\toprule
 & \multicolumn{2}{c}{Nearest neighbor} & \multicolumn{2}{c}{k-nearest neighbors} \\
 & Method & Direction & Method & Direction \\
\midrule
\emph{ompl\_lin} & Dub. & 1-way & Dub. & 1-way \\
\emph{1way\_both} & Dub.Appx. & 1-way & Dub.Appx. & 1-way \\
\emph{2way\_both} & Dub.Appx. & 1-way & Dub.Appx. & 2-way \\
\emph{1way\_onlyk} & Dub. & 1-way & Dub.Appx. & 1-way \\
\emph{2way\_onlyk} & Dub. & 1-way & Dub.Appx. & 2-way \\
\bottomrule
\end{tabular}
\label{tab:PL_Planning_Impr_Nearest_Configs}
\end{table}

\begin{figure}[htbp]
\centering
\subfloat[Number of iterations.]{\includegraphics[width=0.48\textwidth]{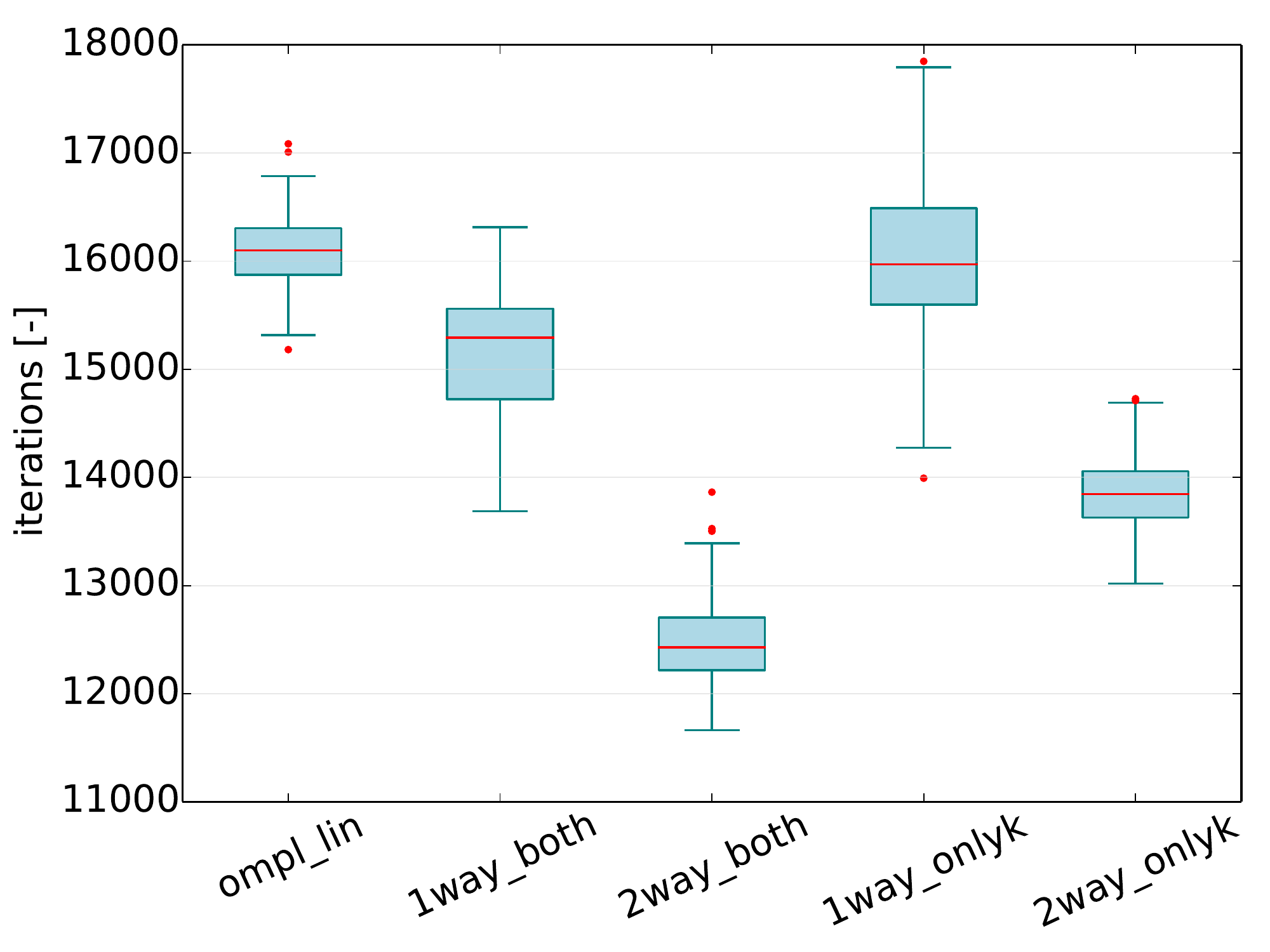}\label{fig:PL_Planning_Impr_nnkliterations15s}}
\hfill
\subfloat[Path length in meters.]{\includegraphics[width=0.48\textwidth]{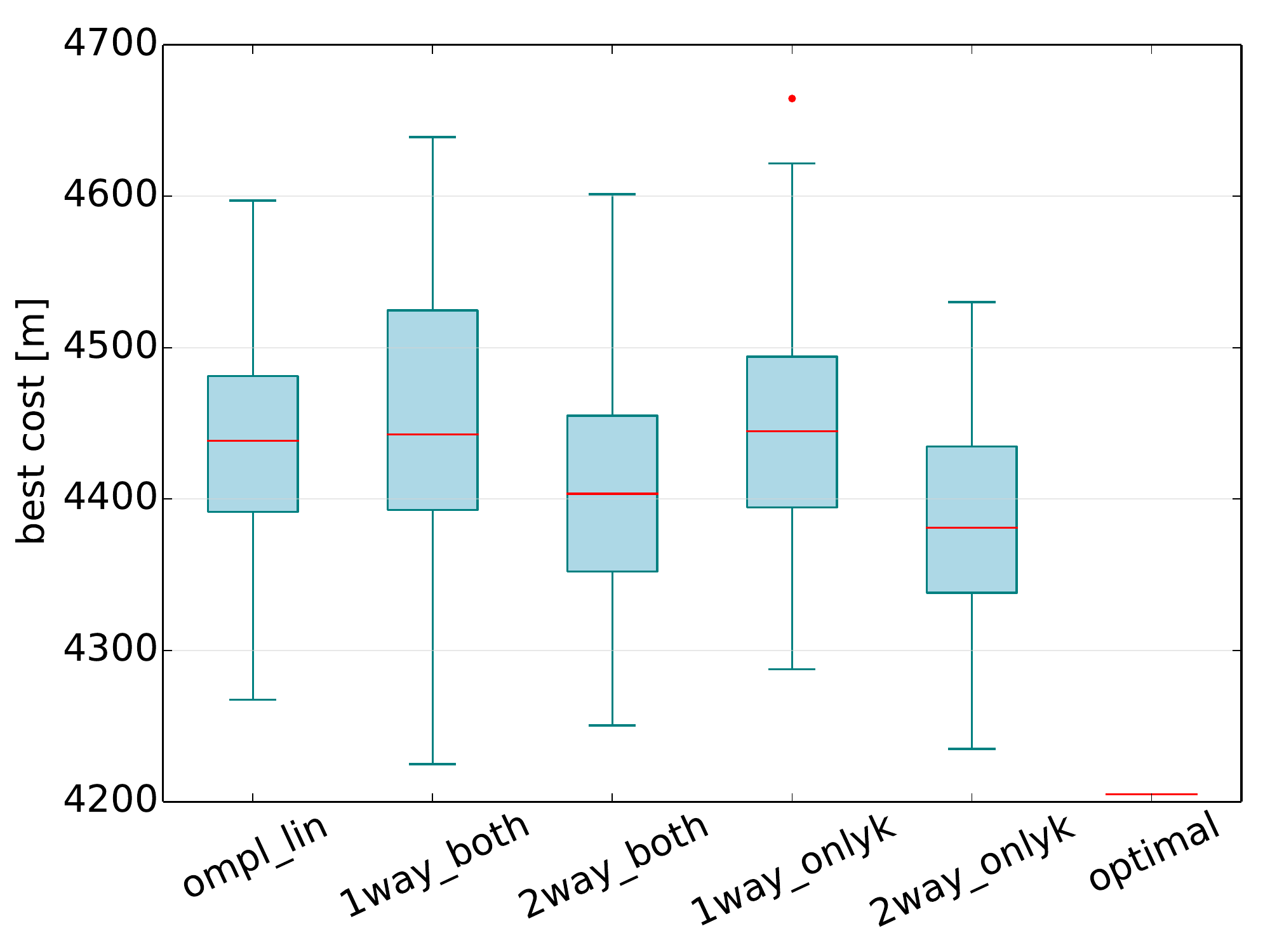}\label{fig:PL_Planning_Impr_nnkllength15s}}
%\subfloat[Path length in meters.]{\includegraphics[width=0.48\textwidth]{images/Planning/Improvements/NeighborCalc/test2.pdf}\label{fig:PL_Planning_Impr_nnkllength15s}}
\caption{Comparison of the OMPL and custom nearest-neighbor search methods in the \emph{Kandertal Low} scenario after \unit[15]{s} planning time. The optimal path is shown for reference.}
\label{fig:PL_Planning_Impr_NN}
\end{figure}

The assessment uses the \emph{Kandertal Low} scenario and the nearest/k-nearest neighbor search configurations in \cref{tab:PL_Planning_Impr_Nearest_Configs}: First, the RRT* standard neighbor search method which calculates the Dubins distance to \emph{all} neighbors but is still called \emph{linear}\footnote{Linear describes the computational complexity of the search method and not how the distance is computed, see \url{http://ompl.kavrakilab.org/classompl_1_1NearestNeighborsLinear.html}} in OMPL, and second, the Dubins approximation of \cref{eqn:PL_Planning_dubapprox2}. \Cref{fig:PL_Planning_Impr_nnkliterations15s} shows the number of iterations completed in \unit[15]{s} planning time, which generally decreases for the custom neighbor search implementation: While \emph{ompl\_lin} computes the distance from the input state to every state in the motion tree once (complexity $O\left(n\right)$), our custom implementation involves sorting $\hat{d}_\text{dubins}$ and thus has complexity $O\left(n \log \left(n\right)\right)$. Hence, the custom implementation is slower although fewer Dubins paths are computed. Nevertheless, as shown in \cref{fig:PL_Planning_Impr_nnkllength15s}, the path cost tends to improve. The \emph{2way\_both} and \emph{2way\_onlyk} configurations decrease the remaining error with respect to the optimal path by 25\% because they always select the correct neighborhood and therefore the optimal connection between every state in the motion tree is found. This is not always the case for the standard one-way implementation. Note that Achermann~\cite{Achermann2017MT} presents further test scenarios. The overall conclusion is that the \emph{2way\_onlyk} configuration yields the largest improvement and should thus be used.

%%%%%%%%%%%%%%%%%%%%%%%%%%%%%%%%%%%%%%%%%%%%%%%%%%%%%%%
\subsubsection{Code Optimization}
\label{sec:PL_Planning_codeoptimization}
%%%%%%%%%%%%%%%%%%%%%%%%%%%%%%%%%%%%%%%%%%%%%%%%%%%%%%%
%Note: Needed for journal paper?? Can remove.
The Dubins path computation and interpolation is the path planning bottleneck and makes up \unit[80]{\%} of the total planning time. Code optimizations were therefore implemented: First, trigonometric and square-root evaluations for constant angles (e.g. $\gamma_\text{max}$) were replaced by pre-computed functions. In addition, trigonometric functions are only evaluated with single instead of double precision, giving a speed up of factor four per evaluation at negligible precision loss (on the order of millimeters). Second, the Dubins path interpolation is improved: The Dubins airplane path consists of at most six segments: Start helix, left/right turn, additional turn in case of the optimal path for the medium altitude case, straight line or left/right turn, another left/right turn, and the goal helix. The segment types, lengths, and climbing angles uniquely define the Dubins airplane path. Previously, for interpolating a state in the $i$-th segment, as required numerous times during collision checking, all previous $i-1$ segments were recomputed. Our implementation calculates the segment ends for a path only once and then caches them for further reuse.

\begin{figure}[htbp]
\centering
\subfloat[Number of iterations]{\includegraphics[width=0.48\textwidth]{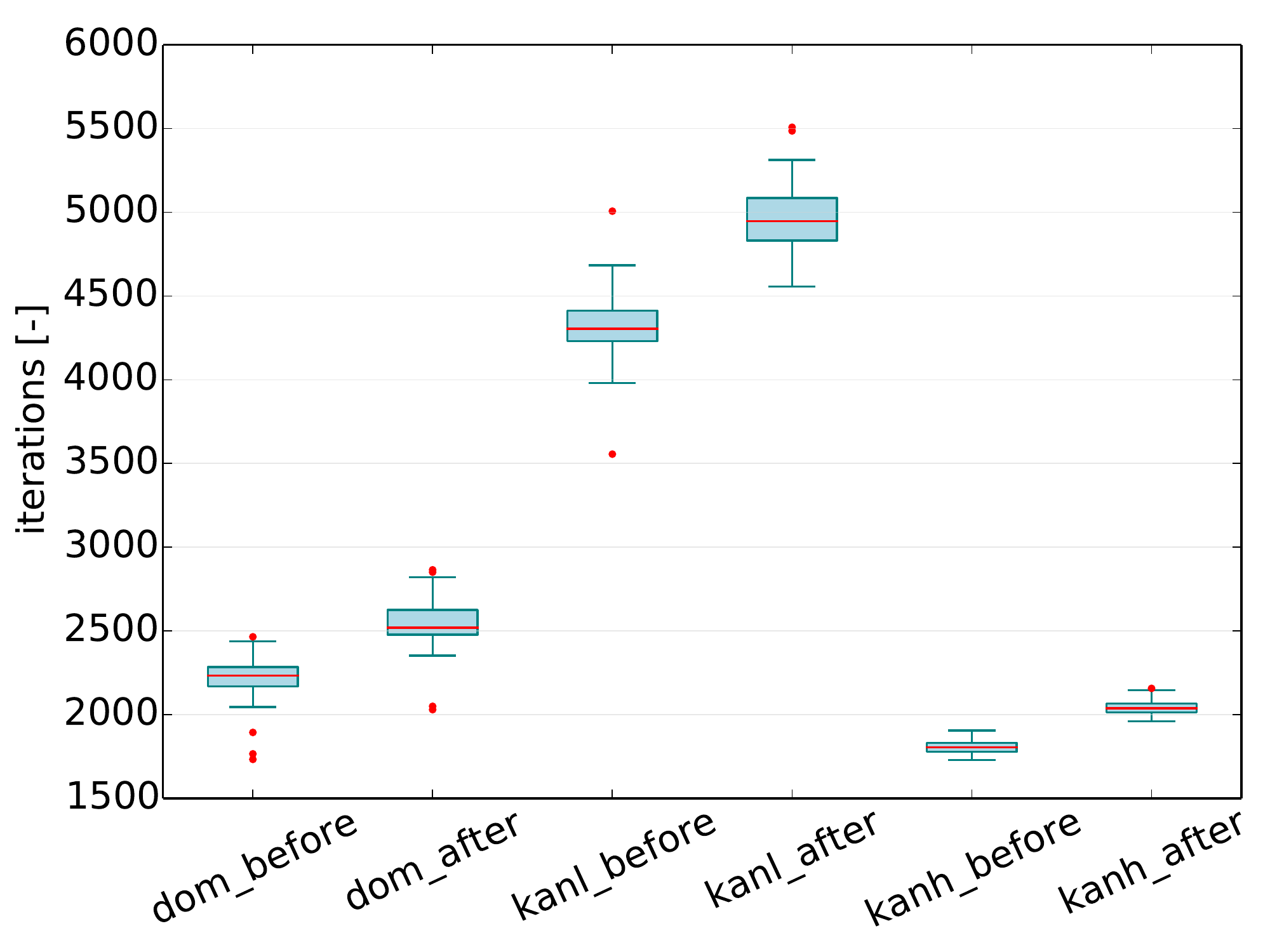}\label{fig:PL_Planning_codeoptiterations}}
\hfill
\subfloat[Solution cost after two seconds normalized by the cost after two hours of planning.]{\includegraphics[width=0.48\textwidth]{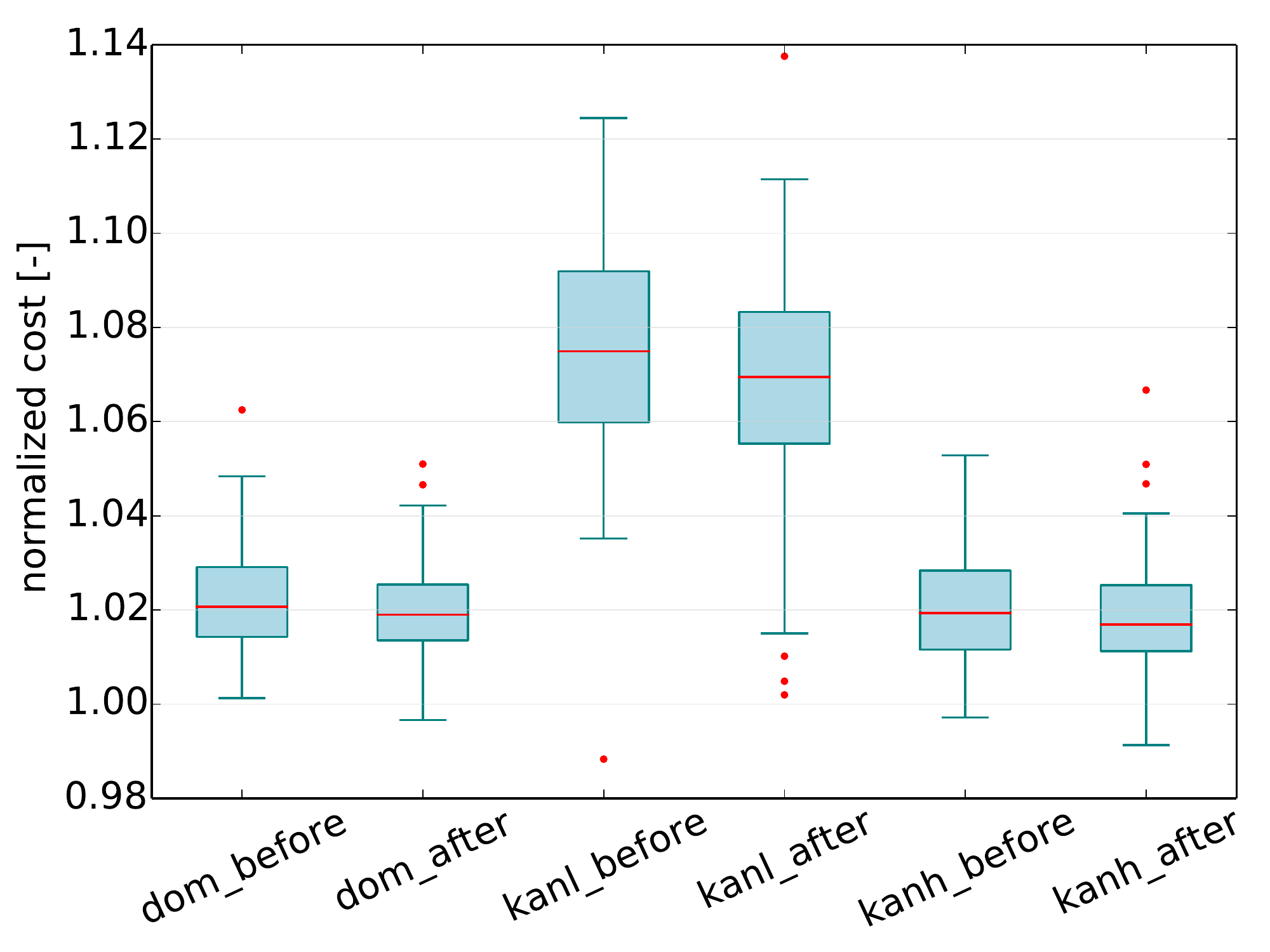}\label{fig:PL_Planning_codeoptpathlength}}
\caption{Iterations and path cost before and after the code optimization when planning \unit[2]{s}. The scenarios are \emph{Dom} (dom\_), \emph{Kandertal Low} (kanl\_) and \emph{Kandertal High} (kanh\_).}
\end{figure}

%%%%%%%%%%%%%%%%%%%%%%%%%%%%%%%%%%%%%%%%%%%%%%%%%%%%%%%
\paragraph{Preliminary results}
%%%%%%%%%%%%%%%%%%%%%%%%%%%%%%%%%%%%%%%%%%%%%%%%%%%%%%%
The performance improvements are evaluated in the \emph{Dom}, \emph{Kandertal Low}, and \emph{Kandertal High} scenarios (where \emph{Kandertal High} uses the same map as \emph{Kandertal Low} but a higher-altitude goal state). The number of iterations after two seconds of planning (\cref{fig:PL_Planning_codeoptiterations}) increases by around \unit[10]{\%}. The motion tree is therefore also \unit[10]{\%} larger. Due to the asymptotic convergence of the planner, the effect on the path length is much smaller (\cref{fig:PL_Planning_codeoptpathlength}). The largest performance gain is seen for \emph{Kandertal Low} where the cost is decreased by about 0.5 percent points (7\% of the remaining error to the solution after 2 hours of planning). Nevertheless, the Dubins path interpolation speed up is important because the time-optimal planning of \cref{sec:PL_Planning_TimeOptimal} requires even more Dubins path calculations than the shortest-path planning.

%%%%%%%%%%%%%%%%%%%%%%%%%%%%%%%%%%%%%%%%%%%%%%%%%%%%%%%
\subsubsection{Performance Results}
\label{sec:PL_Planning_Improvements_Results}
%%%%%%%%%%%%%%%%%%%%%%%%%%%%%%%%%%%%%%%%%%%%%%%%%%%%%%%
\setcounter{topnumber}{3}

\begin{figure}[htbp]
\centering
\subfloat[Number of states in the motion tree.]{\includegraphics[width=0.48\textwidth]{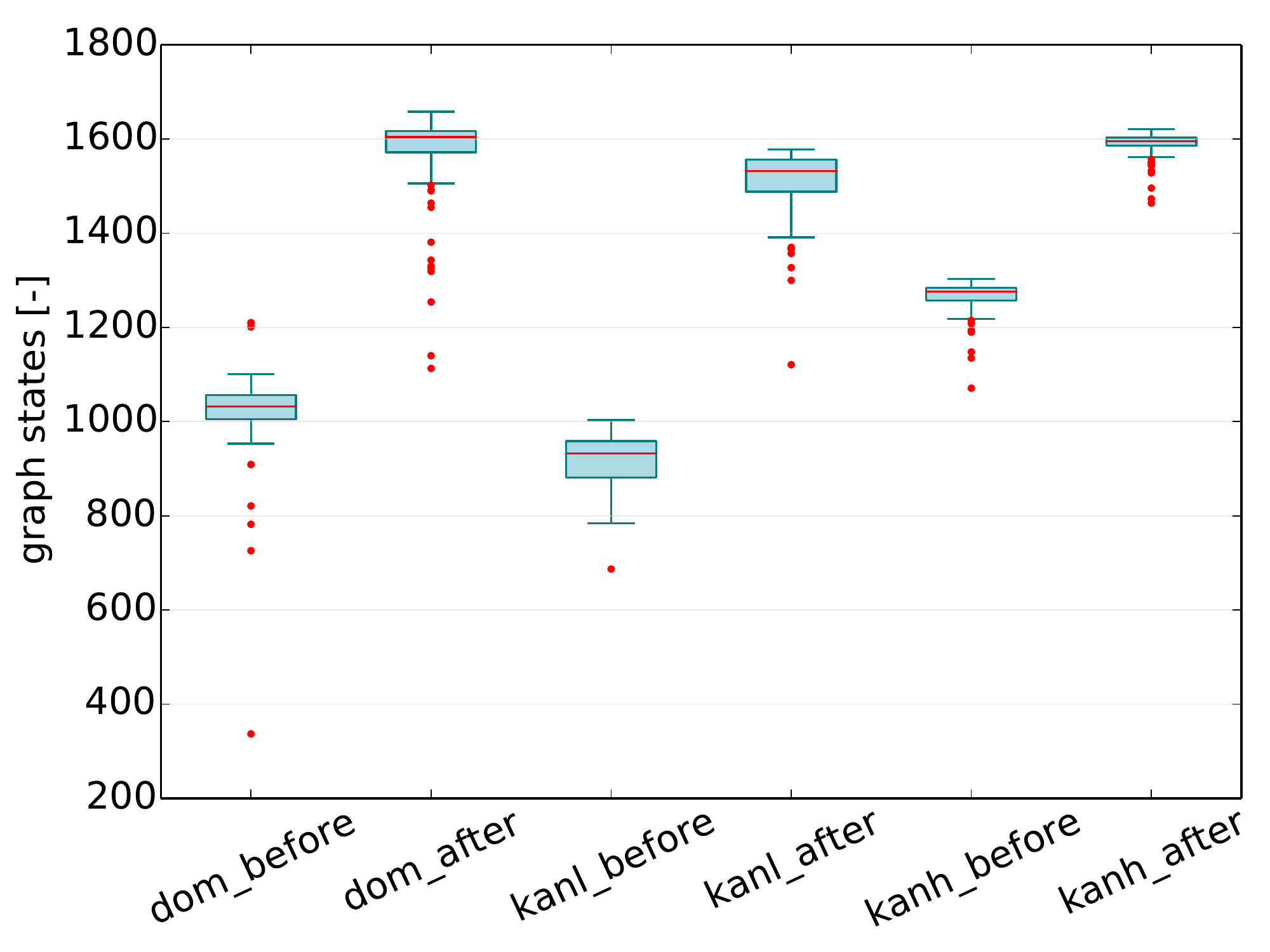}\label{fig:PL_Planning_beforeafterstates}}
\hfill
\subfloat[Cost of solution path after two seconds normalized by cost after two hours of planning.]{\includegraphics[width=0.48\textwidth]{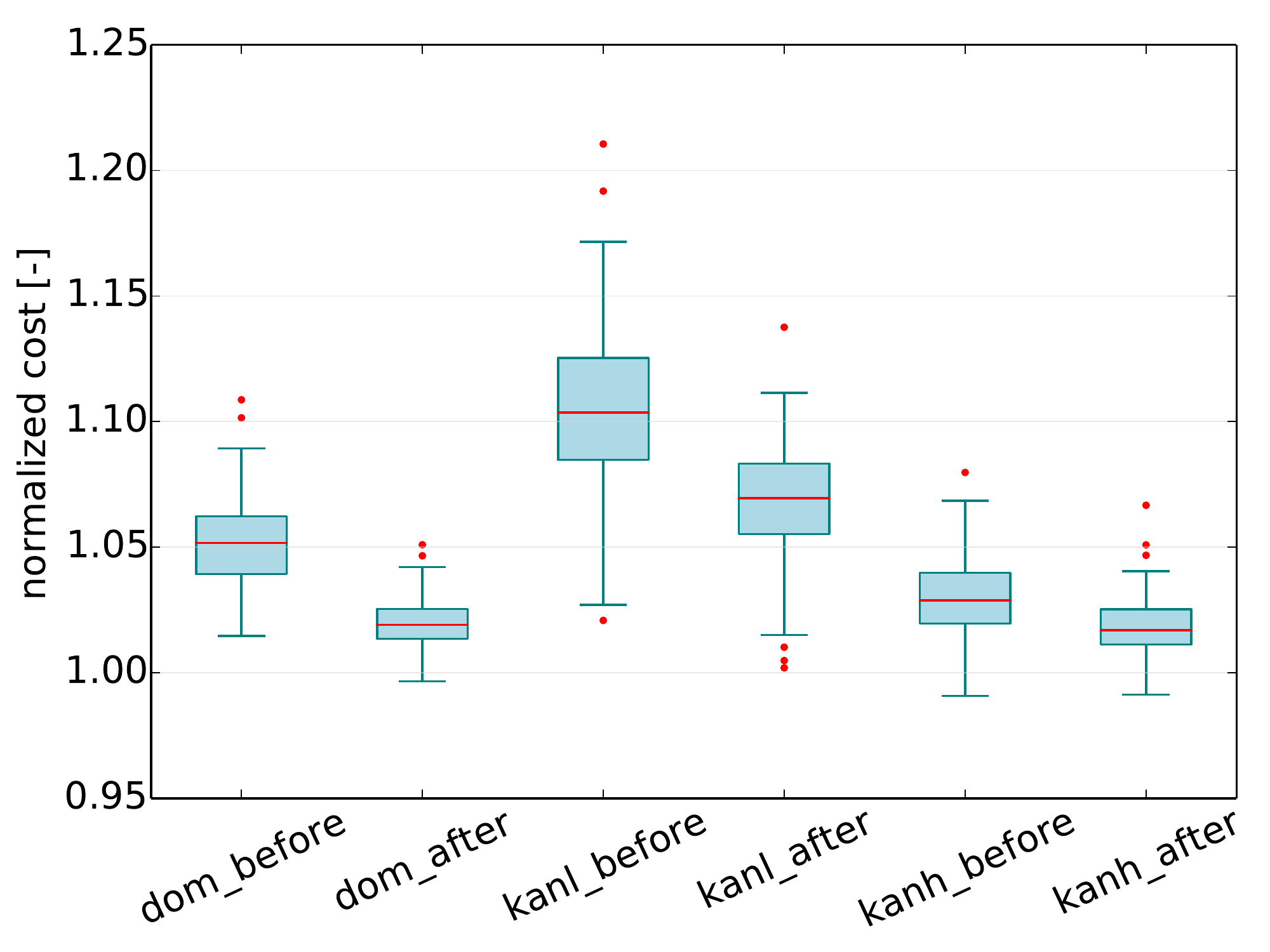}\label{fig:PL_Planning_beforeaftercost}}
\caption{Number of motion tree states and path cost after two seconds of planning time before and after the improvements implemented in this paper. The scenarios are \textit{Dom}(dom\_), \textit{Kandertal Low}(kanl\_), and \textit{Kandertal High}(kanh\_).}
\end{figure}

\begin{figure}[htbp]
\centering
\subfloat[Time required to find a solution at most 5\% longer than the path after two hours of planning. The planning time is limited to {\unit[300]{s}}.]{\includegraphics[width=0.48\textwidth]{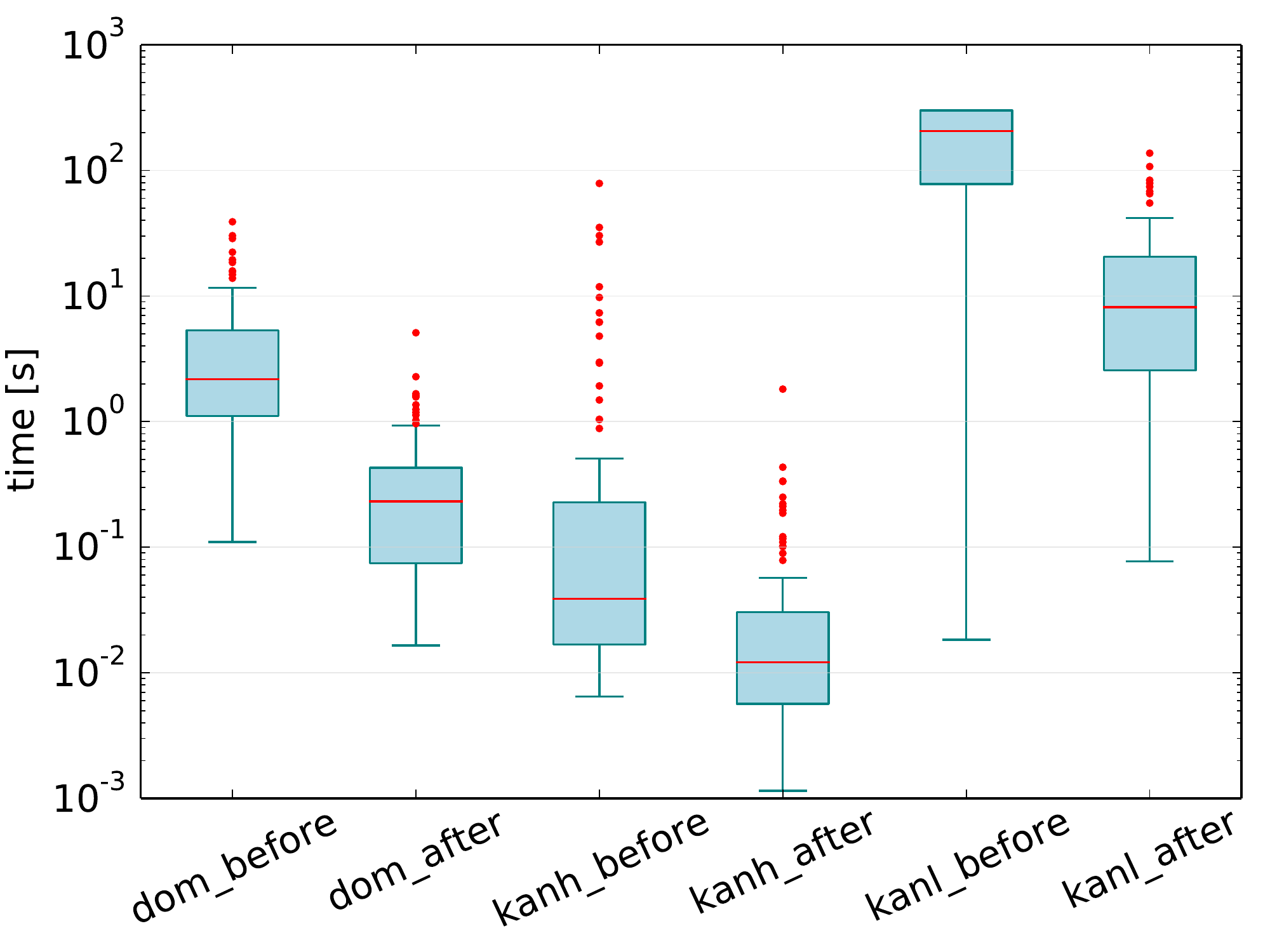}\label{fig:PL_Planning_t5ptimebeforeafter}}
\hfill
\subfloat[Convergence behavior, i.e. normalized path cost over two hours of planning time.]{\includegraphics[width=0.48\textwidth]{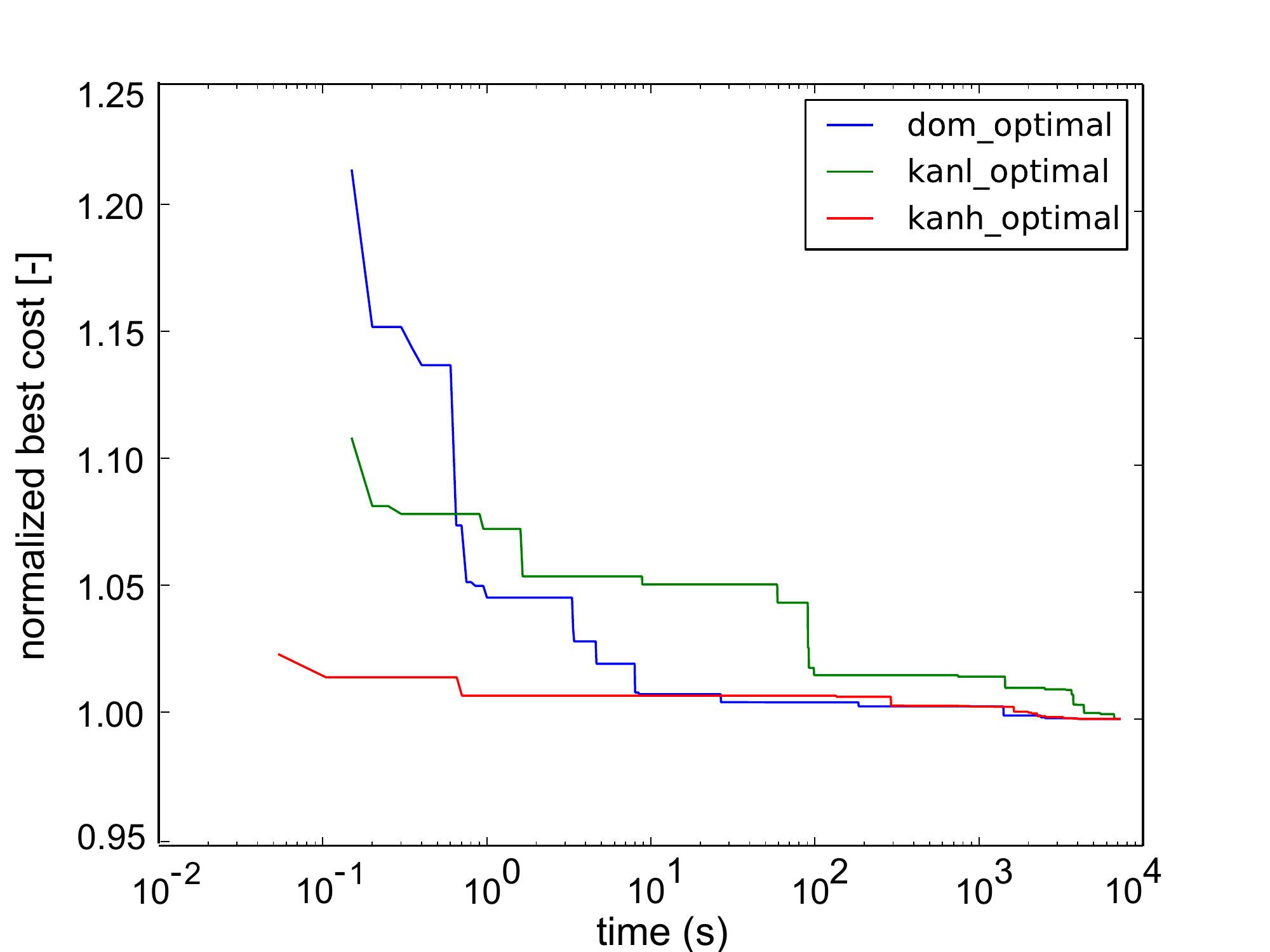}\label{fig:PL_Planning_convergence2h}}
\caption{Performance metrics before and after the implementation of the new methods. The scenarios are \textit{Dom}(dom\_), \textit{Kandertal Low}(kanl\_), and \textit{Kandertal High}(kanh\_).}
\end{figure}

This section assesses the RRT* planning performance improvements resulting from the features implemented in \cref{sec:PL_Planning_Improvements_InformedSubset,sec:PL_Planning_Improvements_CollisionChecking,sec:PL_Planning_Improvements_Sampling,sec:PL_Planning_Improvements_NeighborSearch,sec:PL_Planning_codeoptimization}. The selected optimum configuration is a) obstacle-aware sampling with a shifting parameter $k=5$, b) height checking with the original 2.5D map, and c) custom nearest neighbor search only for k-nearest but in both directions. The number of motion tree states after two seconds of planning (\cref{fig:PL_Planning_beforeafterstates}) is increased by \unit[30--60]{\%}. \Cref{fig:PL_Planning_beforeaftercost} shows that the path length errors with respect to the optimal path are reduced from 5.2 to 1.9 percent points (63\% reduction), from 10.4 to 7.0 percent points (33\% reduction) and from 2.9 to 1.7 percent points (41\% reduction) respectively in the \emph{Dom}, \emph{Kandertal low} and \emph{Kandertal high} scenarios. Overall, the error relative to the optimal path is approximately cut in half. \Cref{fig:PL_Planning_t5ptimebeforeafter} shows that a 5\% error with respect to the optimal path is reached 10x, 20x and 5x faster than without the optimizations in the \textit{Dom}, \textit{Kandertal Low} and \textit{Kandertal High} scenarios respectively. The presented methods thus significantly improve the planning performance. \Cref{fig:PL_Planning_convergence2h} visualizes the convergence of the planner: For the map sizes, resolution and the non-informed RRT* planner used here, solutions with $<8\%$ error to the optimal solution are on average available after one second and a $<5\%$ error is reached within two seconds. This is a main reason why we consider the framework capable of \emph{real-time} path planning for a UAV. 

%%%%%%%%%%%%%%%%%%%%%%%%%%%%%%%%%%%%%%%%%%%%%%%%%%%%%%%
%%%%%%%%%%%%%%%%%%%%%%%%%%%%%%%%%%%%%%%%%%%%%%%%%%%%%%%
\subsection{Wind-Aware Path Planning}
\label{sec:PL_Planning_TimeOptimal}
%%%%%%%%%%%%%%%%%%%%%%%%%%%%%%%%%%%%%%%%%%%%%%%%%%%%%%%
%%%%%%%%%%%%%%%%%%%%%%%%%%%%%%%%%%%%%%%%%%%%%%%%%%%%%%%

To extend the shortest-path planning to time-optimal planning in 3D wind fields, first, \cref{sec:PL_Planning_dubinswithwind} describes the iterative Dubins airplane path calculation in wind. Second, \cref{sec:PL_Planning_timeoptimalmotioncostheuristic} presents the adapted motion cost heuristic. Third, the obstacle-aware sampling of \cref{sec:PL_Planning_Improvements_Sampling} is still partially possible, however, given that in wind the path does not generally fulfill $z>z_\text{s,g}^\text{min}$ this assumption is removed. Fourth, the standard OMPL neighbor search is used given that \cref{eqn:PL_Planning_dubapprox} is not valid in wind. 

%%%%%%%%%%%%%%%%%%%%%%%%%%%%%%%%%%%%%%%%%%%%%%%%%%%%%%%
\subsubsection{Computing the Dubins Airplane Path in Presence of Wind}
\label{sec:PL_Planning_dubinswithwind}
%%%%%%%%%%%%%%%%%%%%%%%%%%%%%%%%%%%%%%%%%%%%%%%%%%%%%%%

\begin{figure}[htbp]
\centering
\subfloat[The Dubins path in wind with respect to the ground and air. The wind-induced shift between the end positions of the two paths is $\smash{\Delta q_\text{goal}^\text{wind}}$.]{\includegraphics[height=6.6cm]{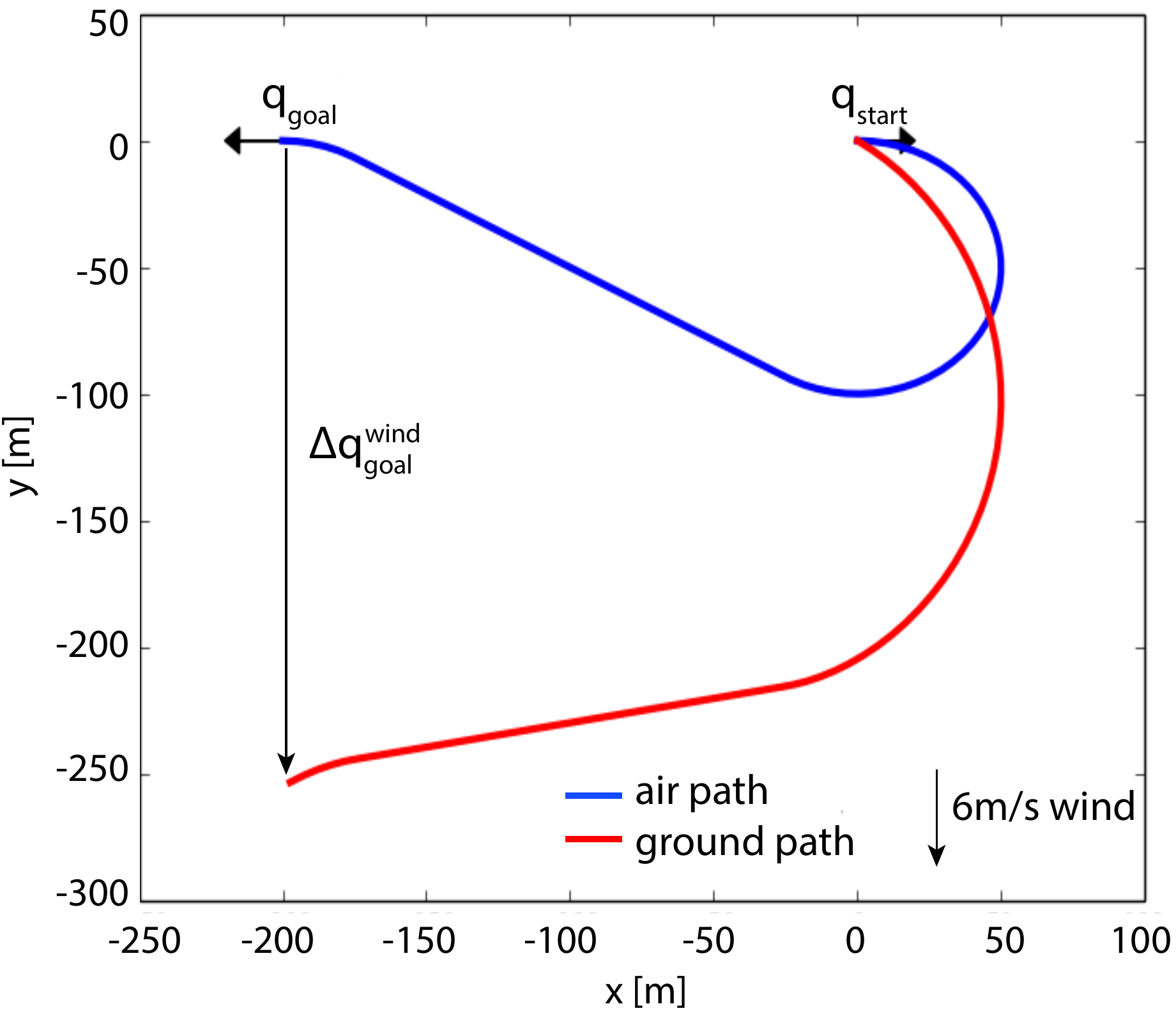}\label{fig:PL_Planning_winddriftmcgee}}
\hfill
\subfloat[The success rate of computing the Dubins airplane path in presence of wind versus the maximum number of allowed iterations.]{\includegraphics[height=6.6cm]{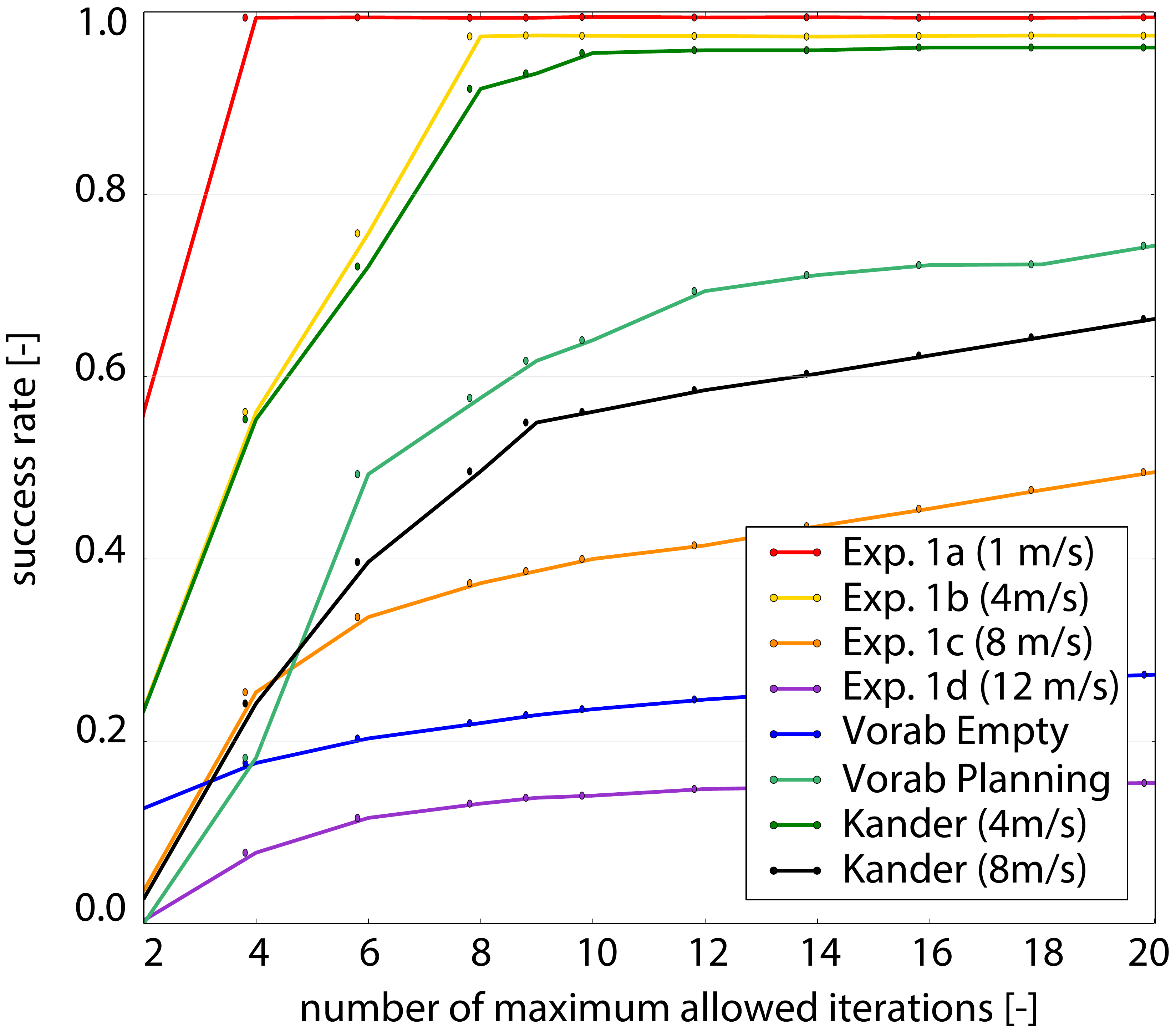}\label{fig:PL_Planning_convergende1t4}}
\caption{Approach and performance of the Dubins airplane path calculation in wind}
\end{figure}

Airplane motion planning in wind fields is discussed by numerous authors~\cite{Ceccarelli2007,Otte2016,Chakrabarty2013PlanningDynWindFields}. \citet{McGee2005} present an algorithm to compute the Dubins airplane path for a uniform wind field. Given that this paper also employs Dubins airplane paths, we extend McGee's approach for spatially varying wind fields. \Cref{alg:PL_Planning_dubinspathwithwind} presents our iterative approach: First, the Dubins path without wind is computed as described in \cref{sec:PL_Planning_Fundamentals_DubinsNoWind}. Second, the resulting path is simulated with wind and the resulting goal pose error due to the wind drift $\smash{\Delta q_\text{goal}^\text{wind}}$ is determined (\cref{fig:PL_Planning_winddriftmcgee}). Third, the new shifted \emph{virtual} goal state $\smash{q_\text{goal}^\text{virt}=q_\text{goal}-\Delta q_\text{goal}^\text{wind}}$ is determined. 
%Third, the goal state is shifted by $-\Delta q_\text{goal}^\text{wind}$. This yields the new shifted \emph{virtual} goal state $q_\text{goal}^\text{virt}$.
The three steps are then repeated iteratively with the new virtual goal state until the ground-relative Dubins path end point and the original goal state have converged to within \unit[1]{m} Euclidean distance. %Because the convergence is slow for certain start/goal configurations the process is aborted after a maximum number of iterations.

\begin{algorithm}[htb]
\caption{Computing the Dubins airplane path in spatially-varying wind}
\label{alg:PL_Planning_dubinspathwithwind}
\begin{algorithmic} [1]
\Procedure{computeDubinsPathWithWind}{$q_\text{start}, q_\text{goal}, \text{windfield}$}
\State $q_\text{goal}^\text{virt} \gets q_\text{goal}$
\While {$\left(\text{converged} == \text{false} \right)$}
  \State $\text{airpath} \gets \textsc{computeDubinsPathWithoutWind} \left(q_\text{start}, q_\text{goal}^\text{virt}\right)$
  \State $\Delta q_\text{goal}^\text{wind} \gets \textsc{computeWindDrift} \left(\text{airpath}, \text{windfield}\right)$
  \State $q_\text{goal}^\text{virt} \gets \textsc{shiftState} \left(q_\text{goal}, \Delta q_\text{goal}^\text{wind}\right)$
  \State $\text{converged} \gets \textsc{isConverged} \left(\text{airpath}, \text{windfield}\right)$
\EndWhile
\EndProcedure
\end{algorithmic}
\end{algorithm}

\Cref{alg:PL_Planning_computewinddrift} describes the wind drift computation. The aircraft position is retrieved through interpolating the air-relative Dubins path and adding the current wind drift. \textsc{GetWindData} determines the wind at the current ground-relative position. The wind drift over the time segment $\Delta t$ is accumulated and the process repeated until the Dubins path is complete. A small $\Delta t_\text{max}$ increases the calculation accuracy but also the computation time because more Dubins path interpolations are required (which is why the improvements of \cref{sec:PL_Planning_codeoptimization} are important).

\begin{algorithm}[htb]
\caption{Computing the wind drift through numerical integration.}
\label{alg:PL_Planning_computewinddrift}
\begin{algorithmic} [1]
\Procedure{computeWindDrift}{$\text{airpath}, \text{windfield}$}
\State $t \gets 0.0$
\State $\Delta q_\text{goal}^\text{wind} \gets \left(0.0, 0.0, 0.0\right)$
\While {$t \leq 1.0 $}
  \State $\Delta t \gets \min\left(\Delta t_\text{max}, 1.0 - t \right)$
  \State $q_\text{air}  \gets \textsc{interpolate} \left(\text{airpath}, t\right)$
  \State $q_\text{ground} \gets q_\text{air} + \Delta q_\text{goal}^\text{wind}$
  \State $\text{winddata} \gets \textsc{getWindData} \left(\text{windfield}, q_\text{ground} \right)$
  \State $\Delta q_\text{goal}^\text{wind} \gets \textsc{incrementWindDrift} \left(\Delta t, \text{winddata}\right)$
  \State $t \gets t + \Delta t $
\EndWhile
\State \Return $\Delta q_\text{goal}^\text{wind}$
\EndProcedure

\end{algorithmic}
\end{algorithm}

%%%%%%%%%%%%%%%%%%%%%%%%%%%%%%%%%%%%%%%%%%%%%%%%%%%%%%%
\paragraph{Preliminary Results}
%%%%%%%%%%%%%%%%%%%%%%%%%%%%%%%%%%%%%%%%%%%%%%%%%%%%%%%
The experimental scenarios conducted to examine the convergence of the Dubins path calculation in the presence of wind are given below. The airspeed is $v_\text{air}=\unitfrac[9]{m}{s}$. Each experiment is run 100'000 times with random start and goal states.
\begin{enumerate}
\item \emph{Experiment 1a-d}: Wind field of \cref{fig:PL_Planning_w1} with size \unit[3500]{m} x \unit[3500]{m} x \unit[3000]{m} and wind magnitude a) \unitfrac[1]{m}{s}, b) \unitfrac[4]{m}{s}, c) \unitfrac[8]{m}{s}, and d) \unitfrac[12]{m}{s}.
\item \emph{Experiment 2a/b}: \emph{Vorab} region with a \unit[3000]{m} x \unit[3000]{m} x \unit[2000]{m} wind field calculated by the meteo node (\cref{sec:PL_WindPrediction}). The maximum wind is \unitfrac[37]{m}{s}, i.e. $v_\text{wind}\gg v_\text{air}$. Case a) uses uninformed sampling, case b) uses informed sampling such that the start/goal states are closer to each other.
\item \emph{Experiment 3a/b}: \emph{Kandertal} region with, for exp. 3a) the same wind field as in exp. 1b), and for exp. 3b) the same wind field as in exp. 1c).
\end{enumerate}
\Cref{fig:PL_Planning_convergende1t4} visualizes the percentage of successful Dubins path calculations in wind versus the maximum number of allowed iterations. The success rate increases somewhat asymptotically. In \emph{Exp. 1d} and \emph{Exp. 2a}, where $v_\text{wind}>v_\text{air}$, the success rates stay below 30\%. In \emph{Exp. 1c} and \emph{Exp. 3b}, where $v_\text{wind}$ is only slightly below $v_\text{air}$, the success rate increases slowly and only reaches 95\% after 100 iterations. Such slow convergence can slow down the overall planner significantly. A maximum number of 12 iterations is therefore chosen. Obviously, finding feasible Dubins paths in strong wind is harder and may even become infeasible at $v_\text{wind}>v_\text{air}$. \citet{McGee2005} explain that for a convergence rate of 1.0 the path may need to be lengthened artificially. Given that, first, this method is not implemented yet, and second, the number of iterations is limited, the \emph{optimality} property is violated in strong winds and only \emph{near-optimal} paths are returned. 

%%%%%%%%%%%%%%%%%%%%%%%%%%%%%%%%%%%%%%%%%%%%%%%%%%%%%%%
\subsubsection{Motion Cost Heuristic}
\label{sec:PL_Planning_timeoptimalmotioncostheuristic}
%%%%%%%%%%%%%%%%%%%%%%%%%%%%%%%%%%%%%%%%%%%%%%%%%%%%%%%

To allow the use of informed planning and tree pruning in 3D wind fields, the Euclidean motion cost heuristic of \cref{sec:PL_Planning_Improvements_InformedSubset} needs to be extended: First, the Euclidean distance $d_\text{eucl}$ (\cref{eqn:PL_Planning_eucldist}) between the two states $q_1,q_2$ is computed. Then the maximum components of the wind in $x$-, $y$-, and $z$-direction are determined separately for each axis over the whole wind field. This makes sure that the heuristic is actually a lower bound. The magnitude of this vector projected in the direction of the vector from $q_1$ to $q_2$ is identified and denoted $v_\text{wind}^\text{proj}$. Then the time heuristic $h$ is
\begin{equation}
h = \frac{d_\text{eucl}}{v_\text{air} + v_\text{wind}^\text{proj}} \; .
\end{equation}
%Note: This heuristic can easily and needs definitely to be improved for an actual journal publication.

\begin{figure}[htbp]
\centering
\includegraphics[width=0.6\textwidth]{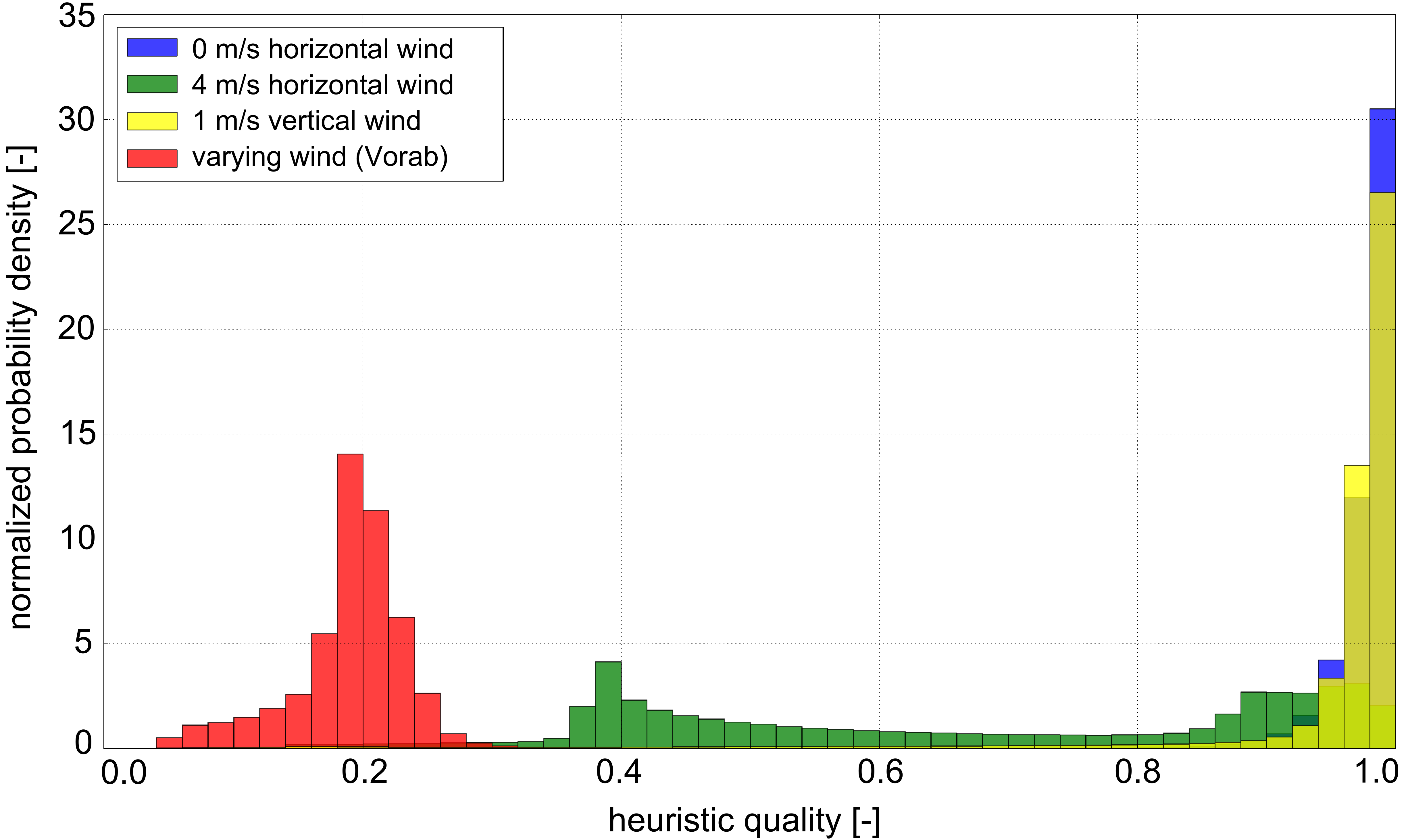}
\caption{The distribution of the heuristic quality, i.e. the fraction of the motion heuristic divided by the actual motion cost, in four different wind fields.}
\label{fig:PL_Planning_histogramheuristic}
\end{figure}

%%%%%%%%%%%%%%%%%%%%%%%%%%%%%%%%%%%%%%%%%%%%%%%%%%%%%%%
\paragraph{Preliminary results}
%%%%%%%%%%%%%%%%%%%%%%%%%%%%%%%%%%%%%%%%%%%%%%%%%%%%%%%
The heuristic must be a lower bound for the actual path cost, but should approximate it as closely as possible such that the heuristic quality $l(q_1,q_2)=h(q_1,q_2)/c(q_1,q_2)\rightarrow 1$. \Cref{fig:PL_Planning_histogramheuristic} assesses the distribution of the heuristic quality in four different cases: a) zero wind b) uniform \unitfrac[4.0]{m}{s} horizontal wind, c) uniform \unitfrac[1.0]{m}{s} vertical wind and d) \emph{Exp. 2a} from \cref{sec:PL_Planning_dubinswithwind}. For each setup 1'000'000 different start and goal states are sampled from a \unit[4000]{m} x \unit[4000]{m} x \unit[500]{m} cuboid. The heuristic approximates the zero-wind test case, which is equivalent to a shortest-path problem, best. On average, $l=0.979$. In the time-optimal case, for a vertical wind of \unitfrac[1]{m}{s} the average quality is $l=0.940$. In the \unitfrac[4.0]{m}{s} horizontal wind the heuristic quality drops to $l=0.664$, and for the wind field from \emph{Exp. 2a} it drops to $l=0.189$. The reason is that one small map region has strong wind of \unitfrac[37]{m}{s}. By taking the maximum wind magnitude over the whole wind field, this small region impacts the heuristic over the whole wind field. The very simple heuristic presented here is valid (i.e. $h(q_1,q_2)<c(q_1,q_2)$) and very fast to compute because it only needs to be calculated once a new wind field is available. It is clearly better than not using a heuristic at all, but it underestimates the cost especially for wind fields with large spatial variations.

%%%%%%%%%%%%%%%%%%%%%%%%%%%%%%%%%%%%%%%%%%%%%%%%%%%%%%%
\subsection{Results}
\label{sec:PL_Planning_Results}
%%%%%%%%%%%%%%%%%%%%%%%%%%%%%%%%%%%%%%%%%%%%%%%%%%%%%%%

\begin{figure}[htbp]
\centering
\subfloat[Scen. \textit{w0}, top down view. {\unitfrac[4.5]{m}{s}} wind in the $x$-$y$ plane.]{\includegraphics[height=4.9cm]{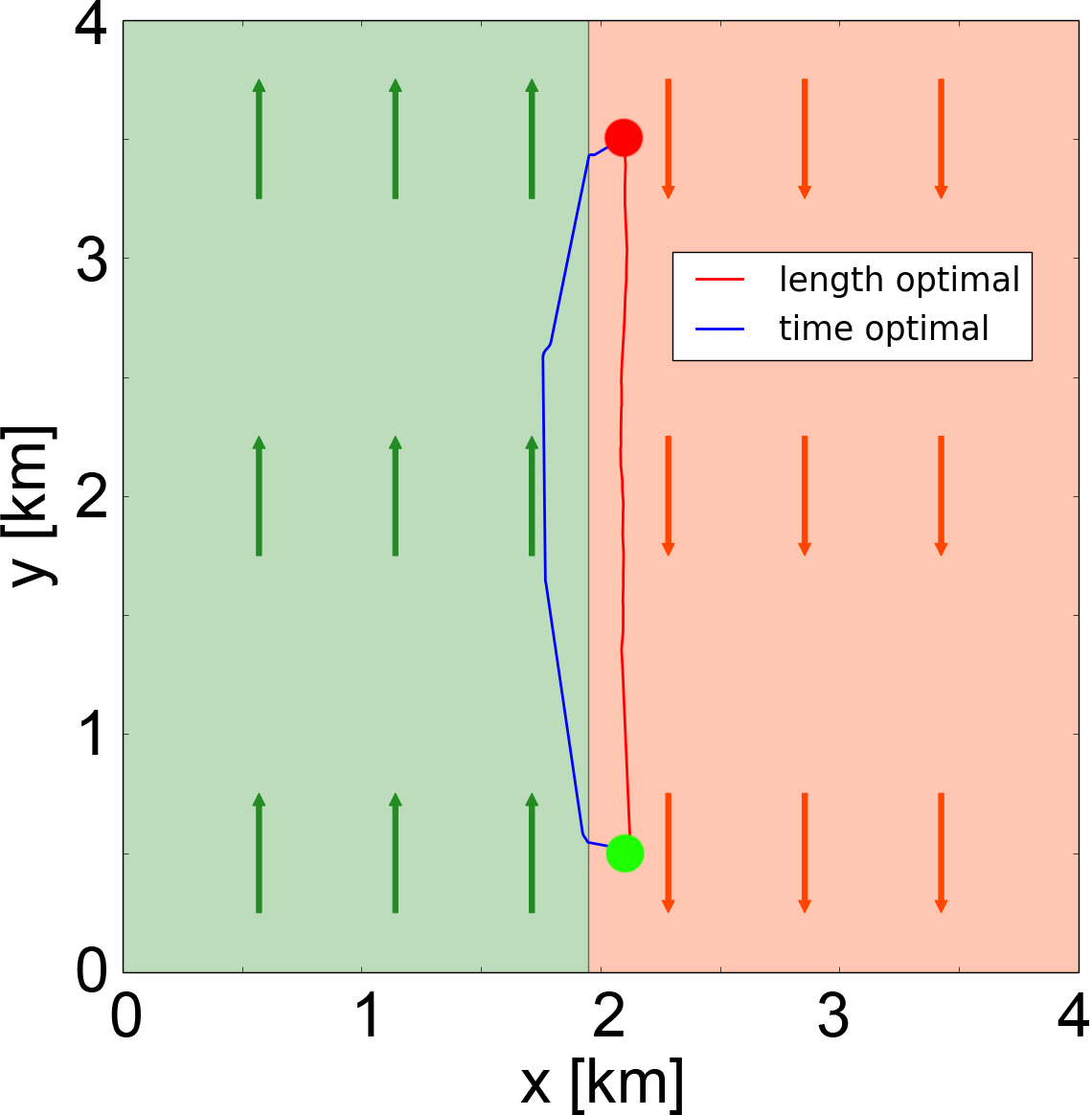}\label{fig:PL_Planning_w0}}
\hfill
\subfloat[Scen. \textit{w1}, top down view. {\unitfrac[6.0]{m}{s}} wind in the $x$-$y$ plane.]{\includegraphics[height=4.9cm]{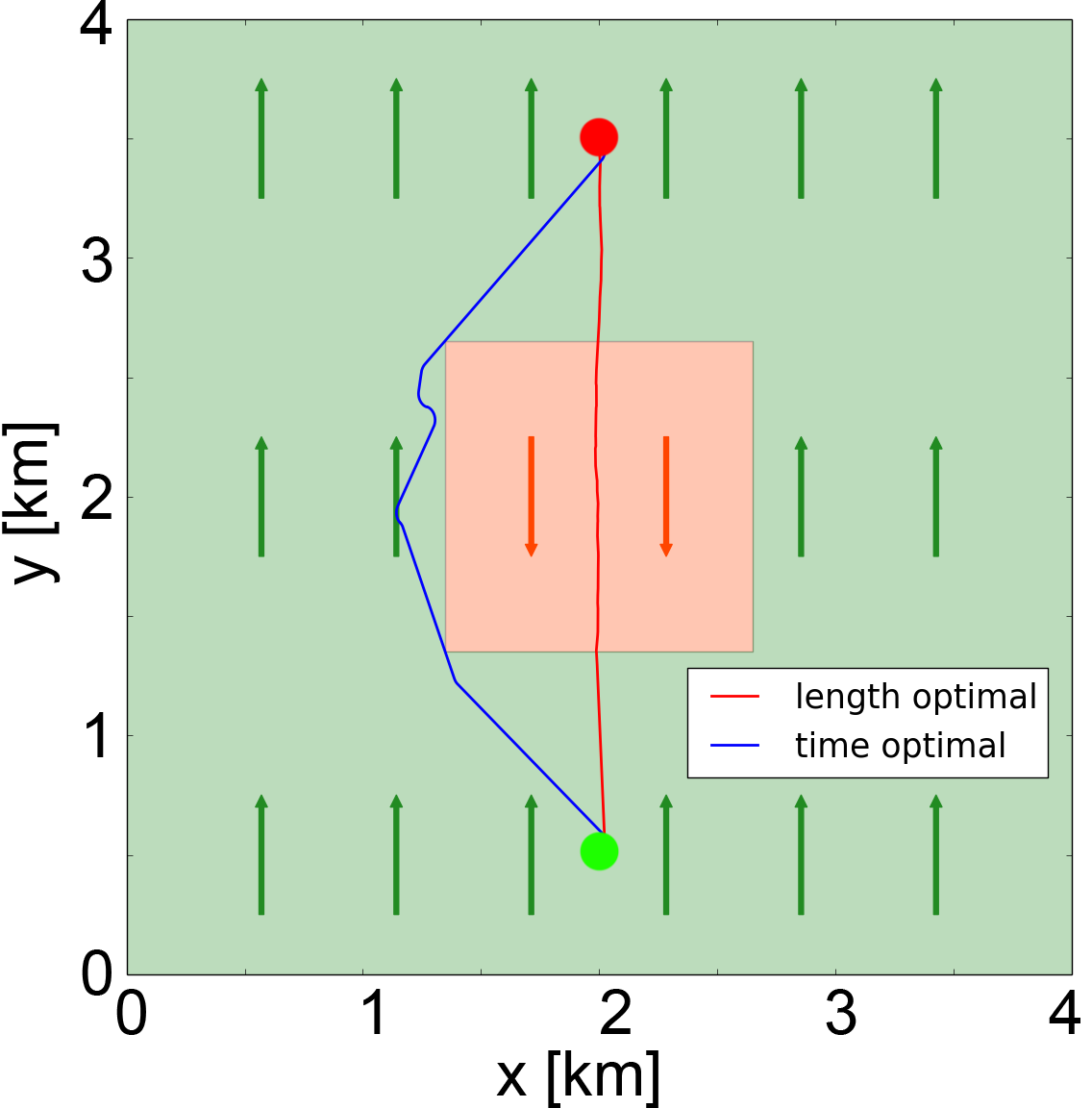}\label{fig:PL_Planning_w1}}
\hfill
\subfloat[Scen. \textit{w2}, side view. {\unitfrac[1.0]{m}{s}} wind in $z$-direction.]{\includegraphics[height=4.9cm]{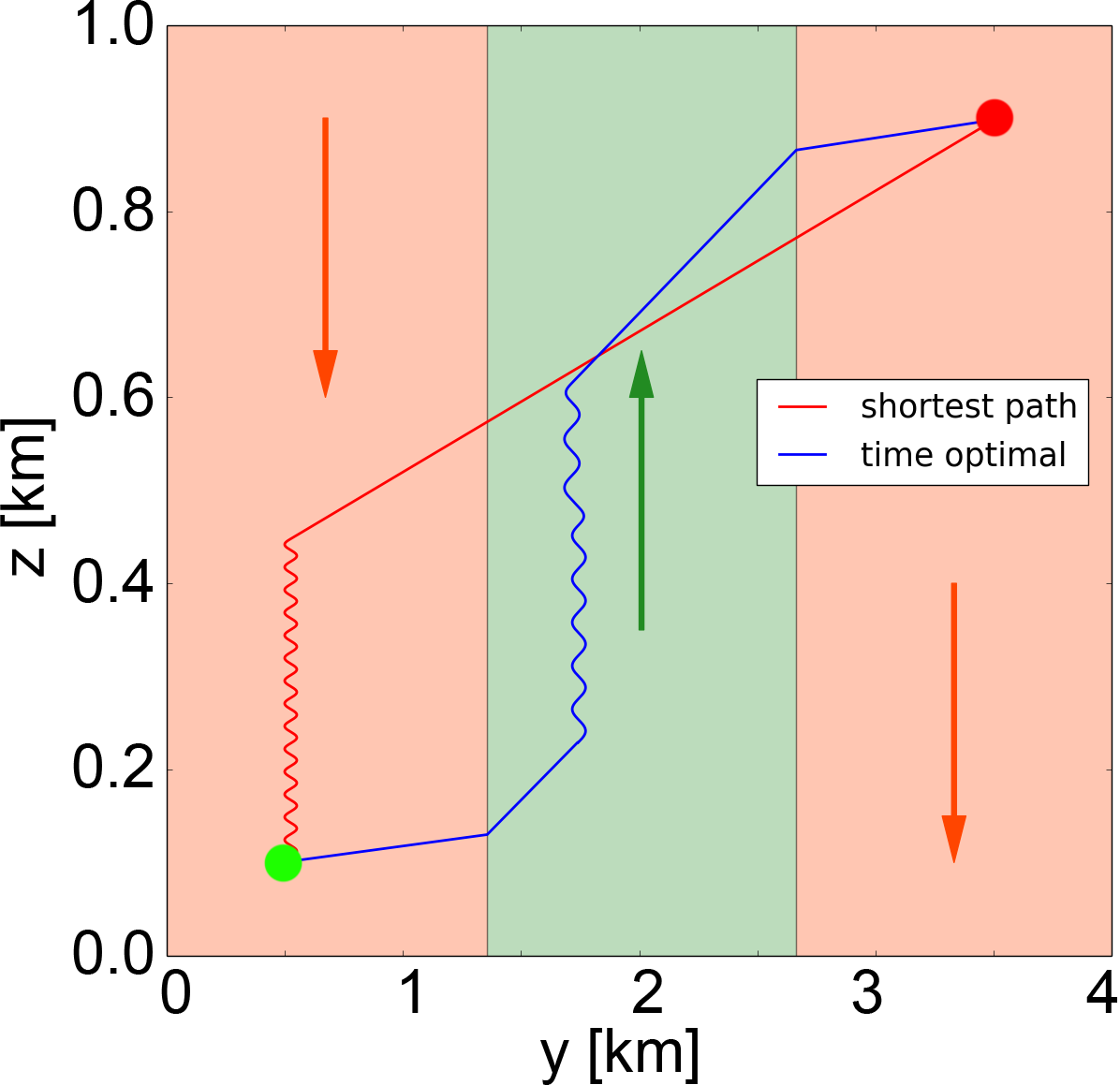}\label{fig:PL_Planning_w2}}
\caption{Synthetic test scenarios for planning in wind. The arrows represent wind, the green and red dot start and goal. The shortest and time-optimal path are shown.}
\end{figure}

%%%%%%%%%%%%%%%%%%%%%%%%%%%%%%%%%%%%%%%%%%%%%%%%%%%%%%%
%\paragraph{Testing Setup}
%%%%%%%%%%%%%%%%%%%%%%%%%%%%%%%%%%%%%%%%%%%%%%%%%%%%%%%
The wind-aware and shortest-path planning are compared by running each of them 100 times for \unit[15]{s}. After each run the planned path is simulated, checked for terrain collisions, and the actual flight time is recorded. Six scenarios that comprise both synthetic test cases and cases with real terrain and wind fields are used:
\begin{itemize}
\item \textit{w0}: No obstacles, horizontal wind of \unitfrac[4.5]{m}{s} (\cref{fig:PL_Planning_w0}).
\item \textit{w1}: No obstacles, horizontal wind of \unitfrac[6.0]{m}{s} (\cref{fig:PL_Planning_w1}).
\item \textit{w2}: No obstacles, vertical wind with \unitfrac[1]{m}{s} ( \cref{fig:PL_Planning_w2}).
\item \textit{w3}: \emph{Kandertal low} terrain (\cref{fig:PL_Planning_Impr_TestcasesKandertalA}) with a synthetic wind field that contains horizontal wind of \unitfrac[6]{m}{s} from $q_\text{start}\rightarrow q_\text{goal}$ for $z>\unit[1850]{m}$ and wind in the opposite direction below this plane.
\item \textit{vorab 1/2}: \emph{vorab 1} is the same as \emph{Exp. 2a} in \cref{sec:PL_Planning_dubinswithwind}. The maximum wind is \unitfrac[37]{m}{s} but close to the ground the wind speed is much lower. \emph{vorab 2} uses an even stronger wind field with a maximum wind of \unitfrac[59]{m}{s}.
\end{itemize}

%%%%%%%%%%%%%%%%%%%%%%%%%%%%%%%%%%%%%%%%%%%%%%%%%%%%%%%
\paragraph{Path cost and feasibility}
%%%%%%%%%%%%%%%%%%%%%%%%%%%%%%%%%%%%%%%%%%%%%%%%%%%%%%%

\begin{figure}[htbp]
\centering
\includegraphics[width = 0.65\textwidth]{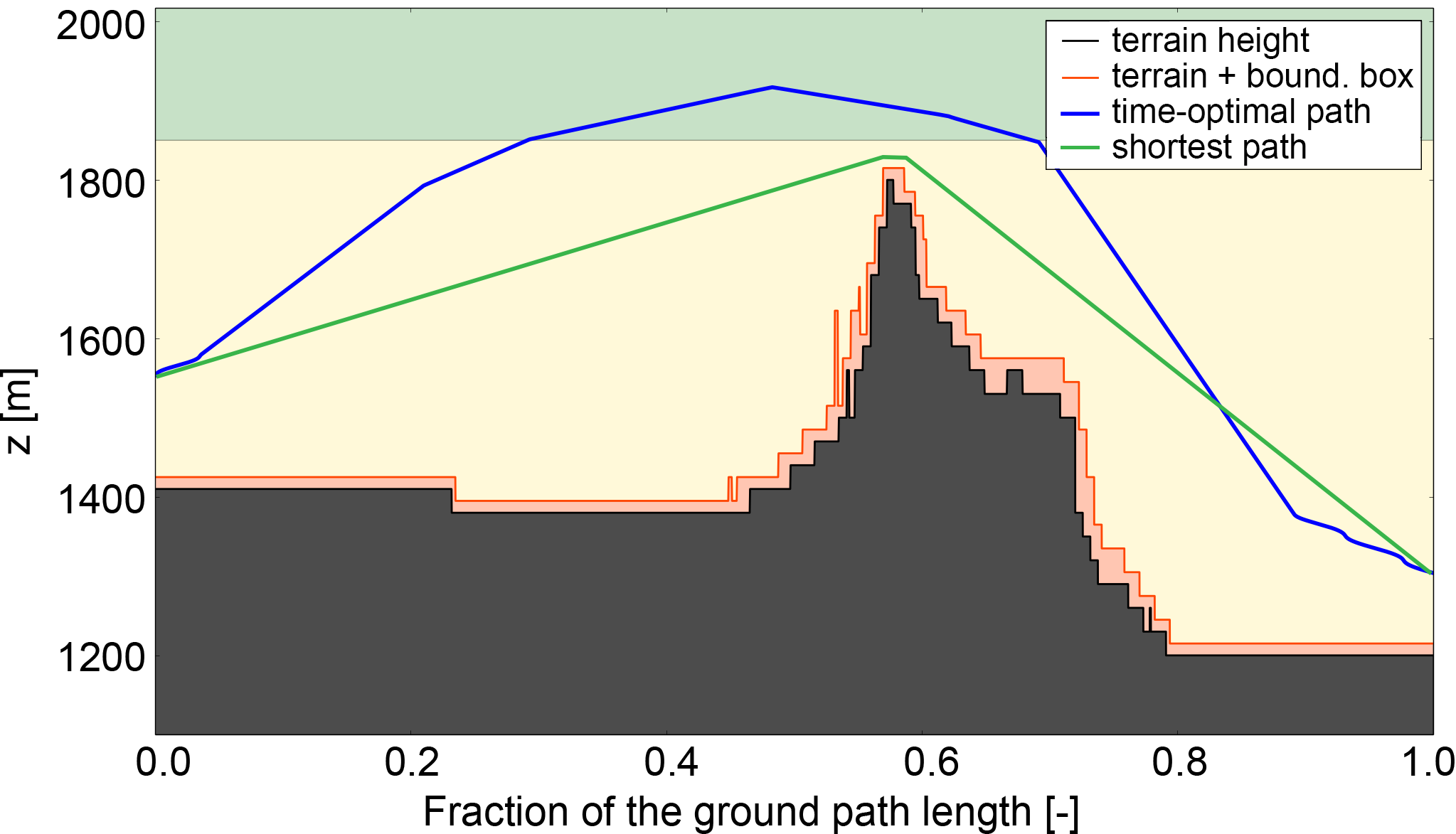}
\caption{A vertical slice along the path of the \emph{w3} scenario. The $x$-axis is the fraction of the ground-relative path length. The winds above \unit[1850]{m} altitude blow from start$\rightarrow$goal and below that in the opposite direction. The shortest path does not consider this fact, but the time-optimal planner clearly leverages the favorable winds at higher altitude.}
\label{fig:PL_Planning_w3}
\end{figure}

\begin{figure}[htbp]
\centering
\subfloat[Path cost, i.e. time required to actually fly the path in wind, normalized by the cost of the time-optimal planner after {\unit[2]{h}} of planning.]{\includegraphics[width=0.48\textwidth]{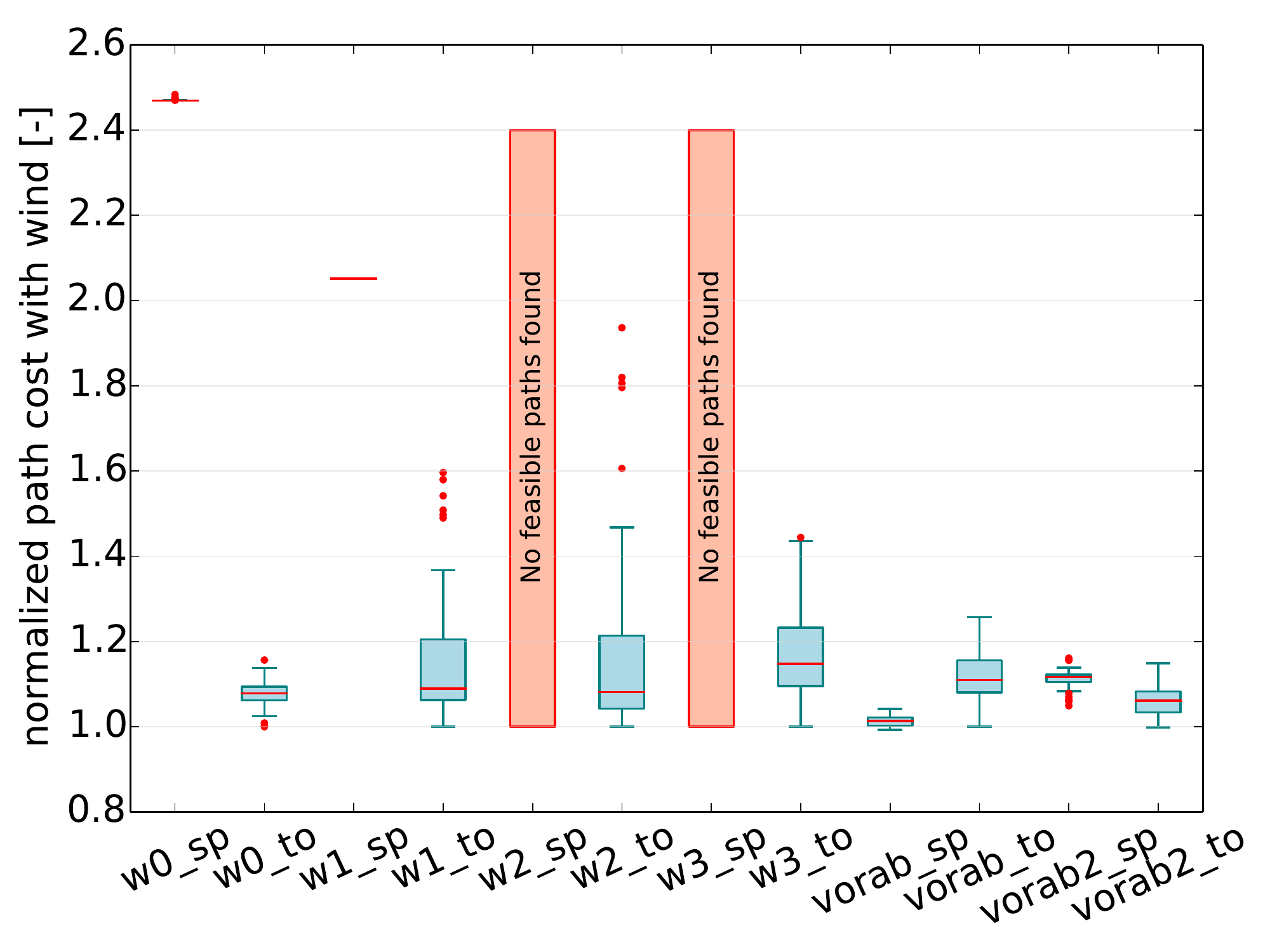}\label{fig:PL_Planning_w0costpath}}
\hfill
\subfloat[Percentage of feasible paths. Paths are infeasible if the wind causes a terrain collision or if the Dubins path in wind cannot be computed.]{\includegraphics[width=0.48\textwidth]{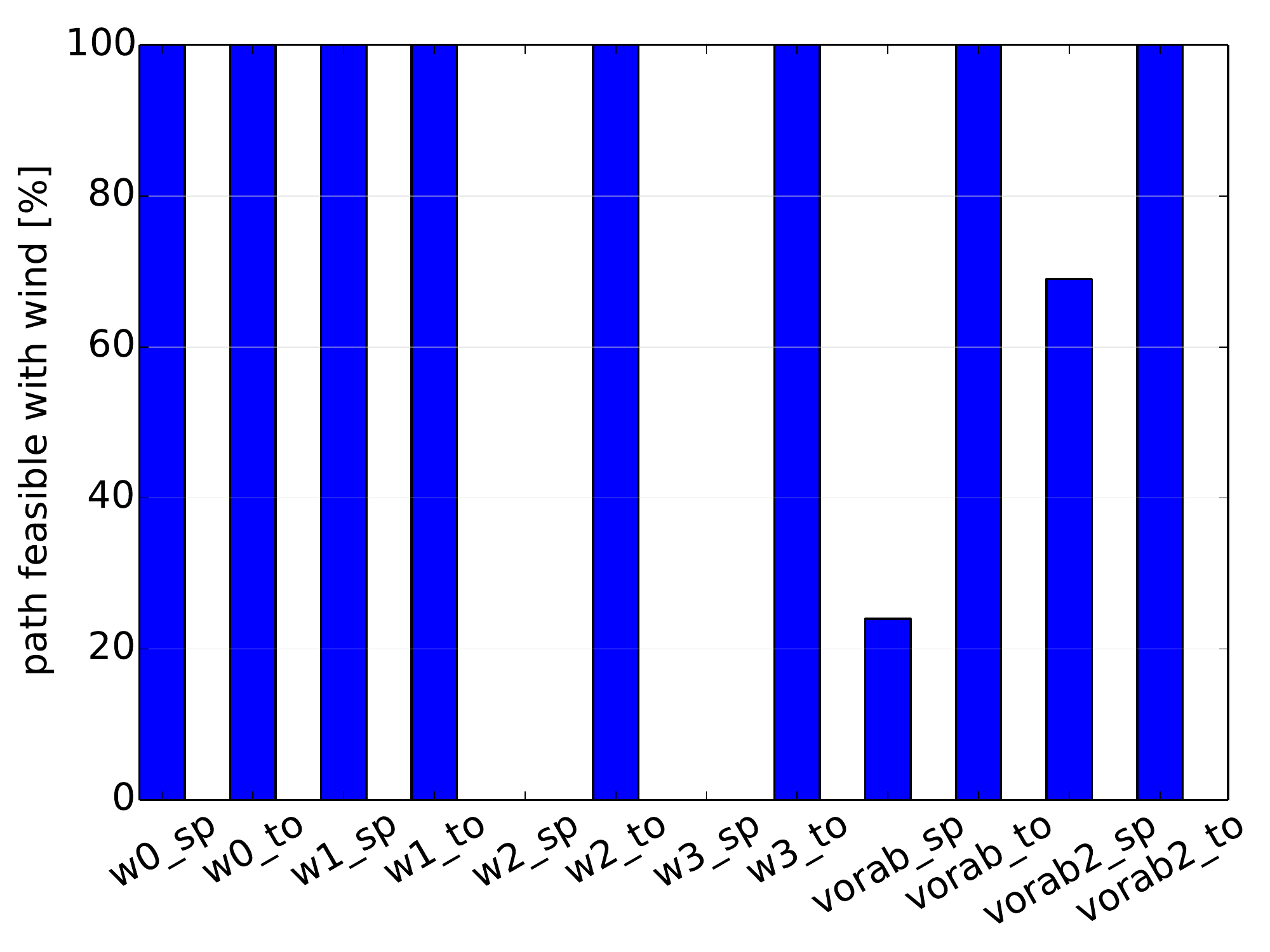}\label{fig:PL_Planning_tofeasible}}
\caption{The cost and percentage of feasible paths for the time-optimal (\_to) and shortest-path (\_sp) planner in case of wind. The TO-planner \emph{always} produces feasible (and thus safe) paths while the SP-planner often causes the aircraft to crash into terrain.}
\end{figure}

As shown in \cref{fig:PL_Planning_w0,fig:PL_Planning_w1}, in scenarios \textit{w0} and \textit{w1} the time-optimal planner always chooses the tailwind areas of the map whereas the shortest-path planner flies through the significant headwind. The time-optimal planner therefore cuts the path cost of the shortest-path planner in half (\cref{fig:PL_Planning_w0costpath}). Similarly, in scenario \emph{w2} shown in \cref{fig:PL_Planning_w2}, the time-optimal planner optimally leverages the column of rising air to gain altitude more quickly. For real missions this means that the planner can exploit thermal updrafts as described in our previous work~\cite{Oettershagen_JFR2018}. More importantly, as shown in \cref{fig:PL_Planning_tofeasible}, in scenarios \textit{w2} and \textit{w3} the time-optimal planner finds feasible paths in 100\% of all cases whereas the shortest-path planner does not provide feasible paths at all. The \emph{w3} scenario (\cref{fig:PL_Planning_w3}) also clearly shows how the planner leverages favorable horizontal winds: It stays at higher altitude and thus in tailwind longer than the shortest path and thereby reaches its goal in less time. The \textit{vorab} scenarios show a mixed picture: The shortest-path planner only finds feasible paths in 22\% and 68\% of all cases for \emph{vorab 1} and \emph{vorab 2} respectively. The rest of the cases are infeasible and unsafe, i.e. result in collision with terrain due to the neglected influence of the wind. While in \emph{vorab 1} the shortest-path planner finds paths with lower cost than the time-optimal planner, these costs are only calculated using the 22\% feasible paths. In addition, in \emph{vorab 2} the time-optimal planner finds lower cost paths by again actively using the stronger winds at higher altitude. 

\begin{figure}[htbp]
\centering
\subfloat[Number of iterations]{\includegraphics[width=0.48\textwidth]{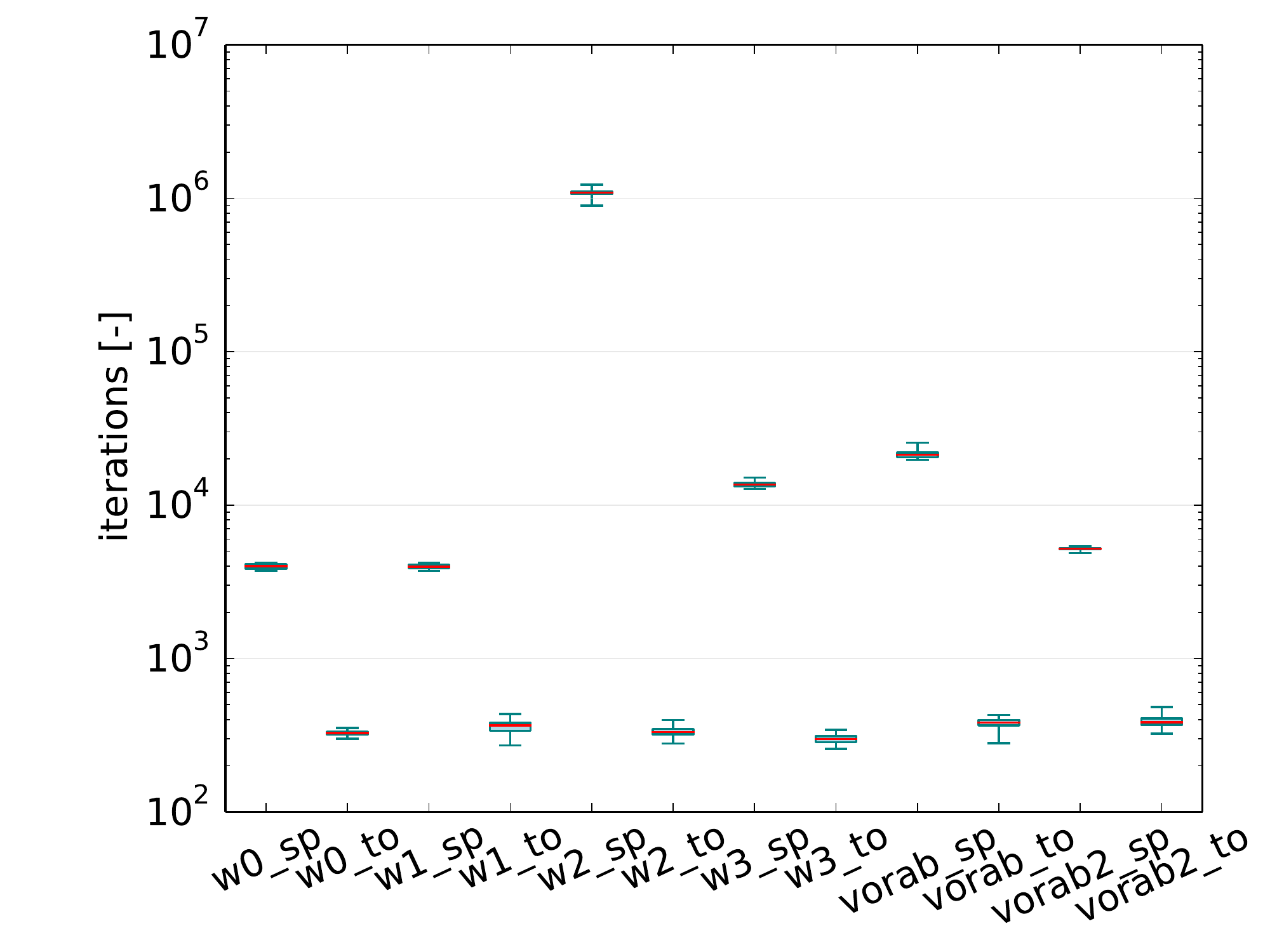}\label{fig:PL_Planning_toiterations}}
\hfill
\subfloat[Normalized cost]{\includegraphics[width=0.48\textwidth]{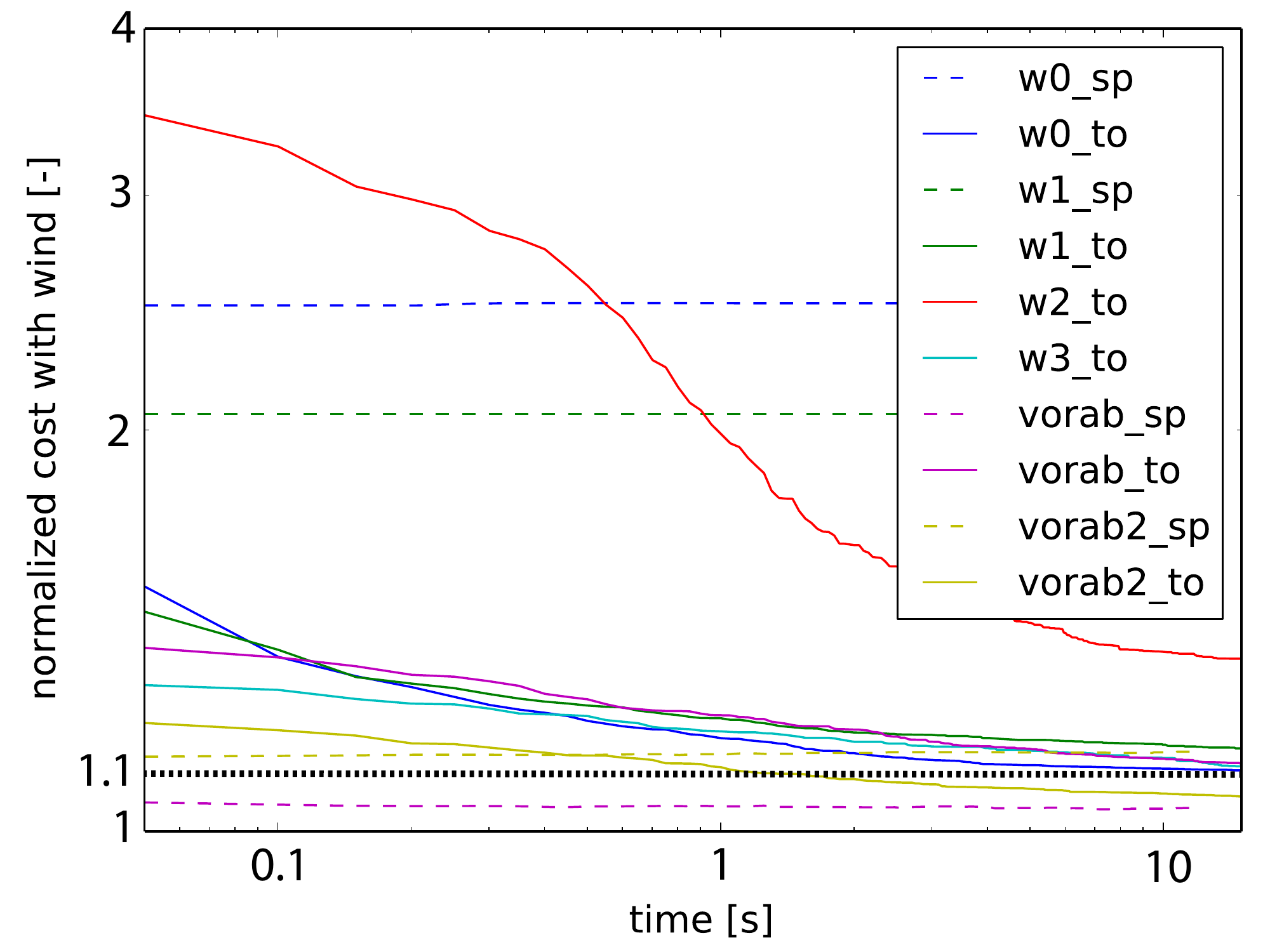}\label{fig:PL_Planning_progresscost1}}
\caption{Number of iterations (left) and cost convergence (right) over \unit[15]{s} for the time-optimal (\_to) and shortest-path (\_sp) planners. On the right, the cost is normalized with the time-optimal path cost after two hours of planning. The costs for w0\_sp and w1\_sp do not converge because the calculated shortest path is not optimal in wind. The results for the w2\_sp and w3\_sp scenarios are not shown as no valid path was found.}
\end{figure}

The reason why the \emph{feasible} shortest paths can be of lower simulated cost than the \emph{feasible} time-optimal paths is that the time-optimal planning, and particularly the iterative Dubins path calculation in wind, is slow. \Cref{fig:PL_Planning_toiterations} shows the amount of iterations performed by both planners in \unit[15]{s}. For the time-optimal planner, the number of iterations is up to 50x lower. As a result, the convergence of the time-optimal planner is much slower (\cref{fig:PL_Planning_progresscost1}). While the shortest-path planner requires only a couple of milliseconds to converge to below 10\% error with respect to the path after two hours of planning, the time-optimal planner needs several seconds. As a result, especially in large maps it may be advantageous to first plan a shortest path and, once more time is available, to then create the wind-aware path. 

%%%%%%%%%%%%%%%%%%%%%%%%%%%%%%%%%%%%%%%%%%%%%%%%%%%%%%%
\paragraph{Influence of the bounding box}
%%%%%%%%%%%%%%%%%%%%%%%%%%%%%%%%%%%%%%%%%%%%%%%%%%%%%%%

\begin{figure}[htbp]
\centering
\subfloat[Percentage of feasible paths.]{\includegraphics[width=0.48\textwidth]{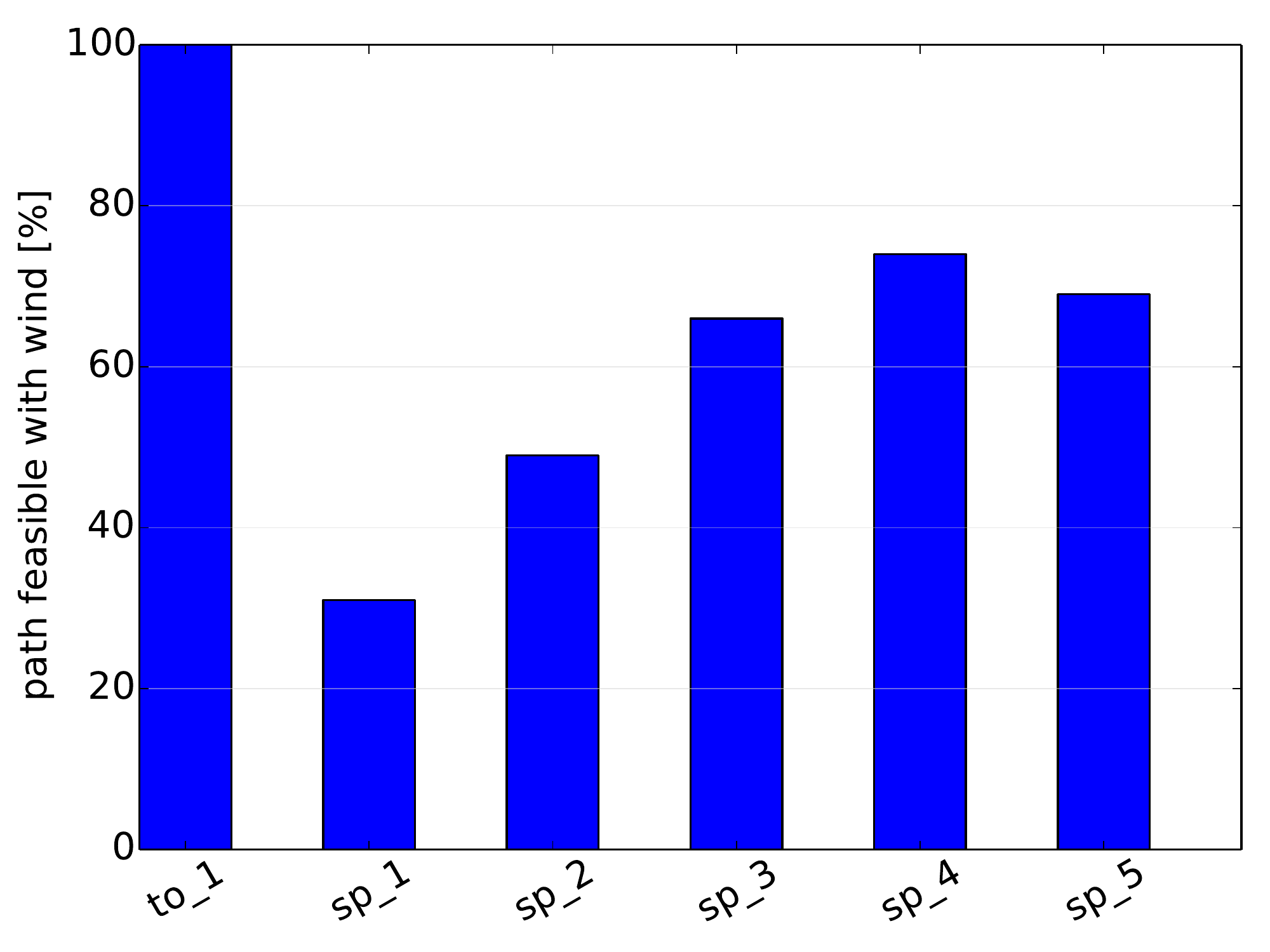}\label{fig:PL_Planning_tobbxfeasible}}
\hfill
\subfloat[Path cost.]{\includegraphics[width=0.48\textwidth]{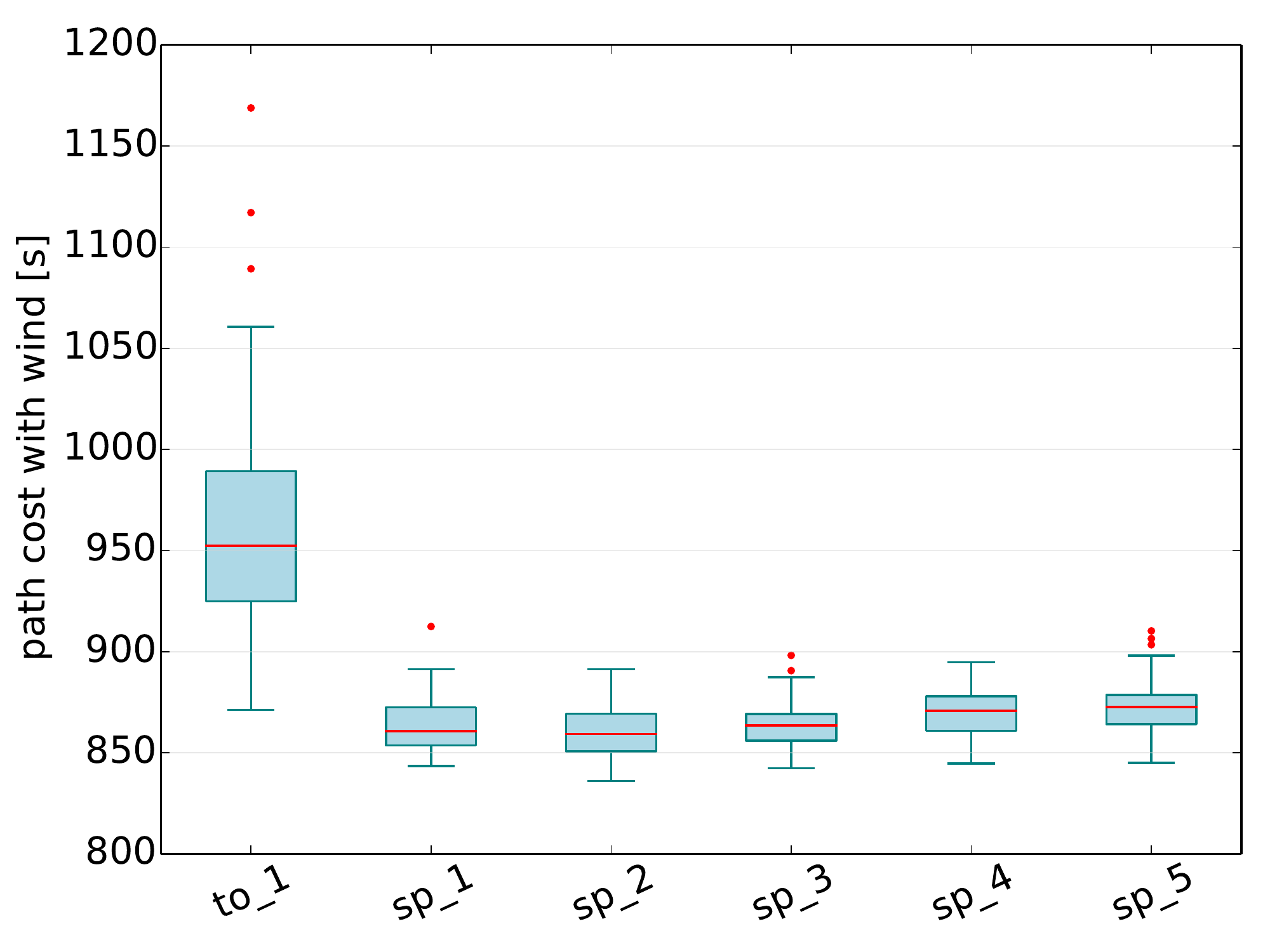}\label{fig:PL_Planning_tobbxcost}}
\caption{Percentage of feasible paths and path cost with different bounding box sizes in the \emph{vorab 1} scenario. The time-optimal (to\_) and shortest-path planner (sp\_) are run for \unit[15]{s}. The bounding box side length is a multiple (appendix of the experiment name) of the original bounding box side length.}
\end{figure}

Given that many shortest paths result in terrain collisions, it is investigated whether simply choosing a larger bounding box can mitigate the problem. \Cref{fig:PL_Planning_tobbxfeasible} shows the percentage of feasible paths with the shortest-path planning and enlarged bounding boxes. More feasible paths are found and only a slight increase of the solution cost is observed (\cref{fig:PL_Planning_tobbxcost}). Therefore, while no feasibility guarantees for the path can be given, increasing the bounding box size is an efficient way to increase the robustness of the shortest-path planner in wind when the time-optimal planner cannot be used (e.g. on very large maps).

%%%%%%%%%%%%%%%%%%%%%%%%%%%%%%%%%%%%%%%%%%%%%%%%%%%%%%%
\paragraph{Effect of the wind magnitude}
%%%%%%%%%%%%%%%%%%%%%%%%%%%%%%%%%%%%%%%%%%%%%%%%%%%%%%%

\begin{figure}[htbp]
\centering
\subfloat[Solution path cost.]{\includegraphics[width=0.48\textwidth]{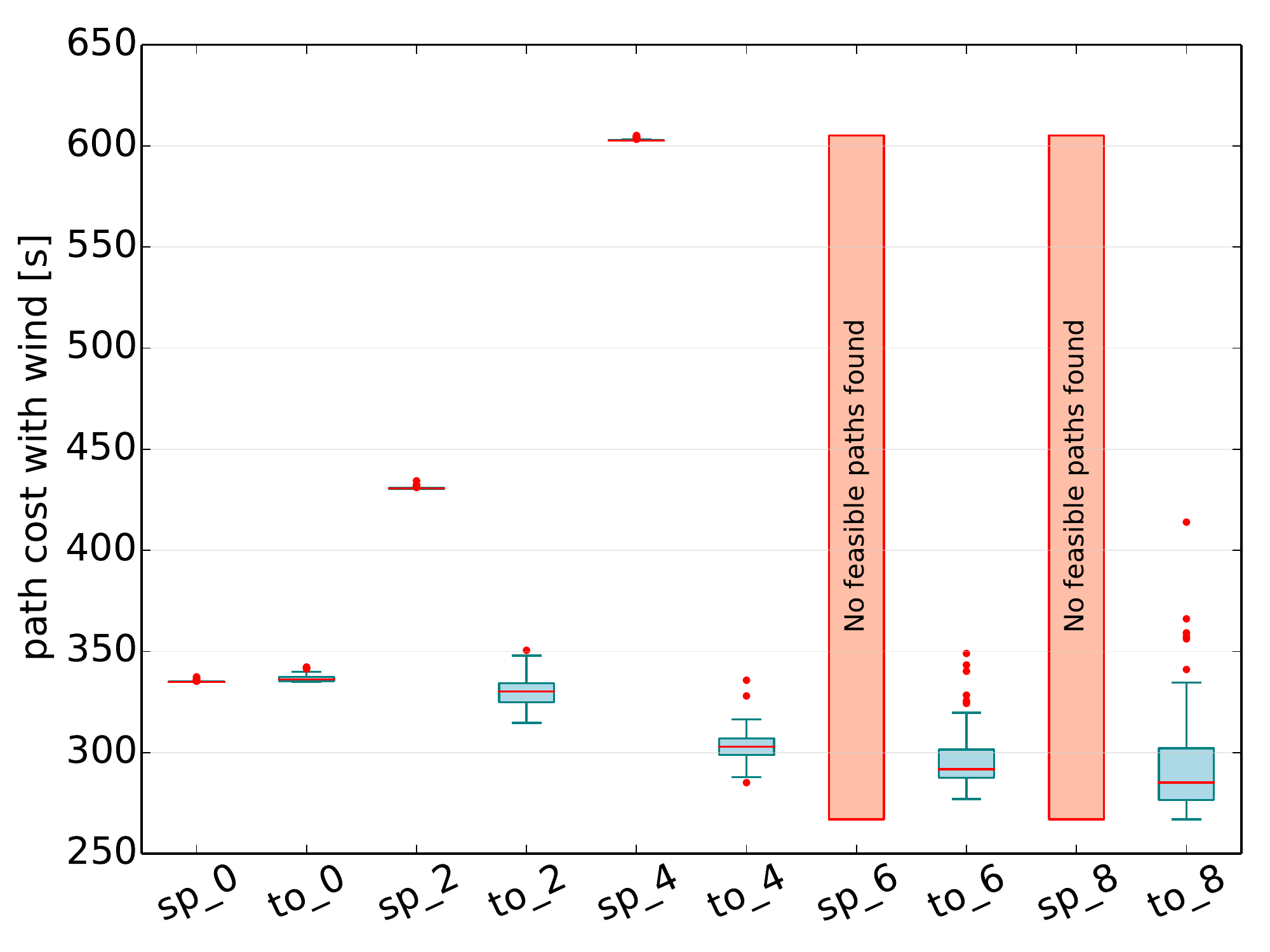}\label{fig:PL_Planning_windw0cost}}
\hfill
\subfloat[Fraction of feasible paths.]{\includegraphics[width=0.48\textwidth]{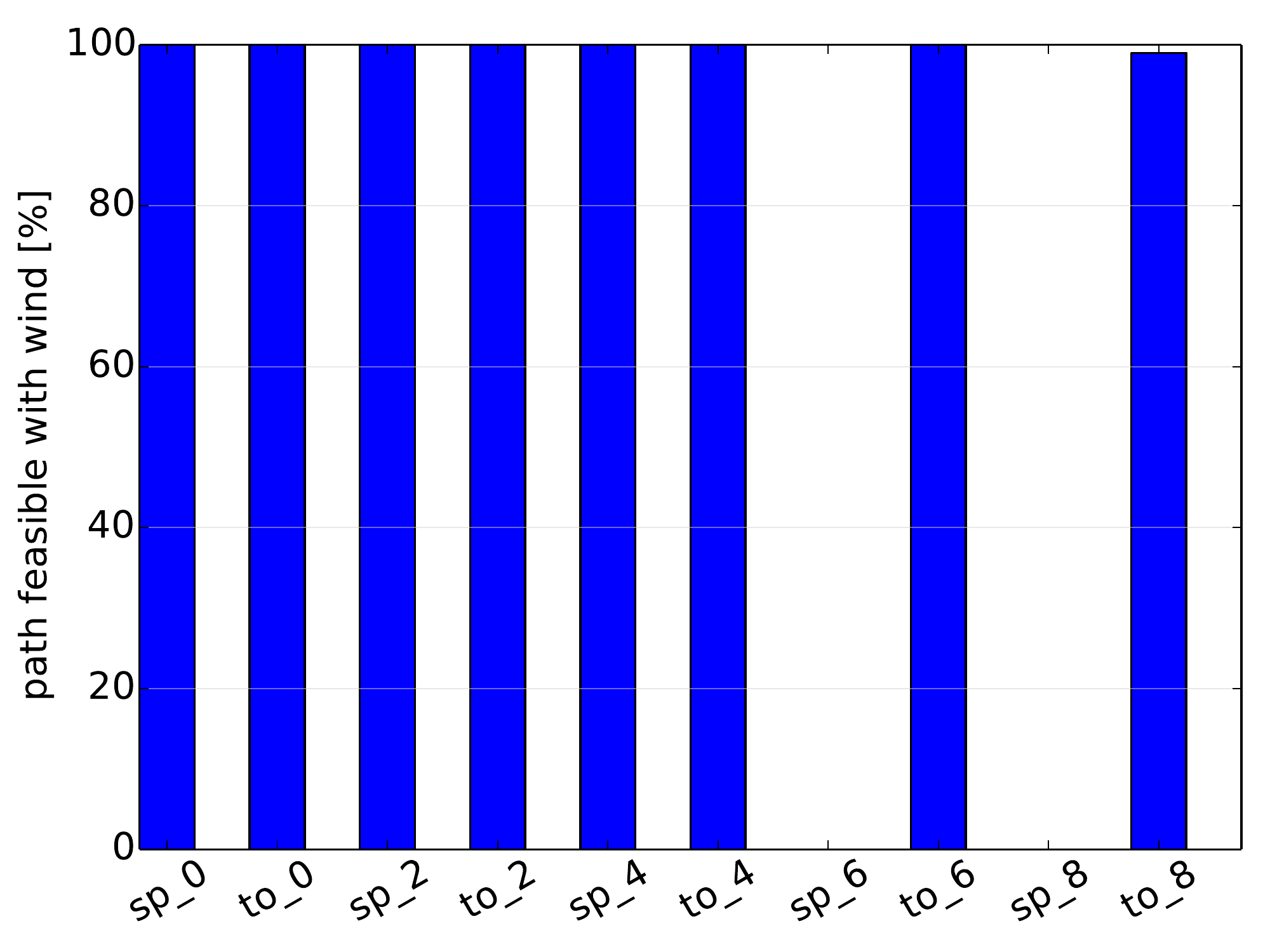}\label{fig:PL_Planning_windw0feasible}}
\caption{Path cost (left) and feasible paths (right) after {\unit[15]{s}} of planning in the \textit{w0} scenario for the shortest-path (sp\_) and time-optimal planning (to\_) and different wind magnitudes. The wind magnitude in {\unitfrac[]{m}{s}} is the number in the experiment name.}
\end{figure}

\Cref{fig:PL_Planning_windw0cost,fig:PL_Planning_windw0feasible}  show the cost and the fraction of feasible paths for the \textit{w0} scenario versus wind magnitudes between \unitfrac[0--8]{m}{s}. Stronger wind results in a lower percentage of feasible paths for the shortest-path planning. The time-optimal planning results in \unit[100]{\%} feasible paths except for when $v_\text{wind}\approx v_\text{air}$. For \unitfrac[0]{m}{s} wind speed both planners yield nearly the same cost because the iterative Dubins path calculation already converges after the first run, i.e. it is not more computationally expensive than the non-iterative Dubins calculation in the shortest-path planner. For larger wind speeds more iterations are required, however, the time-optimal planner then also exploits the stronger wind field better. Given sufficient convergence time, the cost advantages of the time-optimal planner increase at higher wind speed. Note that Achermann~\cite{Achermann2017MT} presents results for the other scenarios which confirm these findings.

%%%%%%%%%%%%%%%%%%%%%%%%%%%%%%%%%%%%%%%%%%%%%%%%%%%%%%%%%%%%%%%%%
%%%%%%%%%%%%%%%%%%%%%%%%%%%%%%%%%%%%%%%%%%%%%%%%%%%%%%%%%%%%%%%%%
\section{Conclusion}
\label{sec:PL_Conclusion}
%%%%%%%%%%%%%%%%%%%%%%%%%%%%%%%%%%%%%%%%%%%%%%%%%%%%%%%%%%%%%%%%%
%%%%%%%%%%%%%%%%%%%%%%%%%%%%%%%%%%%%%%%%%%%%%%%%%%%%%%%%%%%%%%%%%

This paper has presented initial work on a complete real-time environment-aware UAV navigation system: It combines the literature's first 3D wind field prediction system that runs onboard a UAV and a more standard yet performance-optimized wind-aware path planner. The conclusions and potential future work are summarized below.

\paragraph{3D wind field prediction from onboard a UAV}
\label{sec:PL_WindPred_Conclusion}

The onboard 3D wind field prediction is based on downscaling low-resolution wind data from a global weather model. This data is first interpolated to retrieve the initial wind field $\vec{u}^\text{I}$. Using simple potential flow theory, $\vec{u}^\text{I}$ is adjusted to observe the terrain boundaries in a high-resolution 2.5D height map, mass conservation and the atmospheric stratification. Due to the model's simplicity, typical flow fields (i.e. $\unit[1]{km^3}$ size at \unit[25]{m} resolution) are calculated in below \unit[10]{s} on a standard laptop computer, thus allowing near real time execution on the onboard computer of a UAV. 

In synthetic test cases the method yields good qualitative results in the sense that the flow matches an intuitive solution: The flow accelerates within orifices such as a steep valley and both rises and accelerates in front of a ramp. However, given that the model has no actual \emph{physical} understanding of the flow the quantitative results are less accurate, i.e. the wind speed is often underestimated. The comparison to 1D LIDAR profiles collected in the Swiss Alps shows an average horizontal wind error decrease of \unit[41]{\%} over the \emph{zero-wind assumption} that is mostly used in UAV path planning today. In other words, a horizontal wind speed error of \unit[82]{\%} --- close to \emph{AtlantikSolar}'s nominal airspeed --- is reduced to \emph{only} \unit[49]{\%} of \emph{AtlantikSolar}'s airspeed. This clearly improves flight safety by helping the aircraft to avoid dangerous high-wind areas. However, the vertical wind error is increased from \unit[30]{\%} for the zero-wind assumption to \unit[38]{\%} error (again, relative to the aircraft climb or sink rate) for the downscaled wind field. All in all, as shown in \cref{tab:PL_WindPred_rmse}, the method decreases the overall error with respect to a zero-wind assumption by 23\%. It is thus definitely more accurate than not considering any wind during the planning stage at all, and is also slightly better than the pure interpolation of a global weather model.

Still, significant uncertainties in this initial evaluation remain. For example, the increased vertical wind error is \emph{mostly} caused by the \emph{Vorab} LIDAR measurements, which occasionally contain local anomalies such as mountain waves or thermal flows that are also hard to predict for more sophisticated models and human intuition. The accuracy of the LIDAR data might also occasionally be compromised by clouds, fog or precipitation that were not filtered out. To remedy these uncertainties and to improve the framework, the following future work is proposed:
\begin{itemize}
\item \emph{Weather data:} The current LIDAR data is not extensive and reliable enough to allow a final conclusion on the methods accuracy. Given the lack of such data in the literature, comprehensive 3D wind field data should be collected by UAVs with accurate airspeed probes.
\item \emph{Short-term improvements:} Although the method decreases the overall error over the no-wind assumption by 23\%, the vertical errors are still about \unitfrac[0.8]{m}{s}: While this may be accurate enough for standard UAVs, relying only on this model for unsupervised flight of fragile low-power solar-powered UAVs in cluttered terrain is not recommended. Proposed short-term solutions are an increase of the computational domain to resolve large-scale effects (e.g. mountain waves), or to for now only use the horizontal wind prediction that delivers improvements of 41\%.
\item \emph{Future work}: A central problem of the proposed approach is the often low quality of the initial wind field $\vec{u}^\text{I}$, which can not be corrected sufficiently by the simple potential-field based method because its very assumption is that   $\vec{u}^\text{I}$ \emph{is} already a decent estimate. Two very different solutions are:
\begin{itemize}
\item \emph{Improve $\vec{u}^\text{I}$:} UAVs usually measure the wind vector along their trajectory. Fusing this information into the initial wind field can augment $\vec{u}^\text{I}$ substantially. For example, such a method could easily detect the wind direction errors in $\vec{u}^\text{I}$ in \cref{fig:PL_WindPred_valley_breeze}.
\item \emph{Neglect $\vec{u}^\text{I}$}: Methods which do not only \emph{adjust} the initial wind field $\vec{u}^\text{I}$, but which completely \emph{recalculate} the wind field should be investigated. Such methods are more complex and closer to \ac{CFD} approaches, and may require machine learning techniques to speed up the calculation.
\end{itemize}
\end{itemize}

\paragraph{Real-time path planning in wind}
\label{sec:PL_Planning_Conclusion}

The OMPL-based wind-aware path planning framework presented in this paper supports a variety of non-informed (e.g. RRT*) and informed (e.g. IRRT*) planners. It approximates the aircraft dynamics via Dubins airplane motion primitives and retrieves terrain information from 3D octomaps or 2.5D height maps that can be preloaded or generated onboard by a computer vision node. Performance improvements that allow real-time path planning are presented: Obstacle-aware sampling, fast 2.5D map based collision checks and optimal nearest neighbor search are implemented and, for a given planning time, cut the average error with respect to the optimal path in half. To incorporate 3D wind field data, a motion cost heuristic in wind and an iterative Dubins airplane path calculation are implemented. Due to its iterative approach, the time-optimal planner executes up to 50x less iterations per time than the shortest-path planner and thus converges to the optimal solution more slowly. If sufficient computation time is available, then the time-optimal planner's cost advantage over the shortest-path planner increases with wind speed. However, more important than the optimality of the path is its feasibility: While the wind-aware planner always produces collision-free paths in wind, the shortest path often results in terrain collisions. For safe flight through cluttered terrain and complex wind fields, there is therefore no alternative to using a wind-aware path planner.

While the slow convergence of the wind-aware planner is acceptable for flight at high altitude, it is a hindrance when new solutions are required rapidly, such as when flying into previously unmapped terrain where new obstacles may need to be avoided. Speeding up the planner is therefore the main remaining research challenge. For the speed up of local path planning and collision avoidance, two methods are proposed:
\begin{itemize}
\item Wind-aware planning, but with a computational area decreased to the direct vicinity of the airplane (e.g. the area reachable within \unit[20]{s}).
\item Shortest-path planning, but with an enlarged bounding box. As shown in this paper, especially in weaker winds this provides an initial solution on the order of milliseconds. Once more computation time is available, this path should then be improved using the wind-aware planner. 
\end{itemize}
To also speed up the wind-aware planning for large computational domains, we propose to investigate Probabilistic Roadmap (PRM)~\cite{Kavraki1996PRM,Karaman2011RRTstarPRMstar} approaches. These multi-query planners store the graph of wind-aware Dubins airplane edge costs, and replanning tasks are then performed much more quickly by reusing the stored graph instead of recalculating the Dubins motions. Similarly, when exploring new terrain, the graph only has to be extended with new nodes in this area while the rest can be reused. 

%a) multi-query planner (PRM*) or re-planning only for smaller area.
%c) speed ups:
%	- Iterative dubins airplane path calcs  as in [other refs], assume wind uniform in a certain area?
%d) control-based methods for short time horizons
%	- better motion cost heuristic as indicated.
%+ Daniel has nice Future Work section with many ideas. Put this here or in general future work section.

\paragraph{Overall conclusion}

This paper has contributed initial research on the fully-autonomous environment-aware navigation of UAVs by providing a real-time terrain- and wind-aware path planner and the first-ever method for 3D wind field prediction from onboard a UAV. The wind-aware path planner finds feasible collision-free paths even in strong wind and the wind prediction achieves an important improvement over the currently employed no-wind assumption especially in the horizontal wind error. However, a significant body of work remains: As a result, based on the future research challenges outlined in this paper, ETH's Autonomous Systems Lab has recently acquired an INTEL-funded project on real-time machine-learning based 3D wind field prediction in complex terrain. In addition, a research proposal that incorporates wind-aware path planning in Alpine environments was recently handed in to the Swiss National Science Foundation (SNF). This paper has therefore laid the groundwork for follow-up research on first fully-autonomous real-life missions with UAVs in complex terrain.

\subsubsection*{Acknowledgements}
We'd like to thank Prof. Bruno Neininger and  David Braig from the Zurich University of Applied Sciences both for providing the LIDAR wind measurements and assistance in evaluating the data.

%\begingroup
%\bibliographystyle{styles/spmpsci}      % mathematics and physical sciences
\bibliographystyle{plainnat}
\bibliography{refs/refs_all}   % name your BibTeX data base
%\printbibliography
%\endgroup

%\bibliographystyle{apalike}

\end{document}